\documentclass[numsec,webpdf,modern,medium,namedate,preprint]{oup-authoring-template}

\graphicspath{{Fig/}}

\hypersetup{
    colorlinks=true,    
    citecolor=blue,       
    linkcolor=blue,      
    urlcolor=blue,       
    breaklinks=true      
}

\theoremstyle{thmstyleone}
\newtheorem{theorem}{Theorem}
\newtheorem{proposition}{Proposition}
\newtheorem{lemma}{Lemma}       
\newtheorem{corollary}{Corollary} 

\theoremstyle{thmstyletwo}

\newtheorem{remark}{Remark}

\theoremstyle{thmstylethree}
\newtheorem{definition}{Definition}
\newtheorem{assumption}{Assumption}

\usepackage{setspace}
\usepackage[fontsize=12pt]{fontsize}

\newtheorem{innercustomthm}{Theorem}
\newenvironment{mythm}[1]
  {\renewcommand\theinnercustomthm{#1}\innercustomthm}
  {\endinnercustomthm}

\usepackage{bm, dsfont}
\usepackage{enumerate}
\usepackage{mathrsfs, amsmath}
\usepackage{booktabs}

\newcommand{\norm}[1]{\left\|{#1}\right\|}
\newcommand{\abs}[1]{\left|{#1}\right|}

\DeclareMathOperator*{\argmin}{arg\,min}
\DeclareMathOperator*{\argmax}{arg\,max}
\def\R{\mathbb{R}}
\def\N{\mathbb{N}}
\def\Q{\mathbb{Q}}

\def\1{\mathds{1}}
\def\eps{\varepsilon}

\def\P{\mathbb{P}}
\def\Q{\mathbb{Q}}
\def\E{\mathbb{E}}

\let\oldparagraph\paragraph
\renewcommand{\paragraph}[1]{\oldparagraph{\textbf{#1}}}

\allowdisplaybreaks
\begin{document}

\journaltitle{Journals of the Royal Statistical Society}
\appnotes{Original article}

\firstpage{1}


\title[Self-Supervised Learning as Distribution Matching]{Bringing Generative Learning to Representation Learning: Self-Supervised Transfer Learning as Distribution Matching}

\author[1]{Yuling Jiao}
\author[2,3]{Wensen Ma}
\author[3,$\ast$]{Defeng Sun}
\author[4]{Hansheng Wang}
\author[5]{Yang Wang}

\makeatletter
\let\DMpreprintauthors\@author
\def\@author{{\fontsize{11.5}{15}\selectfont\DMpreprintauthors}}
\makeatother

\authormark{Jiao et al.}

\address[1]{\orgdiv{School of Artificial Intelligence}, \orgname{Wuhan University}, \orgaddress{\postcode{430072}, \state{Wuhan}, \country{China}}}
\address[2]{\orgdiv{School of Mathematics and Statistics}, \orgname{Wuhan University}, \orgaddress{\state{Wuhan}, \postcode{430072}, \country{China}}}
\address[3]{\orgdiv{Department of Applied Mathematics}, \orgname{The Hong Kong Polytechnic University}, \orgaddress{\street{Hung Hom}, \state{Kowloon}, \country{Hong Kong SAR, China}}}
\address[4]{\orgdiv{Guanghua School of Management}, \orgname{Peking University}, \orgaddress{\postcode{100871}, \state{Beijing}, \country{China}}}
\address[5]{\orgdiv{Department of Mathematics}, \orgname{The University of Hong Kong}, \orgaddress{\state{Pokfulam}, \country{Hong Kong SAR, China}}}

\corresp[$\ast$]{Address for correspondence: Defeng Sun, Department of Applied Mathematics, The Hong Kong Polytechnic University, Hung Hom, Kowloon, Hong Kong SAR, China. \href{mailto:defeng.sun@polyu.edu.hk}{defeng.sun@polyu.edu.hk}}




\abstract{Most self-supervised learning objectives defend against collapse but leave the target representation law unspecified. We formulate representation learning as Distribution Matching (DM), learning an augmentation-invariant encoder whose push-forward law matches an explicit reference law. The reference law specifies what the learned representation distribution should look like, while the matching mechanism determines how this target is matched. Within DM, generative learning and representation learning can be viewed as directionally inverse distributional processes: generative learning maps a tractable reference to data, whereas representation learning maps data to a designed reference law. For the Mallows-distance instantiation studied here, we connect the population objective to class-centre separation and classification error and derive a finite-sample bound on downstream classification error. Simulations and image benchmarks show manifold rectification, fine-grained structure and transferability.}


\maketitle
\onecolumn
\raggedbottom
\section{INTRODUCTION}\label{section: intro}

\begin{figure}[t]
    \centering
    \includegraphics[width=0.78\linewidth]{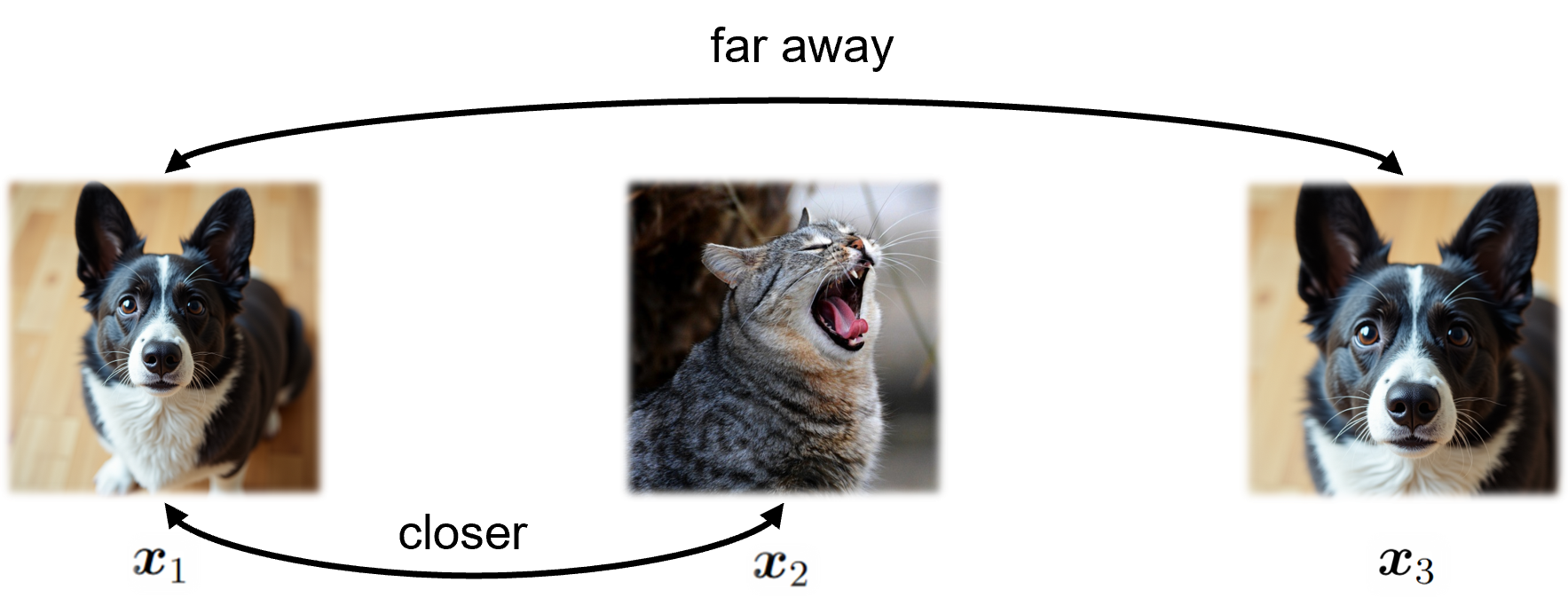}
    \caption{\textbf{Raw Euclidean distance can conflict with semantic meaning.} Although $\bm{x}_3$ is a cropped view of $\bm{x}_1$, $\norm{\bm{x}_1-\bm{x}_2}_2<\norm{\bm{x}_1-\bm{x}_3}_2$.}
    \par\smallskip\noindent\footnotesize\textit{Alt text:} Three images labelled $\bm{x}_1$, $\bm{x}_2$ and $\bm{x}_3$: $\bm{x}_1$ is a dog, $\bm{x}_2$ is a cat, and $\bm{x}_3$ is a cropped view of the dog. Despite $\bm{x}_1$ and $\bm{x}_3$ being two views of the same observation, their raw Euclidean distance exceeds that between $\bm{x}_1$ and $\bm{x}_2$.
    \label{fig: semantic_dist}
\end{figure}

Transfer learning uses a large source sample to improve inference when labelled target observations are scarce. Existing statistical approaches include high-dimensional parametric models \citep{bastani2021predicting,Li2021LinearTransfer,tian2022transferlearninghighdimensionalgeneralized,li2023estimation,he2024transfusion,zhou2024model,gu2025robustangle,zhao2024residualimportanceweightedtransfer,li2025multidimensional} and non-parametric or distribution-shift procedures \citep{cai2019transferlearningnonparametricclassification,kpotufe2021marginal,reeve2021adaptivetransferlearning,chen2021domainadaptation,maity2021linearadjustmentbasedapproach,fan2023robusttransferlearningunreliable,cai2024transferlearningnonparametricregression,zhou2025doubly,cai2025triplyrobust,zhang2025wassersteintransfer}. Recent developments also address feature-based and network-based transfer \citep{xu2025residual,guo2026privacypreserving}. For images and other high-dimensional structured data, representation learning offers a natural route to transfer learning: features learned from a large source sample can support simple inference on downstream tasks with limited labelled data. Learning useful representations requires addressing two related challenges. First, non-parametric rates deteriorate with the ambient dimension $d$ \citep{stone1982optimal,tsybakov2008introduction}. Secondly, raw Euclidean distance need not agree with semantic similarity. Figure~\ref{fig: semantic_dist} gives a simple example: an augmented view can be farther from its source image than a different observation. Representation learning is therefore not merely dimension reduction; it must replace an uninformative raw geometry by one in which low-complexity downstream inference is meaningful.

Supervised pretraining can learn such a geometry from a large annotated source sample \citep{schroff2015facenet,dhillon2020baseline,jiao2024deep}. It reduces dimension and uses labels to determine which aspects of the signal should be retained. Its usefulness is consequently bounded by the availability and scope of those labels: features irrelevant to the source task may be important for a later target task. Self-supervised learning (SSL) instead aims to learn representations that capture useful semantic information from unlabelled data, without relying on manually annotated labels. A common strategy seeks to learn representations that are invariant to semantics-preserving data augmentations by aligning the representations of different augmented views of the same observation. Alignment alone, however, admits a collapsed solution that maps all observations to the same constant representation.

Most SSL objectives address collapse through what we call a \emph{defensive policy}. Contrastive methods repel negative pairs \citep{chen2020simclr,he2020momentum}; clustering-based methods enforce consistency between balanced cluster assignments \citep{caron2020swav}; whitening and variance-based methods constrain second-order or coordinate-wise variability \citep{zbontar2021barlow,ermolov2021whitening,adrien2022vicreg,zhang2024logDet}. These devices are effective, but often work by imposing constraints that rule out undesirable representations, rather than directly specifying what the learned representation distribution should look like. Our focus is to reformulate SSL as a distribution matching problem and explicitly specify what the learned representation distribution should look like.

We propose \emph{Distribution Matching} (DM), a constructive statistical formulation of SSL centred on explicit reference design. We first prescribe a reference distribution $\P_{\mathcal R}$ that specifies the desired representation geometry. We then learn an encoder $f$ for which the push-forward distribution $\P_f=f_\sharp\P_{\mathcal A}$ matches $\P_{\mathcal R}$ while augmented views of the same observation remain aligned; here $\P_{\mathcal A}$ is the law of augmented source observations. Thus the object to be learned is explicit: a useful representation distribution should resemble the chosen geometric reference. Different matching objectives and learned transport mechanisms provide alternative ways of pursuing this goal. For the concrete instantiation studied here, we choose Mallows distance and a balanced reference mixture of concentrated components around orthogonal directions. Alignment keeps augmented views of the same observation close, while distribution matching makes the representation distribution follow the prescribed reference law. The DM framework separates what to match from how to match it.

A broader benefit of the DM framework is that it places generative learning and representation learning in a common distributional language, under which they can be viewed as directionally inverse processes: generative learning transports a tractable reference law towards a complex data law, whereas representation learning under DM transports the data law towards an explicitly designed representation law while preserving augmentation invariance. DM thus provides a principled route for transferring tools widely used in generative learning to representation learning.

This formulation also allows the reference law to be specified independently of the downstream label space. We examine the implications of this flexibility empirically in Section~\ref{subsection: finer-grained concept} and theoretically in Section~\ref{section: theory}.

\subsection{Contributions}\label{subsection: contributions}

The contributions are threefold.
\begin{enumerate}[(a)]
    \item \textbf{A constructive formulation and a distributional bridge.} DM reformulates SSL as matching the encoder push-forward to a prescribed reference law. Our reference design is motivated by clustering observations according to their semantic meaning through augmentation-induced connections between observations (see Appendix~\ref{section: augmentation distance interpretation} for their interpretation via $(\sigma,\delta)$-augmentation). Its separated components provide a model-design route to understanding and exploring fine-grained representations. DM separates what to match from how to match it and, through its directionally inverse view of generative learning and representation learning, opens a route for transferring generative tools; Mallows--Wasserstein is the concrete instantiation studied here.
    \item \textbf{An end-to-end statistical analysis.} A population result links the DM objective to separation of target class centres and the error of a low-complexity downstream rule. A non-asymptotic analysis then controls the population objective attained by a neural sieve estimator. The result separates the cost of learning the representation using unlabelled data from that of fitting the target rule.
    \item \textbf{Empirical assessment of the proposed geometry.} Simulations and image benchmarks examine manifold rectification, sensitivity to the number of reference components, and cross-dataset transfer. These experiments evaluate not only predictive accuracy but also the structural interpretation supplied by the reference distribution.
\end{enumerate}

\subsection{Related work}\label{subsection: related works}

\citet{wang2020understanding} identify alignment and uniformity as desirable representation properties, establish their connection to the InfoNCE loss used in SimCLR \citep{chen2020simclr} through its infinite-negative limit, and demonstrate the effectiveness of directly optimising both. DM formulates SSL more generally as matching the representation distribution to an explicitly designed reference law, separating what to match from how to match it. The structured reference distribution introduced in this paper, a concrete instance within the DM framework, is motivated by the desired representation geometry: representations should form separated clusters according to semantic meaning. This design not only enables us to explore fine-grained semantic concepts by allowing more reference components than downstream classes but also plays a key role in our theoretical analysis. \citet{butakov2025efficient} induce selected prior laws by injecting noise while retaining the Deep InfoMax objective. More recently, \citet{mikulasch2026latent} formulate SSL as matching a joint latent law to a probabilistic latent model through a KL objective, thereby unifying several contrastive, non-contrastive and predictive methods. Their aim is complementary to ours: we prescribe an interpretable marginal reference geometry and connect its discrepancy to downstream transfer risk.

Population analyses of contrastive SSL include the latent-class framework of \citet{arora2019theoretical} and the augmentation-based analysis of \citet{huang2023towards}, with extensions to adversarial settings in \citet{duan2025advssl}; graph-based analyses include \citet{haochen2021spectral,haochen2022beyond}. Related statistical analyses investigate representation geometry, information retention and downstream performance \citep{lin2025sufficient,oko2025contrastive,gui2025intrinsic,cai2026multimodal,li2026hierarchical,dong2025contrastivenetwork,wang2023metric,wang2025augmentation,wang2025linear,wang2026token}. Our analysis relates the discrepancy between the learned representation law and the reference law to class-centre separation and downstream transfer risk.

The Wasserstein-based instance of DM studied here is also related to Wasserstein generative modelling \citep{arjovsky2017wasserstein,gulrajani2017wgangp}, but reverses its statistical direction: generative modelling maps a simple reference distribution towards the data distribution, whereas DM learns an encoder that maps the data distribution towards an explicitly designed representation distribution.

Inspired by DM's directionally inverse view of generative learning and
representation learning, subsequent Flow-Based Distribution Matching (FBDM)
transfers flow matching \citep{lipman2023flowmatching}, a widely used generative learning framework, to
representation learning and achieves competitive performance with
state-of-the-art SSL methods across benchmarks ranging from CIFAR to ImageNet
\citep{jiao2026fbdm}.

\section{METHODOLOGY}\label{section: methodology}

\subsection{Statistical setting and self-supervision}

Let $\mathcal{D}_S=\{X_S^{(i)}:1\leq i\leq n_S\}$ be i.i.d. unlabelled observations from $\P_S$ on $[0,1]^d$, and let $\mathcal{D}_T=\{(X_T^{(i)},Y_i):1\leq i\leq n_T\}$ be i.i.d. observations from $\P_T$, with $Y_i\in\{1,\ldots,K\}$ for $K$ target classes and typically $n_S\gg n_T$. We seek an encoder $f:\R^d\to\R^{d^\star}$, $d^\star\ll d$, learned without source labels and subsequently used for target classification. Let $\mathcal X\subseteq[0,1]^d$ contain the supports of the source, target and augmented observations. For constants $L,R>0$, the population analysis uses
\begin{equation}\label{eq: functional class}
    \mathcal{F}=\left\{f:\R^d\to\R^{d^\star}: f\in\mathrm{Lip}(L),\ \norm{f(\bm{x})}_2=R\ \text{for all }\bm{x}\in\mathcal X\right\}.
\end{equation}
The fixed-radius condition is widely used in self-supervised learning, where representations are commonly normalised onto a hypersphere \citep{chen2020simclr,wang2020understanding}.

Let $\mathcal A$ be a family of transformations equipped with an augmentation law $U_{\mathcal A}$. For $X\sim\P_S$ and $A\sim U_{\mathcal A}$ sampled independently, the augmentation distribution $\P_{\mathcal A}$ is defined, for every measurable $E\subseteq\mathcal X$, by
\begin{equation}\label{eq: Pa definition}
    \P_{\mathcal A}(E)=\int_{\mathcal A}\P_S\{A^{-1}(E)\}\,dU_{\mathcal A}(A).
\end{equation}
The alignment loss is
\begin{equation}\label{eq: alignment loss}
    \mathcal L_{\mathrm{align}}(f)=\E_{X\sim\P_S}\E_{A_1,A_2\sim U_{\mathcal A}}
    \norm{f(A_1(X))-f(A_2(X))}_2^2,
\end{equation}
where $A_1$ and $A_2$ are independent. Alignment encodes augmentation invariance but, by itself, is minimized by a constant encoder.

\subsection{Distribution Matching}

\begin{figure}[!t]
    \centering
    \includegraphics[width=0.47\linewidth]{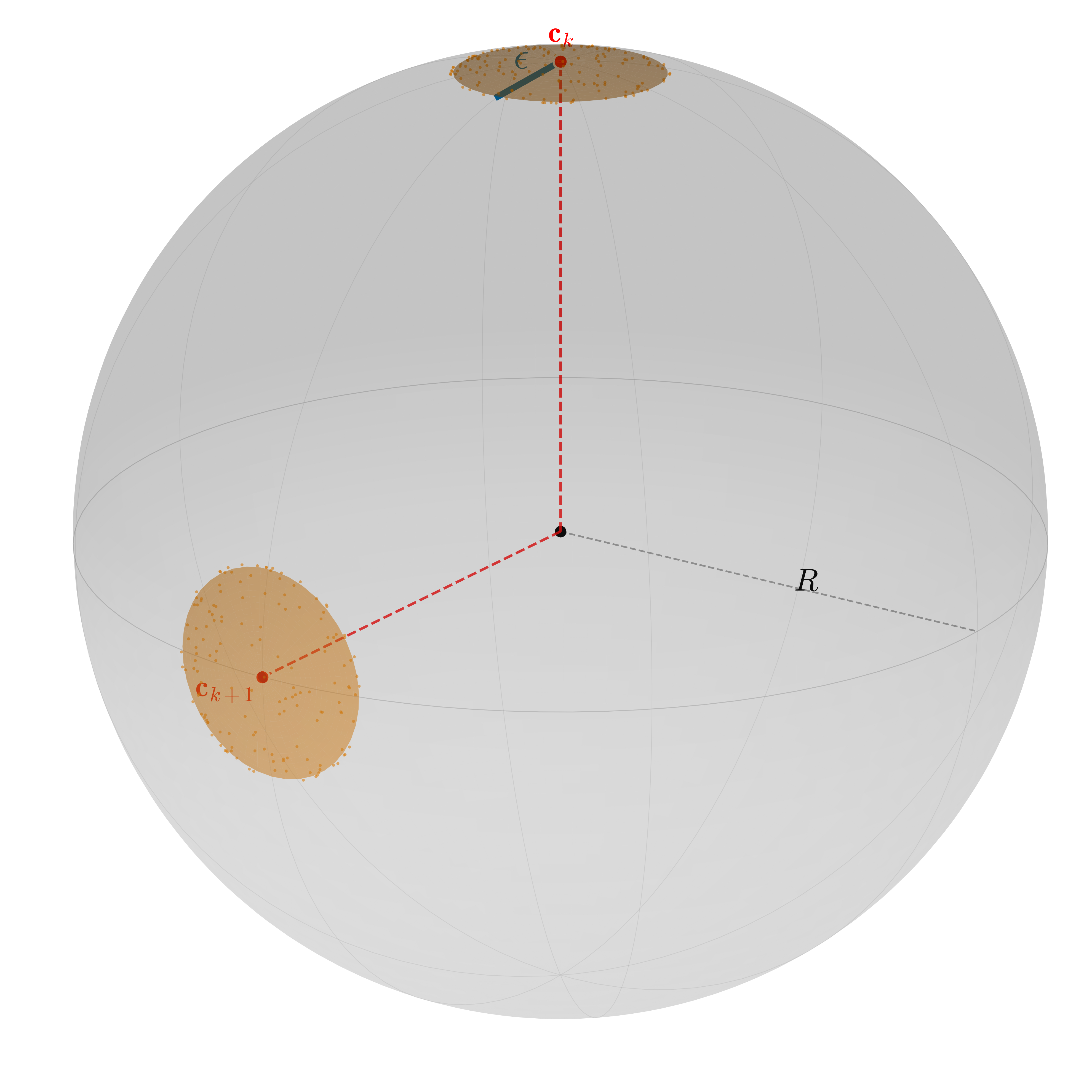}
    \caption{\textbf{The prescribed reference distribution.} To draw from
    $\P_{\mathcal R}$, a component is selected with probability $\alpha_k$,
    its centre direction $\bm c_k$ is perturbed along a uniformly sampled unit
    direction by an amount $\epsilon$, as in (\ref{eq: reference component}),
    and the result is normalised to the
    radius-$R$ hypersphere. Two components are shown; $\epsilon$ controls
    their within-component spread.}
    \par\smallskip\noindent\footnotesize\textit{Alt text:} A grey radius-$R$
    hypersphere contains two orange point clouds concentrated around the
    orthogonal centre directions $\bm c_k$ and $\bm c_{k+1}$; the parameter
    $\epsilon$ indicates the within-component spread around $\bm c_k$.
    \label{fig: reference distribution}
\end{figure}

DM replaces defensive collapse prevention by an explicit target for the representation law. Let $\P_f=f_\sharp\P_{\mathcal A}$ denote the distribution of $f(A(X))$. As one possible choice within the DM framework, we prescribe a reference distribution $\P_{\mathcal R}$ on the radius-$R$ hypersphere in $\R^{d^\star}$. Let $K'\leq d^\star$ be the number of reference components. Choose $K'$ mutually orthogonal centres $\{\bm c_k\}_{k=1}^{K'}$, with each $\bm c_k$ a signed standard basis vector. The $k$th component is defined by
\begin{equation}\label{eq: reference component}
    \mathcal P_k=R\frac{\bm c_k+\epsilon\bm\gamma_{d^\star}/\norm{\bm\gamma_{d^\star}}_2}
    {\norm{\bm c_k+\epsilon\bm\gamma_{d^\star}/\norm{\bm\gamma_{d^\star}}_2}_2},
    \qquad \bm\gamma_{d^\star}\sim\mathcal N(\bm0,\mathbf I_{d^\star}).
\end{equation}
Figure~\ref{fig: reference distribution} visualises this construction.
Each component is concentrated around an orthogonal direction on the
radius-$R$ hypersphere, with its spread controlled by $\epsilon$.
If $\mathcal C\in\{1,\ldots,K'\}$ has probabilities $\alpha_k$, then $\mathcal R=\sum_{k=1}^{K'}\1\{\mathcal C=k\}\mathcal P_k$ has law $\P_{\mathcal R}$. We use $\alpha_k=1/K'$ unless prior information suggests otherwise.

At the framework level, DM learns an encoder whose push-forward law
$f_\sharp\P_{\mathcal A}=\P_f$ matches the prescribed reference law
$\P_{\mathcal R}$ while preserving augmentation invariance. Its
distributional direction is
\begin{equation*}
    \P_{\mathcal A}\xrightarrow{f_\sharp}\P_f
    \rightsquigarrow\P_{\mathcal R}.
\end{equation*}
The first arrow denotes the encoder push-forward, and the matching arrow
$\rightsquigarrow$ denotes the reference-directed learning goal. The
reference law specifies what to match; the learning mechanism specifies how
to pursue that goal. Possible mechanisms include direct minimisation of a
distributional discrepancy and learning a reference-directed transport
through generative tools such as flow matching.

For a chosen discrepancy $\mathcal D$, combining distribution matching with
augmentation alignment gives the population objective
\begin{equation}\label{eq: DM objective}
    f^\star\in\argmin_{f\in\mathcal F}\mathcal L(f)
    :=\mathcal L_{\mathrm{align}}(f)+\lambda\mathcal D(\P_f,\P_{\mathcal R}),
\end{equation}
where $\lambda>0$ and $\mathcal D$ may be any discrepancy between probability
distributions for which the objective is well defined. Its choice specifies
how deviations are penalised, not the reference law to be matched.

\begin{samepage}
An example is first-order Mallows--Wasserstein distance. For probability measures
$\nu_1,\nu_2$ with finite first moments, this distance with Euclidean ground
cost is
\begin{equation}\label{eq: Mallows distance}
    \mathcal W(\nu_1,\nu_2)=\inf_{\pi\in\Pi(\nu_1,\nu_2)}
    \int\norm{\bm z_1-\bm z_2}_2\,d\pi(\bm z_1,\bm z_2),
\end{equation}
where $\Pi(\nu_1,\nu_2)$ is the set of couplings \citep{mallows1972note,Villani}.
\end{samepage}
For the concrete algorithm and theoretical analysis developed below, we take
$\mathcal D=\mathcal W$, as defined above. We use $\mathcal W$ for its convenience in our analysis.
Equation~\eqref{eq: DM objective} combines invariance at the observation
level with the prescribed marginal geometry of the representation.

\subsection{Estimation and computation}\label{subsection: estimation and computation}

For the Mallows--Wasserstein instantiation, Kantorovich--Rubinstein duality,
with the fixed Lipschitz scale absorbed into
$\lambda$, rewrites \eqref{eq: DM objective} as
\begin{equation}\label{eq: dual objective}
    \mathcal L(f)=\mathcal L_{\mathrm{align}}(f)+\lambda
    \sup_{g\in\mathcal G}\left[\E_{\mathcal R\sim\P_{\mathcal R}}\{g(\mathcal R)\}
    -\E_{Z\sim\P_f}\{g(Z)\}\right].
\end{equation}
Here, for a fixed constant $0<C<\infty$,
\[
    \mathcal G=\{g\in\mathrm{Lip}(C):g(\bm0)=0\},
\]
and $\mathrm{Lip}(C)$ denotes the class of real-valued $C$-Lipschitz functions.
For computation, we parameterise the encoder and critic by neural networks
$f_{\bm\theta}$ and $g_{\bm\omega}$, respectively. In the image experiments,
$f_{\bm\theta}$ is a convolutional encoder and $g_{\bm\omega}$ is an MLP;
the architectures are detailed in Section~\ref{subsection: Benchmark Classification Performance on Real-World Image Datasets}. The encoder output is normalised to
radius $R$, enforcing the spherical range condition used in the population
analysis.

From each $X_S^{(i)}$ we form two views $X_{S,1}^{(i)}$ and $X_{S,2}^{(i)}$, and independently draw $\mathcal R^{(i)}\sim\P_{\mathcal R}$. Write $\widetilde{\mathcal D}_S=\{(X_{S,1}^{(i)},X_{S,2}^{(i)},\mathcal R^{(i)}):1\leq i\leq n_S\}$ for the augmented source dataset used in the empirical objective, including its independent reference draws. The empirical criterion is
\begin{gather}
    \widehat{\mathcal L}(\bm\theta,\bm\omega)=
    \widehat{\mathcal L}_{\mathrm{align}}(\bm\theta)+
    \lambda\widehat{\mathcal W}(\bm\theta,\bm\omega),\nonumber\\
    \widehat{\mathcal L}_{\mathrm{align}}(\bm\theta)=\frac1{n_S}\sum_{i=1}^{n_S}
    \norm{f_{\bm\theta}(X_{S,1}^{(i)})-f_{\bm\theta}(X_{S,2}^{(i)})}_2^2,\nonumber\\
    \widehat{\mathcal W}(\bm\theta,\bm\omega)=\frac1{n_S}\sum_{i=1}^{n_S}
    \left\{g_{\bm\omega}(\mathcal R^{(i)})-
    \frac{g_{\bm\omega}(f_{\bm\theta}(X_{S,1}^{(i)}))+
    g_{\bm\omega}(f_{\bm\theta}(X_{S,2}^{(i)}))}{2}\right\}.\label{eq: nn objective}
\end{gather}
Let $\Omega_{\mathcal F}$ and $\Omega_{\mathcal G}$ be the feasible parameter
spaces for the encoder and critic, respectively, inducing the hypothesis classes
\begin{equation*}
    \widehat{\mathcal F}:=\{f_{\bm\theta}:\bm\theta\in\Omega_{\mathcal F}\},
    \qquad
    \widehat{\mathcal G}:=\{g_{\bm\omega}:\bm\omega\in\Omega_{\mathcal G}\}.
\end{equation*}
The ideal empirical minimax estimator is defined by
\begin{equation}\label{eq: saddle point}
    (\hat{\bm\theta}_{n_S},\hat{\bm\omega}_{n_S})\in
    \argmin_{\bm\theta\in\Omega_{\mathcal F}}
    \max_{\bm\omega\in\Omega_{\mathcal G}}
    \widehat{\mathcal L}(\bm\theta,\bm\omega).
\end{equation}
In computation, the critic's hard Lipschitz constraint is relaxed using a gradient
penalty \citep{gulrajani2017wgangp}; implementation details are in Section~\ref{subsection: Benchmark Classification Performance on Real-World Image Datasets}.

\paragraph{Stochastic optimisation.}
Equation~\eqref{eq: nn objective} estimates the dual expectations directly from minibatches. The critic and encoder can therefore be trained by alternating stochastic-gradient updates, as detailed in Appendix~\ref{subsection: Benchmark Classification Performance on Real-World Image Datasets}. That appendix also provides a critic-free batch-wise acceleration, whose CIFAR-10 accuracy and training speed are reported in Table~\ref{tab: batchwise results}.

\subsection{Downstream inference}\label{subsection: downstream inference}

With the encoder fixed, the fitted affine linear probe is
\begin{equation}\label{eq: linear probe (gradient descent)}
    G_{\widehat{\bm W},\hat{\bm b},f_{\hat{\bm\theta}_{n_S}}}(\bm x)=
    \argmax_{1\leq k\leq K}\{\widehat{\bm W}f_{\hat{\bm\theta}_{n_S}}(\bm x)+\hat{\bm b}\}_k,
\end{equation}
where $\widehat{\bm W}\in\R^{K\times d^\star}$ and $\hat{\bm b}\in\R^K$
are obtained by minimising the empirical target loss:
\begin{equation}\label{eq: downstream erm}
    (\widehat{\bm W},\hat{\bm b})\in\argmin_{\bm W,\bm b}
    \frac1{n_T}\sum_{i=1}^{n_T}
    \ell\bigl(\bm Wf_{\hat{\bm\theta}_{n_S}}(X_T^{(i)})+\bm b,Y_i\bigr).
\end{equation}
Here $\ell:\R^K\times\{1,\ldots,K\}\to\R_+$ may be any
multiclass classification loss, such as cross-entropy or the multiclass hinge
loss. In the experiments, we use cross-entropy; implementation details are
given in Section~\ref{subsection: Benchmark Classification Performance on Real-World Image Datasets}.
The relation between $K'$ and $K$ is examined empirically in Section~\ref{subsection: finer-grained concept}; Remark~\ref{remark: overspecified theory} gives the corresponding theoretical condition.

\section{MAIN EMPIRICAL RESULTS}\label{section: experiment}

We evaluate the representations learned by \textit{Distribution Matching} (DM) in three ways: within-domain transfer from self-supervised pretraining to classification, sensitivity to the number $K'$ of reference components, and cross-dataset transfer. Section~\ref{subsection: Benchmark Classification Performance on Real-World Image Datasets} gives the data splits, architectures, optimisation settings and augmentation protocol.

\subsection{Transfer within a common input domain}

We first consider within-domain transfer from self-supervised pretraining to downstream classification. This is a standard benchmark protocol widely used in the SSL research community to evaluate learned representations \citep{chen2020simclr,zbontar2021barlow}. For each of CIFAR-10, CIFAR-100 \citep{Krizhevsky09cifar} and STL-10 \citep{coates2011stl10}, an encoder is pretrained on an unlabelled source sample and then frozen before a classifier is fitted on labelled target observations, following the official training and test splits; risk is evaluated on the held-out test set. Section~\ref{subsection: Benchmark Classification Performance on Real-World Image Datasets} specifies the training partitions used at each stage. For example, the CIFAR-100 columns in Table~\ref{tab:experiment_result} evaluate transfer from self-supervised pretraining to downstream classification within the same input domain. The number of target categories is $K=10$ for CIFAR-10 and STL-10 and $K=100$ for CIFAR-100. For the benchmark comparisons in Table~\ref{tab:experiment_result}, we set $K'=384$; Section~\ref{subsection: finer-grained concept} examines the role of this choice. Table~\ref{tab:experiment_result} compares DM with representative SSL methods using an affine linear probe and a $k$-NN probe.

DM gives the largest reported accuracy in all six columns. The improvements under both linear and $k$-NN probes show that the learned backbone features support both linear classification and classification using neighbouring observations. To examine the role of reference resolution, we next vary the number $K'$ of reference components.

\subsection{Resolution of the reference distribution}\label{subsection: finer-grained concept}

The role of $K'$ is most transparent when the downstream labels are semantically coarse. A single target category may contain several finer concepts, such as breed or coat colour within a broad ``dog'' category. Choosing $K'>K$ gives the unlabelled stage more reference components than there are target categories, allowing it to distinguish such structure without fine-grained labels. The components are not assumed to be named subclasses or to correspond one-to-one to the target categories. Instead, the labelled transfer stage learns how to aggregate the richer component structure into the $K$ target classes. Remark~\ref{remark: overspecified theory} provides a conditional theoretical analysis of fine-component recovery and aggregation.

Table~\ref{tab:ablation_study} reports the corresponding CIFAR-100 ablation, where $K=100$. As $K'$ increases from 32 to 384, linear-probe accuracy rises monotonically from 45.78\% to 67.71\%, while $k$-NN accuracy rises from 22.61\% to 56.18\%. The continued improvement after $K'$ exceeds $K$ is consistent with the interpretation that a finer reference preserves structure hidden by coarse target labels.

The structured reference also offers an additional level of interpretability. Uniformity encourages global dispersion \citep{wang2020understanding}, whereas our separated reference supplies explicit component-wise geometry for inspecting how a coarse category is subdivided. Figure~\ref{fig: spherical groups} shows normalised projected outputs of a saved CIFAR-10 DM model, grouped by their nearest signed coordinate direction. Within the same coarse category, different groups contain recurring differences in appearance and viewpoint: horse head-and-neck views versus whole horses, often with riders; frontal truck cabs versus box-shaped delivery bodies; and light, fluffy dog faces versus faces with longer hair and darker muzzles. These patterns also occur among randomly selected group members. They make the richer within-class geometry directly visible, complementing the quantitative $K'$ ablation. Section~\ref{subsection: qualitative spherical groups} gives the grouping and display protocol.

\begin{center}
\begin{minipage}{\linewidth}
    \centering
    \includegraphics[width=\linewidth]{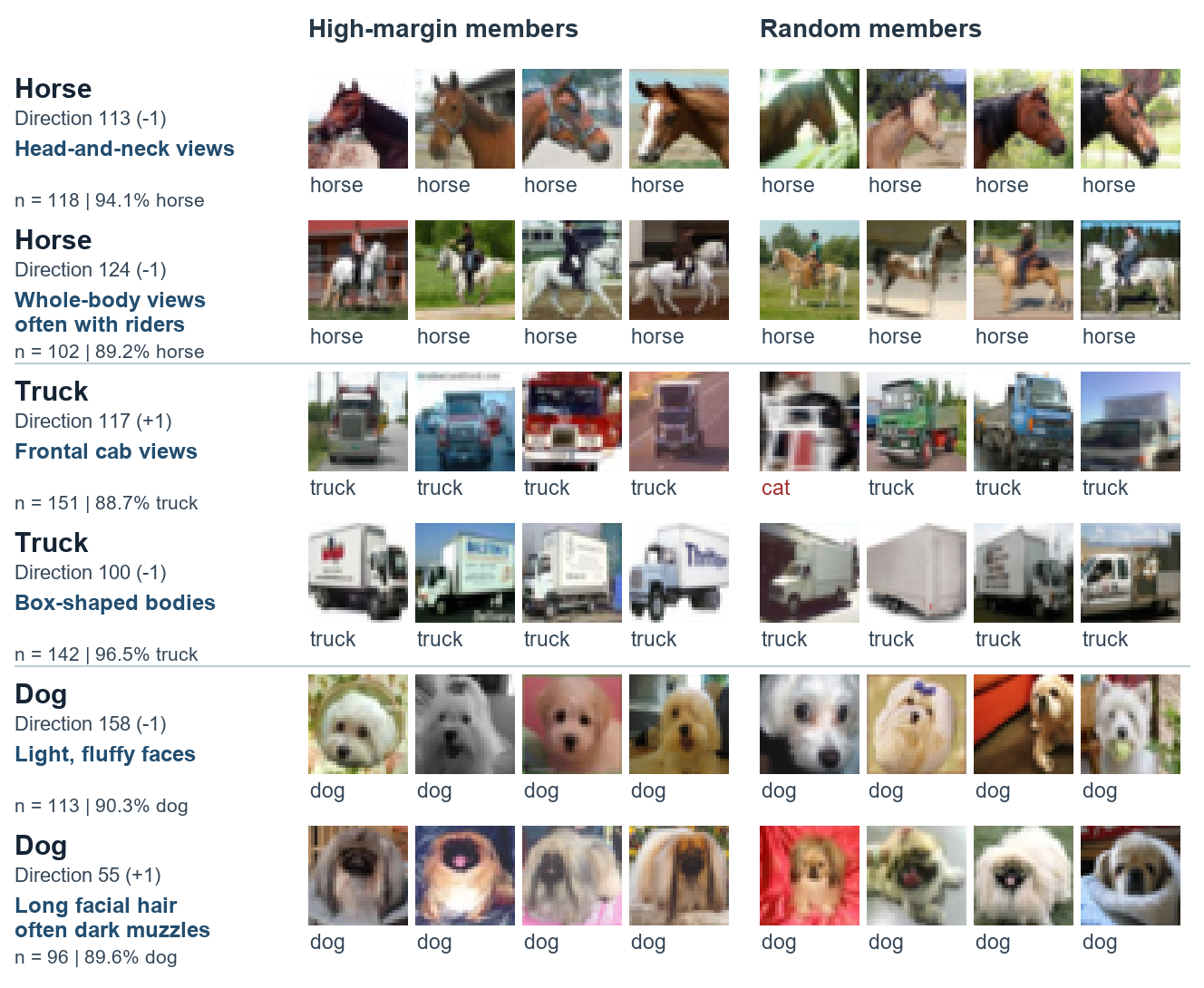}
    \captionof{figure}{\textbf{Qualitative organisation within coarse categories.} Rows show selected signed-coordinate groups of normalised projected CIFAR-10 test representations; descriptors summarise appearance and viewpoint patterns observed after grouping. Left: four high-margin members. Right: four random members, sampled after excluding each group's eight highest-margin members. Here $n$ is the full group size, and percentages give the coarse-class proportions in that group. Coarse-class labels are used only after assignment for summaries and display selection. Directions identify spherical coordinates, not verified original training-component indices; Section~\ref{subsection: qualitative spherical groups} gives the protocol.}
    \label{fig: spherical groups}
\end{minipage}
\end{center}

\subsection{Cross-dataset transfer}\label{subsection: transferability}

Finally, we pretrain $f_{\hat{\bm\theta}_{n_S}}$ without labels on CIFAR-100 ($n_S=50,000$), freeze the backbone, and fit probes to its features on CIFAR-10. This setting involves both domain shift and a change in label space. Table~\ref{tab:transfer_results} shows that the transferred representation supports both linear and $k$-NN classification on CIFAR-10, demonstrating cross-dataset transferability.

\section{THEORETICAL GUARANTEES}\label{section: theory}

We analyse the Mallows--Wasserstein instantiation $\mathcal D=\mathcal W$
of \eqref{eq: DM objective}.

For nonnegative quantities $a$ and $b$, we write $a\lesssim b$ if $a\leq Cb$, and $a\asymp b$ if $cb\leq a\leq Cb$, for constants $0<c\leq C<\infty$ that may depend on fixed model parameters but not on the sample sizes.

The population analysis relates the DM objective to target-centre separation and nearest-centroid classification error. The sample analysis then controls $\mathcal L(f_{\hat{\bm\theta}_{n_S}})$ for the empirical neural estimator. Proofs of the population and sample results are in Section~\ref{section: proof of theorem 1} and Appendix~\ref{section: Appendix B}, respectively.

\subsection{Target rule and structural shift}\label{subsection: target rule and structural shift}

For each labelled target observation, draw independent $A_1^{(i)},A_2^{(i)}\sim U_{\mathcal A}$ and set $X_{T,j}^{(i)}=A_j^{(i)}(X_T^{(i)})$ with $j=1,2$. Write $\widetilde{\mathcal D}_T=\{(X_T^{(i)},Y_i,X_{T,1}^{(i)},X_{T,2}^{(i)}):1\leq i\leq n_T\}$ for the augmented target dataset. With $n_T(k)=\sum_i\1\{Y_i=k\}>0$ for $k=1,\ldots,K$, define
\begin{equation}\label{eq: hat mu T}
    \widehat{\bm\mu}_T(k)=\frac{1}{2n_T(k)}\sum_{i=1}^{n_T}
    \{f(X_{T,1}^{(i)})+f(X_{T,2}^{(i)})\}\1\{Y_i=k\},
\end{equation}
and its population counterpart
\begin{equation}\label{eq: population mu T}
    \bm\mu_T(k)=\E_{(X_T,Y)\sim\P_T}\left[\E_{A\sim U_{\mathcal A}}
    \{f(A(X_T))\}\mid Y=k\right].
\end{equation}
In the sample-complexity analysis, we condition on the observed class counts $\{n_T(k)\}_{k=1}^K$, suppressing this conditioning from the notation.
The theoretical downstream rule is the nearest-centroid rule, written in affine form as
\begin{equation}\label{eq: linear probe}
    G_f(\bm x)=\argmax_{1\leq k\leq K}
    \left\{\widehat{\bm W}f(\bm x)+\hat{\bm b}\right\}_k,
    \qquad
    \widehat{\bm W}_{k\cdot}=2\widehat{\bm\mu}_T(k)^\top,
    \quad \hat b_k=-\norm{\widehat{\bm\mu}_T(k)}_2^2.
\end{equation}
This classifier has the same affine form as \eqref{eq: linear probe (gradient descent)}, with weights determined by the empirical class centres. Its risk is
\begin{equation}\label{eq: excess risk}
    \mathrm{Err}(G_f)=\P_T\{G_f(X_T)\neq Y\}.
\end{equation}

Fix tolerances $\epsilon_1,\epsilon_2\geq0$, and let $\{C_T(k)\}_{k=1}^K$ be the target-class partition, with $C_T(k)$ corresponding to $Y=k$. We assume that there exists a measurable partition $\{C_S(k)\}_{k=1}^K$ of $\mathcal X$ whose components correspond to the target classes in the following sense. Write $\P_T(k)$ for the law of $X_T$ conditional on $Y=k$ and $\P_S(k)$ for the law of $X_S$ conditional on $X_S\in C_S(k)$, and set $p_T(k)=\P_T(Y=k)$ and $p_S(k)=\P_S\{X_S\in C_S(k)\}$. The corresponding conditional laws and probabilities satisfy
\begin{equation}\label{eq: shift constraints}
    \max_{1\leq k\leq K}\mathcal W(\P_S(k),\P_T(k))\leq\epsilon_1,
    \qquad
    \max_{1\leq k\leq K}\abs{p_S(k)-p_T(k)}\leq\epsilon_2.
\end{equation}
We fix one such source partition throughout the analysis. For within-dataset SSL, a standard setting for assessing the quality of learned representations, the source and target input distributions coincide, so one may take $C_S(k)=C_T(k)$, giving $\epsilon_1=\epsilon_2=0$.

\begingroup
\clubpenalty=10000
\widowpenalty=10000
\subsection{Augmentation quality and local identifiability}\label{subsection: augmentation quality and local identifiability}

To characterise semantic similarity between observations, we define the augmentation distance as the smallest Euclidean distance between their augmented views:
\begin{equation}\label{def: augmentation distance}
    d_{\mathcal A}(\bm x_1,\bm x_2)=
    \min_{A_1,A_2\in\mathcal A}\norm{A_1(\bm x_1)-A_2(\bm x_2)}_2.
\end{equation}
\endgroup
We take the identity to belong to $\mathcal A$. Because the minimum in \eqref{def: augmentation distance} is taken over two transformed views, $d_{\mathcal A}$ compares augmentation orbits rather than raw observations.

\begin{definition}[$(\sigma,\delta)$-augmentation]\label{def: augmentation}
The family $\mathcal A$ is a $(\sigma,\delta)$-augmentation relative to the target partition if, for every $k$, there is $C_T^\circ(k)\subseteq C_T(k)$ such that
\begin{enumerate}[(a)]
    \item $\P_T\{X_T\in C_T^\circ(k)\}\geq\sigma\P_T\{X_T\in C_T(k)\}$;
    \item $\sup_{\bm x_1,\bm x_2\in C_T^\circ(k)}d_{\mathcal A}(\bm x_1,\bm x_2)\leq\delta$.
    \item $\P_T\{X_T\in\bigcup_{k=1}^K C_T^\circ(k)\}\geq\sigma$.
\end{enumerate}
Thus $\delta$ measures within-class semantic connectivity and $\sigma$ the globally covered probability mass. Section~\ref{section: augmentation distance interpretation} provides a detailed interpretation and geometric illustration.
\end{definition}

\begin{assumption}[Regularity of augmentations]\label{assumption: Lip augmentation}
There is $Q>0$ such that, for all $A\in\mathcal A$ and $\bm x_1,\bm x_2\in[0,1]^d$,
\begin{equation*}
    \norm{A(\bm x_1)-A(\bm x_2)}_2
    \leq Q\norm{\bm x_1-\bm x_2}_2.
\end{equation*}
\end{assumption}
Common augmentation operators used in self-supervised learning satisfy this regularity condition; explicit verification for standard augmentation families is given by \citet{duan2025advssl}.

We next state a quantitative identifiability condition for the optimal transport. For $K'=K$, recall the latent reference-component label $\mathcal C\in\{1,\ldots,K\}$, with $\P(\mathcal C=j)=\alpha_j$ and $\mathcal R\mid\mathcal C=j\sim\P_{\mathcal R}(j)$. Let $\Q^\star$ be a joint law of $(X_S,A,\mathcal C,\mathcal R)$ with $(X_S,A)\sim\P_S\otimes U_{\mathcal A}$ and $(\mathcal C,\mathcal R)$ distributed according to this reference-mixture construction, such that the induced pair $(f(A(X_S)),\mathcal R)$ attains $\mathcal W(\P_f,\P_{\mathcal R})$. Define
\begin{equation}\label{eq: optimal permutation}
    \tau^\star\in\argmax_{\tau\in P_K}\sum_{k=1}^K
    \Q^\star\{X_S\in C_S(k),\mathcal C=\tau(k)\},
\end{equation}
where $P_K$ is the set of permutations of $\{1,\ldots,K\}$.
Write $\P_{f,S}(k)$ for the law of $f(A(X_S))$ conditional on $X_S\in C_S(k)$, and $\P_{\mathcal R}(j)$ for the law of $\mathcal R$ conditional on $\mathcal C=j$.
For the matched pairs, set
\begin{equation*}
    q_k^\star
    :=\Q^\star\{X_S\in C_S(k),\mathcal C=\tau^\star(k)\},
    \qquad k=1,\ldots,K,
\end{equation*}
and define their maximal two-sided leakage by
\begin{equation}\label{eq: transport leakage}
    \mathrm{Leak}(\Q^\star)
    :=\max_{1\leq k\leq K}
    \big\{p_S(k)+\alpha_{\tau^\star(k)}-2q_k^\star\big\}.
\end{equation}
Here, $p_S(k)-q_k^\star$ is the mass of source component $k$ transported to non-matched reference components, whereas $\alpha_{\tau^\star(k)}-q_k^\star$ is the mass received by its matched reference component from other source components.

\begin{assumption}[Representational identifiability]\label{assumption: transport mass}
For a fixed constant $C_{\mathrm{id}}>0$, the optimal coupling $\Q^\star$ satisfies
\begin{equation*}
    q_k^\star>0,\quad k=1,\ldots,K,
    \qquad
    \mathrm{Leak}(\Q^\star)
    \leq C_{\mathrm{id}}\mathcal W(\P_f,\P_{\mathcal R}).
\end{equation*}
\end{assumption}

Assumption~\ref{assumption: transport mass} permits an optimal plan to split component mass, but bounds the off-matched mass by a fixed multiple of the overall matching discrepancy. It therefore rules out a representation in which different semantic components have the same mixture over the reference components even though the overall distributions match exactly. Exact component-wise transport is the special case $\mathrm{Leak}(\Q^\star)=0$. Proposition~\ref{proposition: leakage support gap} provides a geometric sufficient condition for the leakage bound.

\subsection{Population guarantee}\label{subsection: population guarantee}

\begin{theorem}\label{theorem: pop theorem}
Let $\mathcal A$ be a finite $(\sigma,\delta)$-augmentation equipped with the uniform law $U_{\mathcal A}$, and let $f\in\mathcal F$. Under Assumptions~\ref{assumption: Lip augmentation} and \ref{assumption: transport mass},
\begin{equation*}
    \max_{i\neq j}\abs{\bm\mu_T(i)^\top\bm\mu_T(j)}
    \lesssim \mathcal L(f)+\epsilon_1.
\end{equation*}
Furthermore, for every $\eps>0$, if
$\max_{i\neq j}\bm\mu_T(i)^\top\bm\mu_T(j)<R^2\psi(\sigma,\delta,\eps,f)$,
then
\begin{equation*}
    \mathrm{Err}(G_f)\leq(1-\sigma)+
    \mathcal O\!\left(\eps^{-1}\{\mathcal L(f)+\epsilon_1+\epsilon_2\}^{1/2}\right),
\end{equation*}
where $\psi(\sigma,\delta,\eps,f)$ is given in Lemma~\ref{lemma: sufficient condition of small Err} of Section~\ref{section: proof of theorem 1}.
\end{theorem}

The proof has three steps. First, for each matched source/reference pair, we split the component first moments into contributions from matched and off-matched mass. The Mallows cost controls the difference between the matched contributions, while the leakage condition in Assumption~\ref{assumption: transport mass} controls the remaining contributions; together they bound the distance between the source-component mean and its matched reference mean. The reference component means lie in orthogonal directions, so these bounds control cross-class inner products after the source-to-target shift in \eqref{eq: shift constraints}. Secondly, $(\sigma,\delta)$-augmentation and $\mathcal L_{\mathrm{align}}(f)$ control the probability that two views of a target observation are far apart. Finally, on this stable set, the separation condition makes the nearest-centroid score of the correct class larger than every competing score. The uncovered mass contributes $1-\sigma$.

\begin{remark}[What changes when $K'>K$]\label{remark: overspecified theory}
Theorem~\ref{theorem: pop theorem} treats $K'=K$, where each target class corresponds to one reference component and \eqref{eq: hat mu T} estimates its centre. For $K'>K$, suppose the target population admits an unobserved fine-component index $J\in\{1,\ldots,K'\}$ and a surjective aggregation map
$a^\star:\{1,\ldots,K'\}\to\{1,\ldots,K\}$ such that $Y=a^\star(J)$. Suppose further that the latent fine-label problem obtained by using $J$ as the target label admits a corresponding $K'$-component source partition and satisfies the structural-shift, augmentation and representational-identifiability conditions of Theorem~\ref{theorem: pop theorem}, with $J$ and $K'$ replacing $Y$ and $K$. We retain $\epsilon_1$ and $\epsilon_2$ for the resulting fine-level shift tolerances. Define the fine-component centres and the corresponding population centroid head by
\begin{equation}\label{eq: fine component centres}
\begin{aligned}
    \bm\nu_T(j)
    &:=\E\!\left[\E_{A\sim U_{\mathcal A}}\{f(A(X_T))\}\mid J=j\right],\\
    \bm W_{j\cdot}^\star&:=2\bm\nu_T(j)^\top,
    \qquad b_j^\star:=-\norm{\bm\nu_T(j)}_2^2,
    \qquad j=1,\ldots,K'.
\end{aligned}
\end{equation}
Let
\begin{equation}\label{eq: fine oracle margin}
\begin{aligned}
    J_f^\star(\bm x)
    &\in\argmax_{1\leq j\leq K'}
    \{\bm W^\star f(\bm x)+\bm b^\star\}_j,
    &G_f^\star(\bm x)&:=a^\star(J_f^\star(\bm x)),\\
    m_f^\star(\bm x)
    &:=\{\bm W^\star f(\bm x)+\bm b^\star\}_{J_f^\star(\bm x)}
    -\max_{l\neq J_f^\star(\bm x)}
    \{\bm W^\star f(\bm x)+\bm b^\star\}_l.
\end{aligned}
\end{equation}
Ties are resolved by a fixed deterministic rule. When $K'=K$ and $a^\star$ is bijective, relabelling gives $J=Y$ and $\bm\nu_T(k)=\bm\mu_T(k)$. Thus \eqref{eq: fine oracle margin} has the same nearest-centroid form as \eqref{eq: linear probe}; taking $\widehat{\bm\nu}_T(k)=\widehat{\bm\mu}_T(k)$ below recovers $G_f$ exactly.

Under these fine-level conditions, the population-centroid version of
Theorem~\ref{theorem: pop theorem} controls
$\P_T\{J_f^\star(X_T)\neq J\}$ whenever its corresponding separation condition
holds. Moreover,
$\mathrm{Err}(G_f^\star)\leq\P_T\{J_f^\star(X_T)\neq J\}$ because a wrong fine
prediction need not change the aggregated target class. Thus
$\mathrm{Err}(G_f^\star)$ is controlled by the same population bound as the
fine-component error.

For $K'>K$, however, the coarse label reveals $a^\star(J)$ but not $J$. Thus neither the $K'$ fine centres nor their many-to-one aggregation are directly estimable by the construction in \eqref{eq: hat mu T}. To isolate this additional recoverability problem, suppose an algorithm $\mathsf A$, using the coarsely labelled target sample and the fixed encoder, returns centres $\{\widehat{\bm\nu}_T(j)\}_{j=1}^{K'}$ in the radius-$R$ ball and a map $\widehat a$. Let $P_{K'}$ denote the set of permutations of $\{1,\ldots,K'\}$. Assume that, with probability at least $1-\zeta_{n_T}$, there is a permutation $\pi\in P_{K'}$ such that
\begin{equation}\label{eq: aggregation algorithm error}
    \max_{1\leq j\leq K'}
    \norm{\widehat{\bm\nu}_T(\pi(j))-\bm\nu_T(j)}_2
    \leq\eta_{n_T},
    \qquad
    \widehat a(\pi(j))=a^\star(j),\quad j=1,\ldots,K'.
\end{equation}
Set $\widehat{\bm W}_{j\cdot}=2\widehat{\bm\nu}_T(j)^\top$ and
$\hat b_j=-\norm{\widehat{\bm\nu}_T(j)}_2^2$. Define
$\widehat J(\bm x)\in\argmax_{1\leq j\leq K'}
\{\widehat{\bm W}f(\bm x)+\hat{\bm b}\}_j$ and
$\widehat G(\bm x)=\widehat a(\widehat J(\bm x))$. Then, averaging over the
sample used by $\mathsf A$,
\begin{equation}\label{eq: overspecified risk}
    \E_{\widetilde{\mathcal D}_T}\{\mathrm{Err}(\widehat G)\}
    \leq\mathrm{Err}(G_f^\star)
    +\P_T\!\left\{m_f^\star(X_T)\leq8R\eta_{n_T}\right\}
    +\zeta_{n_T}.
\end{equation}
Combining the fine-level population bound with \eqref{eq: overspecified risk}
gives, whenever the corresponding fine-level separation condition holds,
\begin{equation*}
\begin{aligned}
    \E_{\widetilde{\mathcal D}_T}\{\mathrm{Err}(\widehat G)\}
    \leq{}&(1-\sigma)
    +\mathcal O\!\left(\eps^{-1}
    \{\mathcal L(f)+\epsilon_1+\epsilon_2\}^{1/2}\right)\\
    &+\P_T\!\left\{m_f^\star(X_T)\leq8R\eta_{n_T}\right\}
    +\zeta_{n_T}.
\end{aligned}
\end{equation*}
This is the two-stage extension to $K'>K$: DM first separates the latent fine
components, and algorithm $\mathsf A$ then recovers their centres and
aggregation from coarse labels, up to the stated recovery and margin errors.
Section~\ref{section: fine centre aggregation} gives the details.
\end{remark}

\subsection{Non-asymptotic sample complexity}\label{subsection: sample complexity}

Theorem~\ref{theorem: pop theorem} is a population implication for a given
encoder $f$ under the structural Assumptions~\ref{assumption: Lip augmentation}
and \ref{assumption: transport mass}. The remaining statistical question is
whether empirical risk minimisation, implemented here through a saddle-point
estimator, can attain sufficiently small population DM risk and thereby satisfy
the separation condition of Theorem~\ref{theorem: pop theorem} with high
probability. We now address this question by controlling the estimation and
approximation errors of the empirical objective.

For the sample-complexity analysis, let $\rho(t)=\max(t,0)$ act coordinatewise.
The class $\mathcal{NN}_{p,q}(W,D,B)$ consists of functions $h:\R^p\to\R^q$
of the form
\begin{equation}\label{eq: neural network class}
\begin{aligned}
    \bm z_0(\bm x)&=\bm x,\\
    \bm z_l(\bm x)&=\rho\!\left(\bm A_l\bm z_{l-1}(\bm x)+\bm b_l\right),
        \qquad l=1,\ldots,D,\\
    h(\bm x)&=\bm A_{D+1}\bm z_D(\bm x)+\bm b_{D+1},
\end{aligned}
\end{equation}
where $m_0=p$, $m_{D+1}=q$, $\bm A_l\in\R^{m_l\times m_{l-1}}$,
$\bm b_l\in\R^{m_l}$, and $\max_{1\leq l\leq D}m_l\leq W$. Thus $D$ is
the number of hidden affine--ReLU layers and $W$ is the maximum hidden-layer
width. On a specified compact domain $\mathcal X_p\subseteq\R^p$, we further impose
$\sup_{\bm x\in\mathcal X_p}\norm{h(\bm x)}_\infty\leq B$. Let
$\Delta_{n_S}$ be a deterministic upper bound for the ReLU approximation
error with
$\Delta_{n_S}=\widetilde{\mathcal O}\{(D_1W_1)^{-2/d}\}$, and set
\begin{equation*}
    R_{1,n_S}=R-\Delta_{n_S},
    \qquad R_{2,n_S}=R+\Delta_{n_S}.
\end{equation*}
For all sufficiently large $n_S$, $R_{1,n_S}>0$. We use the annular encoder
sieve
\begin{equation*}
\widehat{\mathcal F}
=\left\{f\in\mathcal{NN}_{d,d^\star}(W_1,D_1,B_1)\cap\mathrm{Lip}(\bar L):
R_{1,n_S}\leq\norm{f(\bm x)}_2\leq R_{2,n_S}
\text{ on }\mathcal X\right\},
\end{equation*}
where $\bar L$ is a sufficiently large fixed constant. The annular range
constraint accommodates the uniform ReLU approximation error $\Delta_{n_S}$.
The critic domain is
$\mathcal X_{d^\star}=\overline B(\bm0,R_{2,n_S})$. We take a symmetric,
anchored ReLU sieve
$\widehat{\mathcal G}\subseteq
\mathcal{NN}_{d^\star,1}(W_2,D_2,B_2)\cap\mathcal G$ satisfying
\[
\sup_{g\in\mathcal G}\inf_{\hat g\in\widehat{\mathcal G}}
\norm{g-\hat g}_\infty
=\widetilde{\mathcal O}\{(D_2W_2)^{-2/d^\star}\}.
\]
The critic class $\mathcal G$ is defined in Section~\ref{subsection: estimation and computation}. Section~\ref{section: The Approximation Error} gives the
sieve approximation details. Thus
$(W_1,D_1)$ and $(W_2,D_2)$ describe the widths and depths of the encoder and
critic sieves, respectively; $B_1$ and $B_2$ are fixed independently of
$n_S$.

For brevity, write
\begin{equation}\label{eq: empirical encoder notation}
    \hat f_{n_S}:=f_{\hat{\bm\theta}_{n_S}},
    \qquad
    \hat g_{n_S}:=g_{\hat{\bm\omega}_{n_S}},
\end{equation}
for the empirical saddle-point pair in \eqref{eq: saddle point}. The encoder
satisfies the basic error decomposition
\begin{equation}\label{eq: main error decomposition}
    \mathcal L(\hat f_{n_S})\leq\mathcal L(f^\star)+
    2\mathcal E_{\mathrm{sta}}+\mathcal E_{\mathcal F}+2\mathcal E_{\mathcal G},
\end{equation}
where $\mathcal E_{\mathrm{sta}}$ is the uniform empirical-process error and $\mathcal E_{\mathcal F},\mathcal E_{\mathcal G}$ are the encoder and critic approximation errors; precise definitions are in Section~\ref{subsection: error decomposition}. An approximate saddle-point solution adds its optimisation error to the right-hand side.

\begin{assumption}[Distribution shift scale]\label{assumption: rates}
For the given $n_S>1$, suppose $\epsilon_1\lesssim n_S^{-\alpha}$ and $\epsilon_2\lesssim n_S^{-\beta}$ for some $\alpha,\beta>0$, with implicit constants independent of $n_S$.
\end{assumption}

\begin{assumption}[Augmentation refinement]\label{assumption: existence of augmentation}
There is a sequence of finite $(\sigma_n,\delta_n)$-augmentations $\mathcal A_n$, each equipped with the uniform law and satisfying $\sup_n\abs{\mathcal A_n}<\infty$, such that
$\sigma_n\to1$ and $\delta_n\to0$.
\end{assumption}

\begin{assumption}[Realizable transport geometry]\label{ass: realizable}
There exists an augmentation-invariant map $\phi\in\mathcal F$ such that:
\begin{enumerate}[(a)]
    \item $\phi(A(\bm x))=\phi(\bm x)$ for every $\bm x\in\mathcal X$
    and $A\in\mathcal A$;
    \item
    $\displaystyle\min_{i\neq j}\operatorname{dist}
    (\phi(C_S(i)),\phi(C_S(j)))\geq\rho>0$.
\end{enumerate}
Writing $\nu_k:=\phi_\sharp\P_S(k)$, for each $k=1,\ldots,K$ there exists
a Lipschitz map
\begin{equation*}
    T_k:\phi(C_S(k))\longrightarrow
    \operatorname{supp}(\P_{\mathcal R}(k))
    \quad\text{such that}\quad
    (T_k)_\sharp\nu_k=\P_{\mathcal R}(k).
\end{equation*}
The maps $T_k$ need not be optimal transport maps.
\end{assumption}

For the realizability analysis, relabel the reference components, if necessary,
so that component $k$ corresponds to $C_S(k)$, choose the design weights
$\alpha_k=p_S(k)$, and take the fixed constant $L$ in
\eqref{eq: functional class} no smaller than the Lipschitz constant of the
resulting composite map. Assumption~\ref{ass: realizable} is then sufficient
for an $f^\star\in\mathcal F$ with $\mathcal L(f^\star)=0$: the separated sets
allow the maps $T_k$ to be glued without losing Lipschitz continuity, while
the chosen weights make the component-wise push-forward equal to
$\P_{\mathcal R}$. For $\mathcal A=\mathcal A_{n_S}$, the maps $\phi$ and $T_k$ may depend on $n_S$, but their composite must remain in the same fixed class $\mathcal F$.

\begin{theorem}[End-to-End Risk Bound]\label{theorem: sample theorem}
Fix a sufficiently large sample size $n_S$. Suppose Assumption~\ref{assumption: Lip augmentation} and Assumptions~\ref{assumption: rates}--\ref{ass: realizable} hold. Suppose also that Assumption~\ref{assumption: transport mass} holds uniformly, with a fixed $C_{\mathrm{id}}$, for encoders in $\widehat{\mathcal F}$ with sufficiently small population DM risk. If $d^\star\leq d$, $D_2\lesssim D_1$, $W_2\lesssim W_1$, and the encoder and critic sizes satisfy
\begin{equation*}
    D_2W_2\asymp D_1W_1\asymp n_S^{\frac{d}{2d+4}}.
\end{equation*}
In the regime $K'=K$, for a fixed sufficiently small $\eps>0$ and $\mathcal A=\mathcal A_{n_S}$, the following bound holds:
\begin{equation*}
    \E_{\widetilde{\mathcal D}_S,\widetilde{\mathcal D}_T}
    \{\mathrm{Err}(G_{\hat f_{n_S}})\}
    \leq(1-\sigma_{n_S})+
    \widetilde{\mathcal O}\!\left(n_S^{-\min\{\frac{1}{2d+4},\frac{\alpha}{2},\frac{\beta}{2}\}}\right)
    +\mathcal O\!\left(\frac1{\min_k\sqrt{n_T(k)}}\right).
\end{equation*}
\end{theorem}

The local identifiability condition is stated precisely at the \hyperref[proof: sample theorem]{start of the proof} in Section~\ref{section: Appendix B}. Here $\widetilde{\mathcal O}$ suppresses logarithmic factors. The entropy calculation in Section~\ref{subsection: The Stochastic Error} gives
$\E\mathcal E_{\mathrm{sta}}\lesssim D_1W_1/\sqrt{n_S}$ and
$\mathcal E_{\mathcal F}+\mathcal E_{\mathcal G}\lesssim(D_1W_1)^{-2/d}$, up to logarithmic factors. Balancing these terms yields
$\E\mathcal L(\hat f_{n_S})\lesssim n_S^{-1/(d+2)}$; Assumption~\ref{ass: realizable} removes the population residual. Applying the annular population guarantee in Theorem~\ref{theorem: general version of the pop theorem} and using concentration of \eqref{eq: hat mu T} then give the displayed classification risk bound. Thus the dimension-dependent cost is paid using the unlabelled sample $n_S$, whereas estimation of the downstream rule contributes an inverse-square-root term in the target class counts $n_T(k)$. Remark~\ref{remark: overspecified theory} gives the corresponding conditional extension for a fixed encoder when $K'>K$, including the fine-centre recoverability and margin terms in \eqref{eq: overspecified risk}.

\section{SIMULATION STUDIES}\label{section: simulation}

The simulations examine two implications of the formulation: whether distribution matching can rectify a non-linearly warped representation, and how performance changes with the unlabelled sample size. The reported runs use seed 42; Sections~\ref{subsection: experimental details simulation} and \ref{subsection: experimental details tsne} give the full data-generating and optimisation settings.

\subsection{Non-linear rectification}

We generate $K=10$ latent classes in $\R^{d^\star}$, with $d^\star=32$. Their centres $\{\bm c_k\}_{k=1}^{10}$ are the first ten standard basis vectors, and
\begin{equation*}
    \bm z_i=\bm c_{y_i}+\bm\epsilon_i,
    \qquad \bm\epsilon_i\sim\mathcal N(\bm0,0.4^2\mathbf I_{32}).
\end{equation*}
A fixed, randomly initialised three-layer map $G:\R^{32}\to\R^{512}$ with $\tanh$ activations produces $\bm x_i=G(\bm z_i)$. Positive pairs are obtained by adding independent $\mathcal N(\bm0,0.05^2\mathbf I_{32})$ perturbations before applying the same map $G$. The encoder $f:\R^{512}\to\R^{32}$ is trained from these unlabelled pairs using the DM objective \eqref{eq: DM objective}.

Table~\ref{tab:sim_results} compares multiclass logistic-regression classifiers fitted separately to the observed $X$, the latent $Z$, and the learned $f(X)$, using $n_T=200$ labelled observations. The latent oracle uses the same type of classifier, trained directly on the true latent representations $Z$ by minimising cross-entropy. DM materially improves on raw-space classification and essentially matches the latent-oracle error in this run: a linear classifier fitted to $f(X)$ predicts the labels about as accurately as one fitted directly to $Z$.

\subsection{Dependence on the unlabelled sample size}\label{subsection: rate_sim}

\begin{figure}[t]
    \centering
    \includegraphics[width=0.6\linewidth]{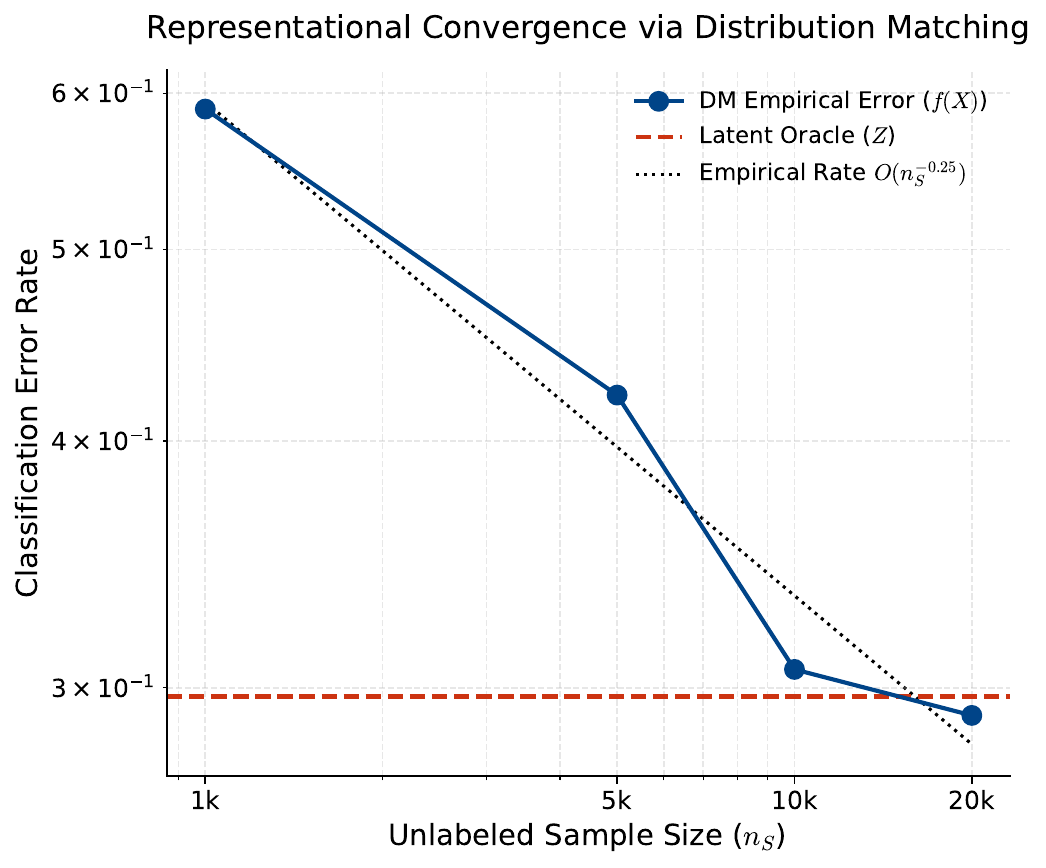}
    \caption{\textbf{Error versus unlabelled sample size.} The DM error decreases over the four values of $n_S$; the latent-oracle error is shown as a reference level.}
    \par\smallskip\noindent\footnotesize\textit{Alt text:} A log--log plot shows the DM classification error decreasing from 0.5890 at $n_S=1,000$ to 0.2905 at $n_S=20,000$, approaching the latent-oracle reference error of 0.2970.
    \label{fig:log_log}
\end{figure}

We next vary $n_S\in\{1,000,5,000,10,000,20,000\}$ while holding the labelled evaluation design fixed and setting the number of training iterations to $T=4000+n_S$. Table~\ref{tab:simulation_convergence} and Figure~\ref{fig:log_log} show a clear decline in error as the unlabelled pool and the training budget grow together, illustrating the combined benefit of additional unlabelled observations and optimisation.

\subsection{Two-dimensional visualisation}\label{subsection: sim_tsne}

\begin{figure}[t]
    \centering
    \includegraphics[width=0.9\linewidth]{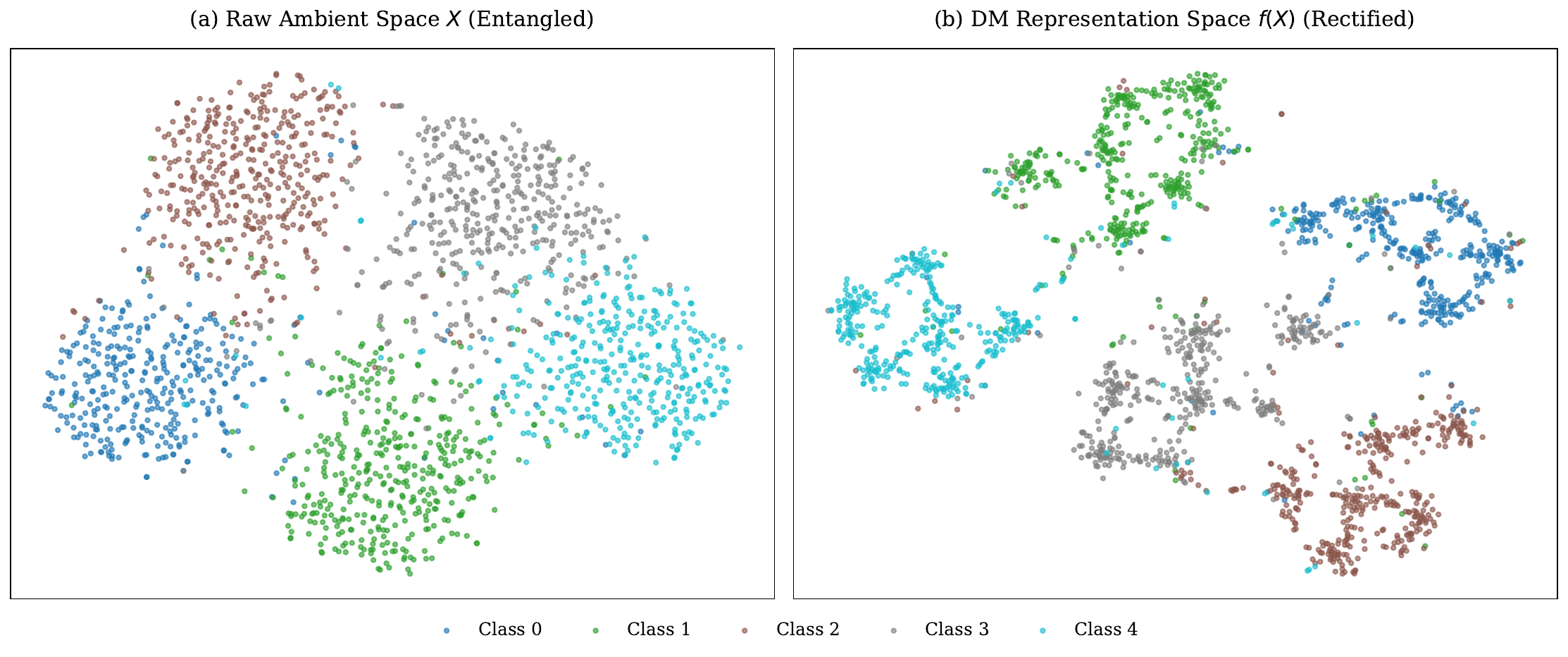}
    \caption{\textbf{t-SNE visualisation.} (a) Standardised observations $X$; (b) normalised DM representations $f(X)$. Colours denote latent classes and are not used in pretraining.}
    \par\smallskip\noindent\footnotesize\textit{Alt text:} Two t-SNE panels compare the same five coloured latent classes. In the standardised ambient space the classes overlap substantially; in the DM representation they form five more compact and separated groups.
    \label{fig:sim_tsne}
\end{figure}

Figure~\ref{fig:sim_tsne} gives a t-SNE visualisation \citep{hinton2008tsne} from a separate five-class simulation, with its configuration specified in Section~\ref{subsection: experimental details tsne}. The raw observations overlap strongly in this projection, whereas the DM representations form more compact and separated groups. Labels are used only for colouring, so the figure displays the geometric structure learned without label supervision.

\section{DISCUSSION}\label{section: outlook}

The principal contribution of DM is a change in statistical formulation. Conventional SSL losses mainly defend against collapse by imposing constraints that exclude undesirable representations; DM instead asks the encoder to solve a distribution matching problem and writes down the desired representation law explicitly. Augmentation alignment keeps the views of each observation close in representation space, while matching the separated reference distribution used here encourages representations to form distinct clusters. Unlabelled pretraining can distinguish fine-grained semantic concepts, while a labelled target sample learns how to aggregate them into the target classes.

We establish population and sample-complexity guarantees for $K'=K$. For $K'>K$, we reduce the population analysis to matching $K'$ fine semantic components with $K'$ reference components, followed by aggregation into $K$ target classes, under the recoverability and margin conditions in Remark~\ref{remark: overspecified theory}. In practice, the downstream classification head is trained on labelled target data by gradient-based optimisation.

At the framework level, DM takes the reference-directed matching goal
$f_\sharp\P_{\mathcal A}\rightsquigarrow\P_{\mathcal R}$ as its starting
point. A distributional discrepancy is one way to pursue this goal, as in
the Mallows-distance algorithm and theory developed here. Alternative
divergences, such as KL or Jensen--Shannon divergence when their density and
support requirements are appropriate, may produce different representations
and estimators. Subsequent work by \citet{mikulasch2026latent} gives concrete
support to this framework-level view: its KL-based latent distribution
matching formulation recovers several existing SSL objectives and yields a
new predictive model. That work matches a joint latent law to a
probabilistic latent model, whereas our focus is an explicit marginal
reference geometry with guarantees for downstream transfer.

The DM framework also yields a broader conceptual benefit: it places generative learning and representation learning in a common distributional language, under which they can be viewed as directionally inverse processes. Generative learning transports a tractable reference law towards the data law, whereas representation learning under DM transports the data law towards a prescribed latent law while enforcing augmentation invariance. This shared distributional viewpoint provides a principled route for transferring tools widely used in generative learning to representation learning.

Inspired by DM's view of generative learning and representation learning as
directionally inverse distributional processes, Flow-Based Distribution
Matching (FBDM) develops a new representation-learning algorithm by
transferring flow matching \citep{lipman2023flowmatching}, a widely used generative learning framework.
It achieves competitive performance with state-of-the-art SSL methods across
multiple benchmarks ranging from CIFAR to ImageNet \citep{jiao2026fbdm}.

Two further directions appear especially useful. First, the reference law could be selected or adapted to known invariances, hierarchical labels or scientific structure rather than fixed a priori. Secondly, the $(\sigma,\delta)$-augmentation condition suggests treating augmentation design as part of the statistical model: augmentation design should preserve semantic meaning and allow semantically related observations to have nearby augmented views. These questions are natural precisely because distribution matching exposes the target geometry instead of leaving it implicit in an anti-collapse mechanism.

\section{Data availability}
The benchmark datasets analysed in this article---CIFAR-10, CIFAR-100 and STL-10---are publicly available from their cited sources. The code used for model fitting and evaluation is available at \url{https://github.com/vincen-github/DM}. The simulation designs are specified in Section~\ref{section: simulation}, with complete settings in Sections~\ref{subsection: experimental details simulation} and \ref{subsection: experimental details tsne}.







\begin{samepage}
\section*{Acknowledgements}
\paragraph{Use of artificial intelligence.}
Google Gemini and OpenAI Codex provided important assistance with manuscript
revision, proof checking and constructive suggestions. The authors developed
the DM formulation, algorithm, original experimental implementations and proof
framework. Codex removed redundant encoder forward passes to improve efficiency.
It identified an invalid component-wise transport-cost decomposition involving
$\Q^\star$ and helped formulate an inter-component transport-leakage bound,
correct the inequalities and develop a sufficient condition for that bound.
Under the authors' direction, Codex
implemented post-training analysis code for checkpoint loading, extraction of
normalised projected representations, signed-coordinate grouping, selection
of high-margin members, fixed-seed random sampling and image-panel generation.
It assisted with visual inspection, group-wise descriptions of recurring
appearance and viewpoint patterns, exposition, notation consistency,
typesetting and cross-reference checks.
The authors reviewed all AI-assisted changes and take full
responsibility for the final mathematical arguments, implementation and
manuscript.
\end{samepage}

\bibliographystyle{abbrvnat}
\bibliography{reference}

\clearpage
\section*{Tables}

\begin{table}[htbp]
    \centering
    \fontsize{10}{12}\selectfont
    \renewcommand{\arraystretch}{1.1}
    \setlength{\tabcolsep}{4.5pt}
    \begin{tabular}{lcccccc}
        \toprule
        \multirow{2}{*}{Method} & \multicolumn{2}{c}{CIFAR-10} & \multicolumn{2}{c}{CIFAR-100} & \multicolumn{2}{c}{STL-10} \\
        \cmidrule(lr){2-3} \cmidrule(lr){4-5} \cmidrule(lr){6-7}
        & Linear & $k$-NN & Linear & $k$-NN & Linear & $k$-NN \\
        \midrule
        Barlow Twins \citep{zbontar2021barlow} & 87.32 & 84.74 & 55.88 & 46.41 & 81.41 & 76.41 \\
        SimCLR \citep{chen2020simclr} & 91.23 & 88.57 & 65.16 & 56.05 & 89.44 & 84.68 \\
        HaoChen et al. \citep{haochen2022beyond} & 86.95 & 82.04 & 56.48 & 48.62 & 81.54 & 77.31 \\
        VICReg \citep{adrien2022vicreg} & 90.16 & 86.10 & 61.63 & 52.59 & 88.63 & 83.13 \\
        \midrule
        \textbf{DM (ours)} & \textbf{92.11} & \textbf{89.17} & \textbf{67.71} & \textbf{56.18} & \textbf{90.22} & \textbf{85.51} \\
        \bottomrule
    \end{tabular}
    \caption{\textbf{Within-domain classification accuracy (\%).} For each benchmark, the encoder is pretrained without labels and its features are evaluated on a downstream classification task within the same input domain, using an affine linear probe and a $k$-NN probe. DM uses $K'=384$ reference components. The reported DM accuracies are higher than those of the listed baselines in each of the six comparisons.}
    \label{tab:experiment_result}
    \vspace{1.5em}
    
    \begin{minipage}[t]{0.49\textwidth}
    \centering
    \fontsize{10}{12}\selectfont
    \renewcommand{\arraystretch}{1.1}
    \setlength{\tabcolsep}{3.5pt}
    \begin{tabular}{lccccc}
        \toprule
        $K'$ & 32 & 64 & 128 & 256 & 384 \\
        \midrule
        Linear (\%) & 45.78 & 49.83 & 55.93 & 62.13 & \textbf{67.71} \\
        $k$-NN (\%) & 22.61 & 32.73 & 44.17 & 52.00 & \textbf{56.18} \\
        \bottomrule
    \end{tabular}
    \captionof{table}{\textbf{Sensitivity to the number of reference components.} CIFAR-100 accuracy increases over the examined range from $K'=32$ to $K'=384$.}
    \label{tab:ablation_study}
    \end{minipage}
    \hfill
    \begin{minipage}[t]{0.47\textwidth}
    \centering
    \fontsize{10}{12}\selectfont
    \renewcommand{\arraystretch}{1.1}
    \begin{tabular}{lcc}
        \toprule
        Transfer task & Linear & $k$-NN \\
        \midrule
        CIFAR-100 $\rightarrow$ CIFAR-10 & 80.64 & 71.47 \\
        \bottomrule
    \end{tabular}
    \captionof{table}{\textbf{Cross-dataset transfer accuracy (\%).} The encoder is pretrained without labels on CIFAR-100 and evaluated using probes fitted on CIFAR-10 ($k=5$).}
    \label{tab:transfer_results}
    \end{minipage}
\end{table}

\begin{table}[htbp]
    \centering
    \fontsize{10}{12}\selectfont
    \renewcommand{\arraystretch}{1.2}
    \begin{tabular}{lccc}
        \toprule
        Feature space & Raw $X$ & Latent oracle $Z$ & \textbf{DM $f(X)$} \\
        \midrule
        Classification error & 0.4160 & 0.3320 & \textbf{0.3310} \\
        \bottomrule
    \end{tabular}
    \caption{\textbf{Classification error after non-linear warping.} Here $d=512$, $n_S=15,000$ and $n_T=200$.}
    \label{tab:sim_results}
\end{table}

\begin{table}[htbp]
    \centering
    \fontsize{10}{12}\selectfont
    \renewcommand{\arraystretch}{1.1}
    \begin{tabular}{lcc}
        \toprule
        Unlabelled size ($n_S$) & Latent oracle $Z$ & \textbf{DM $f(X)$} \\
        \midrule
        $1,000$  & 0.2970 & 0.5890 \\
        $5,000$  & 0.2970 & 0.4220 \\
        $10,000$ & 0.2970 & 0.3065 \\
        $20,000$ & 0.2970 & \textbf{0.2905} \\
        \bottomrule
    \end{tabular}
    \caption{\textbf{Empirical error as $n_S$ increases.} The same target and test designs are used at every source sample size.}
    \label{tab:simulation_convergence}
\end{table}

\clearpage
\begin{appendices}

\section{Population guarantee}\label{section: Appendix A}

The population analysis establishes a fundamental link between the geometric structure induced by the \textit{distribution matching} task and the resulting downstream classification performance. This section generalizes the proof techniques developed in \citet{huang2023towards} and \citet{duan2025advssl} to the distribution-matching objective considered here.

\subsection{Interpretation of the augmentation distance}\label{section: augmentation distance interpretation}

\begin{figure}[htbp]
    \centering
    \includegraphics[width=0.62\linewidth]{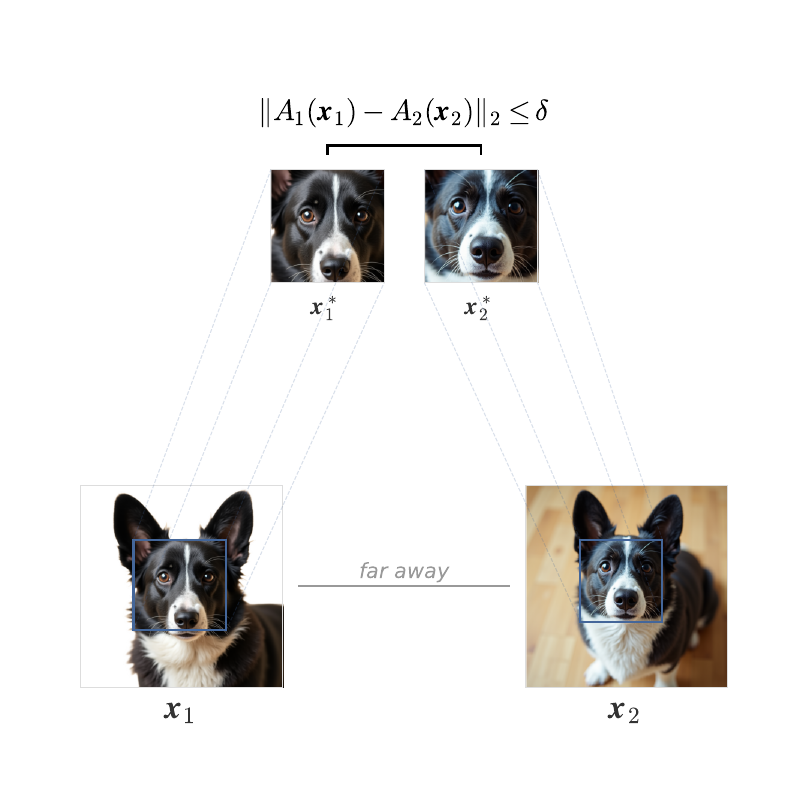}
    \caption{\textbf{Augmentation-induced semantic bridging.} The raw observations $\bm x_1$ and $\bm x_2$ may be far apart, while transformations $A_1,A_2\in\mathcal A$ isolate views $\bm x_1^\star$ and $\bm x_2^\star$ whose Euclidean distance is at most $\delta$. The augmentation distance therefore discounts nuisance variation that can be removed by the allowed transformations.}
    \label{fig: augmentation semantic bridge}
\end{figure}

Equation~\eqref{def: augmentation distance} uses augmentation as a mechanism for removing nuisance variation. Two observations may be far apart under raw Euclidean distance even when they share the same semantic content. If transformations $A_1,A_2\in\mathcal A$ isolate views that retain this shared structure while suppressing nuisance features such as backgrounds or irrelevant image regions, then $d_{\mathcal A}(\bm x_1,\bm x_2)$ is small. Thus $d_{\mathcal A}$ measures proximity between the augmentation orbits of two observations rather than proximity between the raw observations themselves. It is not required to be a metric: distinct observations may have zero augmentation distance when suitable transformed views coincide. Figure~\ref{fig: augmentation semantic bridge} illustrates this semantic-bridging interpretation.

The two parameters in the $(\sigma,\delta)$-augmentation condition separate resolution from coverage. The bound by $\delta$ requires semantic bridging only within the high-probability subsets $C_T^\circ(k)$, and smaller $\delta$ means that the augmentation family connects same-class observations more precisely. The global coverage condition ensures that these subsets contain at least probability $\sigma$; the remaining mass contributes the term $1-\sigma$ in the classification-risk bound. When the target classes form a partition, the classwise coverage inequalities imply the displayed global coverage inequality by summation, but stating it explicitly makes the role of $\sigma$ in the risk bound transparent.

\subsection{Proof of Theorem 1}\label{section: proof of theorem 1}

\begin{lemma}
\label{lemma: sufficient condition of small Err}
Given a $(\sigma, \delta)$-augmentation, if the encoder $f$ with $R_1 \leq \norm{f(\bm{x})}_2 \leq R_2$ for all $\bm{x} \in \mathcal X$ is $L$-Lipschitz and satisfies the condition
\begin{align*}
\bm\mu_T(i)^\top\bm\mu_T(j) < R_2^2\psi(\sigma, \delta, \eps, f)
\end{align*}
for any pair of $(i,j)$ with $i \neq j$, then for any $\eps > 0$, the downstream misclassification rate of $G_f$ obeys
\begin{align*}
\text{\rm Err}(G_f) \leq (1 - \sigma) + U_T(\eps, f),
\end{align*}
where $U_T(\eps, f) = \P_T\big\{X_T: \sup_{A_1,A_2\in\mathcal A}\norm{f(A_1(X_T)) - f(A_2(X_T))}_2 > \eps\big\}$. Set
\begin{equation*}
    \bar d_{\eps}=\min\left\{L\delta+2\eps,\ R_2+\frac{R_1^2}{R_2}\right\}.
\end{equation*}
Then
\begin{align}\label{eq: psi population threshold}
    \psi(\sigma, \delta, \eps,f) &= \Gamma_{\min}(\sigma, \delta, \eps, f) - \sqrt{2 - 2\Gamma_{\min}(\sigma, \delta, \eps, f)} \nonumber\\
    &\quad - \frac{1}{2}\Big(1 - \frac{\min_k\norm{\hat{\bm\mu}_T(k)}_2^2}{R_2^2}\Big) - \frac{2\max_k\norm{\hat{\bm\mu}_T(k) - \bm\mu_T(k)}_2}{R_2}.
\end{align}
Here, the threshold function $\Gamma_{\min}(\sigma, \delta, \eps, f)$ is given by
\begin{align*}
    \Gamma_{\min}(\sigma, \delta, \eps, f) = \begin{cases}
    (2\sigma - 1) - \frac{2U_T(\eps, f)}{\min_ip_T(i)} - \Big(\sigma - \frac{U_T(\eps, f)}{\min_ip_T(i)}\Big)\frac{\bar d_{\eps}}{R}, &R_1 = R_2 = R, \\
    \big(\sigma - \frac{U_T(\eps, f)}{\min_ip_T(i)}\big)\big\{1 + \big(\frac{R_1}{R_2}\big)^2 - \frac{\bar d_{\eps}}{R_2}\big\} - 1, & R_1 < R_2.
\end{cases}
\end{align*}
\end{lemma}

\begin{proof}
For any encoder $f$, define the set of stable observations as $V_T(\eps, f):= \{X_T: \sup_{A_1,A_2\in\mathcal A} \norm{f(A_1(X_T)) - f(A_2(X_T))}_2 \leq \eps\}$. If every $X_T \in \big\{C_T^\circ(1) \cup \cdots \cup C_T^\circ(K)\big\}\cap V_T(\eps, f)$ is correctly classified by $G_f$, the misclassification risk $\mathrm{Err}(G_f)$ can be bounded by $(1 - \sigma) + U_T(\eps, f)$. Indeed,
\begin{align*}
\mathrm{Err}(G_f) &= \sum_{k=1}^K\P_T\big\{G_f(X_T) \neq k, Y = k\big\} \\
&\leq \P_T\Big[\big\{\{C_T^\circ(1) \cup \cdots \cup C_T^\circ(K)\} \cap V_T(\eps, f)\big\}^c\Big] \\
&= \P_T\Big[\big(C_T^\circ(1) \cup \cdots \cup C_T^\circ(K)\big)^c \cup \big\{V_T(\eps, f)\big\}^c\Big] \leq (1 - \sigma) + \P_T\big[\{V_T(\eps, f)\}^c\big] \\
&= (1 - \sigma) + U_T(\eps, f),
\end{align*}
where the final equality follows from $U_T(\eps, f)=\P_T\{V_T(\eps,f)^c\}$. It therefore suffices to demonstrate that for a fixed $1 \leq i \leq K$, any $X_T \in C_T^\circ(i) \cap V_T(\eps, f)$ is correctly identified by $G_f$ provided that for all $j \neq i$:
\begin{align*}
\bm\mu_T(i)^\top\bm\mu_T(j) < R_2^2\psi(\sigma, \delta, \eps, f).
\end{align*}

Without loss of generality, we consider the case $i = 1$. According to the definition of the nearest-centroid rule, $X_T \in C_T^\circ(1) \cap V_T(\eps, f)$ is correctly classified if for any $k \neq 1$, $\norm{f(X_T) - \hat{\bm\mu}_T(1)}_2 < \norm{f(X_T) - \hat{\bm\mu}_T(k)}_2$, which is equivalent to
\begin{align}\label{eq: sufficient condition to be classified as 1}
f(X_T)^{\top}\hat{\bm\mu}_T(1) - f(X_T)^\top\hat{\bm\mu}_T(k) - \Big(\frac{1}{2}\norm{\hat{\bm\mu}_T(1)}^2_2 - \frac{1}{2}\norm{\hat{\bm\mu}_T(k)}^2_2\Big) > 0.
\end{align}
To bound the first term $f(X_T)^\top\hat{\bm\mu}_T(1)$, we decompose it as follows:
\begingroup
\small
\begin{align}
\label{eq: correct score decomposition}
&f(X_T)^{\top}\hat{\bm\mu}_T(1) = f(X_T)^\top\bm\mu_T(1) + f(X_T)^\top\{\hat{\bm\mu}_T(1) - \bm\mu_T(1)\} \nonumber\\
&\geq f(X_T)^{\top}\E_{(X_T, Y) \sim \P_T}\E_{A \sim U_{\mathcal{A}}}\{f(A(X_T)) \vert Y = 1\} - \norm{f(X_T)}_2\norm{\hat{\bm\mu}_T(1) - \bm\mu_T(1)}_2 \nonumber\\
&\geq \frac{1}{p_T(1)}f(X_T)^\top\E_{(X_T, Y) \sim \P_T}\E_{A \sim U_{\mathcal{A}}}\Big[f(A(X_T))\1\big\{X_T \in C_T(1)\big\}\Big] - R_2\norm{\hat{\bm\mu}_T(1) - \bm\mu_T(1)}_2 \nonumber\\
&= \frac{1}{p_T(1)}f(X_T)^{\top}\E_{(X_T, Y) \sim \P_T}\E_{A \sim U_{\mathcal{A}}}\Big[f(A(X_T))\1\big\{X_T \in C_T(1) \cap C_T^\circ(1) \cap V_T(\eps, f)\big\}\Big] \nonumber\\
&\quad + \frac{1}{p_T(1)}f(X_T)^\top\E_{(X_T, Y) \sim \P_T}\E_{A \sim U_{\mathcal{A}}}\Big[f(A(X_T))\1\big\{X_T \in C_T(1) \cap \big(C_T^\circ(1) \cap V_T(\eps, f)\big)^c\big\}\Big] \nonumber \\
&\quad - R_2\norm{\hat{\bm\mu}_T(1) - \bm\mu_T(1)}_2 \nonumber\\
&= \frac{\P_T\big\{C_T^\circ(1) \cap V_T(\eps, f)\big\}}{p_T(1)}f(X_T)^\top\E_{(X_T ,Y) \sim \P_T}\E_{A \sim U_{\mathcal{A}}}\Big\{f(A(X_T)) \big\vert X_T \in C_T^\circ(1) \cap V_T(\eps, f)\Big\} \nonumber \\
&\quad + \frac{1}{p_T(1)}\E_{(X_T, Y) \sim \P_T}\Big[\E_{A \sim U_{\mathcal{A}}}\big\{f(X_T)^\top f(A(X_T))\big\}\1\big\{X_T \in C_T(1) \backslash \big(C_T^\circ(1) \cap V_T(\eps, f)\big)\big\}\Big] \nonumber \\
&\quad - R_2\norm{\hat{\bm\mu}_T(1) - \bm\mu_T(1)}_2 \nonumber\\
&\geq \frac{\P_T\big\{C_T^\circ(1) \cap V_T(\eps, f)\big\}}{p_T(1)}f(X_T)^{\top}\E_{(X_T ,Y) \sim \P_T}\E_{A \sim U_{\mathcal{A}}}\big\{f(A(X_T)) \big\vert X_T \in C_T^\circ(1) \cap V_T(\eps, f)\big\} \nonumber\\
&\quad- \frac{R_2^2}{p_T(1)}\P_T\Big[C_T(1) \backslash \big\{C_T^\circ(1) \cap V_T(\eps, f)\big\}\Big] - R_2\norm{\hat{\bm\mu}_T(1) - \bm\mu_T(1)}_2,
\end{align}
\endgroup
where the second and third inequalities result from the range constraint $\norm{f(\bm{x})}_2 \leq R_2$. We further observe the following probabilistic bounds:
\begin{align}
\label{eq: unstable class mass upper bound}
\P_T\Big[C_T(1) \backslash \big\{C_T^\circ(1) \cap V_T(\eps, f)\big\}\Big] &= \P_T\Big[\{C_T(1) \backslash C_T^\circ(1)\} \cup \big\{C_T^\circ(1) \cap \big(V_T(\eps, f)\big)^c\}\Big] \nonumber \\ 
&\leq (1 - \sigma)p_T(1) + U_T(\eps, f),
\end{align}
and
\begin{align}
\label{eq: stable class mass lower bound}
\P_T\big\{C_T^\circ(1) \cap V_T(\eps, f)\big\} &= \P_T\{C_T(1)\} - \P_T\Big[C_T(1) \backslash \big\{C_T^\circ(1) \cap V_T(\eps, f)\big\}\Big] \nonumber \\
&\geq p_T(1) - \big\{(1 - \sigma)p_T(1) + U_T(\eps, f)\big\} \nonumber \\ 
&= \sigma p_T(1) - U_T(\eps, f).
\end{align}
Set
\begin{equation*}
q_1=\frac{\P_T\{C_T^\circ(1)\cap V_T(\eps,f)\}}{p_T(1)},
\qquad
a_1=\sigma-\frac{U_T(\eps,f)}{p_T(1)},
\end{equation*}
and let $H_1$ denote the conditional inner product defined below. By \eqref{eq: stable class mass lower bound}, $q_1\geq a_1$, and \eqref{eq: correct score decomposition} gives
\begin{align}
\label{eq: correct score lower bound intermediate}
&f(X_T)^{\top}\hat{\bm\mu}_T(1) \nonumber \\
&\geq q_1 H_1-R_2^2(1-q_1)
-R_2\norm{\hat{\bm\mu}_T(1)-\bm\mu_T(1)}_2,
\end{align}
Given that $X_T \in C_T^\circ(1) \cap V_T(\eps, f)$, by the property of $(\sigma, \delta)$-augmentation, for any $X_T^\prime \in C_T^\circ(1)\cap V_T(\eps, f)$, there exist
$(A^\star,A'^\star)\in\argmin_{A,A'\in\mathcal A}
\norm{A(X_T)-A'(X_T^\prime)}_2$ such that
$\norm{A^\star(X_T)-A'^\star(X_T^\prime)}_2\leq\delta$.
Combined with the $L$-Lipschitzness of $f$, this implies $\norm{f(A^\star(X_T)) - f(A'^\star(X_T^\prime))}_2 \leq L\delta$. Since $X_T$ and $X_T'$ are both in $V_T(\eps, f)$ and the identity belongs to $\mathcal A$, for every $A\in\mathcal A$ the triangle inequality gives $\norm{f(X_T)-f(A(X_T^\prime))}_2\leq L\delta+2\eps$. It follows that
\begin{align*}
H_1
&:=f(X_T)^{\top}\E_{(X_T^\prime ,Y^\prime) \sim \P_T}\E_{A \sim U_{\mathcal{A}}}\big\{f(A(X_T^\prime)) \big\vert X_T^\prime \in C_T^\circ(1) \cap V_T(\eps, f)\big\} \\
&= \E_{(X_T^\prime ,Y^\prime) \sim \P_T}\E_{A \sim U_{\mathcal{A}}}\big\{f(X_T)^\top f(A(X_T^\prime))\big\vert X_T^\prime \in C_T^\circ(1) \cap V_T(\eps, f)\big\} \nonumber \\
&\geq \norm{f(X_T)}_2^2-R_2\E\big\{\norm{f(A(X_T^\prime))-f(X_T)}_2\mid X_T^\prime \in C_T^\circ(1) \cap V_T(\eps, f)\big\}\nonumber\\
&\geq R_1^2-R_2(L\delta+2\eps).
\end{align*}
Cauchy--Schwarz also gives $H_1\geq-R_2^2$. Combining the two lower bounds yields
\begin{equation}\label{eq: fEf}
H_1\geq\max\{R_1^2-R_2(L\delta+2\eps),-R_2^2\}
=R_1^2-R_2\bar d_{\eps}.
\end{equation}
Since
$R_1^2+R_2^2-R_2\bar d_{\eps}\geq0$ and $q_1\geq a_1$, substituting \eqref{eq: fEf} into \eqref{eq: correct score lower bound intermediate} gives
\begin{align}\label{eq: correct score lower bound}
f(X_T)^{\top}\hat{\bm\mu}_T(1) &\geq a_1\big(R_1^2 - R_2\bar d_{\eps}\big) - R_2^2(1-a_1) \nonumber\\
&\qquad - R_2\norm{\hat{\bm\mu}_T(1) - \bm\mu_T(1)}_2 \nonumber\\
&= R_2^2\Gamma_1(\sigma, \delta, \eps, f) - R_2\norm{\hat{\bm\mu}_T(1) - \bm\mu_T(1)}_2,
\end{align}
where $\Gamma_1(\sigma, \delta, \eps, f)$ is obtained from the expression for $\Gamma_{\min}$ in Lemma~\ref{lemma: sufficient condition of small Err} by replacing $\min_i p_T(i)$ with $p_T(1)$.

For the cross-term $f(X_T)^\top\hat{\bm\mu}_T(k)$ in
\eqref{eq: sufficient condition to be classified as 1}, the preceding expansion yields
$f(X_T)^{\top}\bm\mu_T(1) \geq R_2^2\Gamma_1(\sigma, \delta, \eps, f)$.
Since $\norm{\bm\mu_T(k)}_2 \leq R_2$ for $1 \leq k \leq K$, we have
\begin{align}\label{eq: competing score upper bound}
f(X_T)^\top\hat{\bm\mu}_T(k)
&\leq f(X_T)^{\top}\bm\mu_T(k)
  + f(X_T)^{\top}(\hat{\bm\mu}_T(k)-\bm\mu_T(k)) \nonumber \\
&\leq f(X_T)^{\top}\bm\mu_T(k)
  + R_2\norm{\hat{\bm\mu}_T(k)-\bm\mu_T(k)}_2 \nonumber \\
&= (f(X_T)-\bm\mu_T(1))^{\top}\bm\mu_T(k)
  + \bm\mu_T(1)^\top\bm\mu_T(k) \nonumber\\
&\qquad + R_2\norm{\hat{\bm\mu}_T(k)-\bm\mu_T(k)}_2 \nonumber\\
&\leq \norm{f(X_T)-\bm\mu_T(1)}_2 R_2
  + \bm\mu_T(1)^\top\bm\mu_T(k) \nonumber\\
&\qquad + R_2\norm{\hat{\bm\mu}_T(k)-\bm\mu_T(k)}_2 \nonumber\\
&\leq R_2\sqrt{\norm{f(X_T)}_2^2
  -2f(X_T)^{\top}\bm\mu_T(1)+\norm{\bm\mu_T(1)}_2^2} \nonumber\\
&\qquad + \bm\mu_T(1)^\top\bm\mu_T(k)
  + R_2\norm{\hat{\bm\mu}_T(k)-\bm\mu_T(k)}_2 \nonumber\\
&\leq R_2\sqrt{2R_2^2-2R_2^2\Gamma_1(\sigma,\delta,\eps,f)}
  + \bm\mu_T(1)^\top\bm\mu_T(k) \nonumber\\
&\qquad + R_2\norm{\hat{\bm\mu}_T(k)-\bm\mu_T(k)}_2 \nonumber\\
&= \sqrt{2}R_2^2\sqrt{1-\Gamma_1(\sigma,\delta,\eps,f)}
  + \bm\mu_T(1)^\top\bm\mu_T(k) \nonumber\\
&\qquad + R_2\norm{\hat{\bm\mu}_T(k)-\bm\mu_T(k)}_2.
\end{align}
Combining \eqref{eq: correct score lower bound} and \eqref{eq: competing score upper bound} into \eqref{eq: sufficient condition to be classified as 1} concludes the proof as follows:
\begin{align*}
&f(X_T)^{\top}\hat{\bm\mu}_T(1) - f(X_T)^{\top}\hat{\bm\mu}_T(k) - \Big(\frac{1}{2}\norm{\hat{\bm\mu}_T(1)}_2^2 - \frac{1}{2}\norm{\hat{\bm\mu}_T(k)}^2_2\Big) \\
&\geq f(X_T)^{\top}\hat{\bm\mu}_T(1) - f(X_T)^{\top}\hat{\bm\mu}_T(k) - \frac{1}{2}R_2^2\big(1 - \min_{k}\norm{\hat{\bm\mu}_T(k)}_2^2 / R_2^2\big) \\
&\geq R_2^2\Gamma_1(\sigma, \delta, \eps, f) - R_2\norm{\hat{\bm\mu}_T(1) - \bm\mu_T(1)}_2 - \sqrt{2}R_2^2\sqrt{1 - \Gamma_1(\sigma, \delta, \eps, f)} \\
&\qquad - \bm\mu_T(1)^\top\bm\mu_T(k) - R_2\norm{\hat{\bm\mu}_T(k) - \bm\mu_T(k)}_2 - \frac{1}{2}R_2^2\big(1 - \min_{k}\norm{\hat{\bm\mu}_T(k)}_2^2 / R_2^2\big) > 0,
\end{align*}
where the last inequality is directly guaranteed by the threshold condition in Lemma~\ref{lemma: sufficient condition of small Err}.
\end{proof}

\begin{proposition}[Geometric sufficient condition for leakage control]
\label{proposition: leakage support gap}
Under the notation of Assumption~\ref{assumption: transport mass}, write
$\bm Z=f(A(X_S))$ and fix $\Delta_{\mathrm{id}}>0$.
Define the total off-matched mass and the part
whose transport distance is less than $\Delta_{\mathrm{id}}$ by
\begin{equation*}
\begin{aligned}
    M_{\mathrm{off}}
    &:=\sum_{k=1}^K\sum_{j\neq\tau^\star(k)}
    \Q^\star\{X_S\in C_S(k),\mathcal C=j\},\\
    M_{\mathrm{off}}^{<\Delta_{\mathrm{id}}}
    &:=\sum_{k=1}^K\sum_{j\neq\tau^\star(k)}
    \Q^\star\big\{X_S\in C_S(k),\mathcal C=j,
    \norm{\bm Z-\mathcal R}_2<\Delta_{\mathrm{id}}\big\}.
\end{aligned}
\end{equation*}
Suppose that, for some fixed $\vartheta_{\mathrm{id}}\in[0,1)$,
\begin{equation*}
    M_{\mathrm{off}}^{<\Delta_{\mathrm{id}}}
    \leq\vartheta_{\mathrm{id}}M_{\mathrm{off}}.
\end{equation*}
Then
\begin{equation*}
    \mathrm{Leak}(\Q^\star)
    \leq\frac{2}{(1-\vartheta_{\mathrm{id}})\Delta_{\mathrm{id}}}
    \mathcal W(\P_f,\P_{\mathcal R}).
\end{equation*}
Consequently, the leakage inequality in
Assumption~\ref{assumption: transport mass} holds with
$C_{\mathrm{id}}=2/\{(1-\vartheta_{\mathrm{id}})\Delta_{\mathrm{id}}\}$.
The separate requirement $q_k^\star>0$ is not implied by this
condition.
\end{proposition}

The condition requires at least a fraction $1-\vartheta_{\mathrm{id}}$ of the
total off-matched mass to have transport distance at least
$\Delta_{\mathrm{id}}$. For example, if, with
$\mathsf S_k=\operatorname{supp}(\P_{f,S}(k))$ and
$\mathsf R_j=\operatorname{supp}(\P_{\mathcal R}(j))$,
\begin{equation*}
    \inf_{\bm z\in\mathsf S_k,\,\bm r\in\mathsf R_j}
    \norm{\bm z-\bm r}_2\geq\Delta_{\mathrm{id}}
    \qquad\text{whenever }j\neq\tau^\star(k),
\end{equation*}
then $M_{\mathrm{off}}^{<\Delta_{\mathrm{id}}}=0$, and setting
$\vartheta_{\mathrm{id}}=0$ gives $C_{\mathrm{id}}=2/\Delta_{\mathrm{id}}$.

\begin{proof}
Since $\tau^\star$ is a permutation and both component families form
partitions,
\begin{equation*}
    M_{\mathrm{off}}
    =\sum_{k=1}^K\{p_S(k)-q_k^\star\}
    =\sum_{k=1}^K\{\alpha_{\tau^\star(k)}-q_k^\star\}.
\end{equation*}
The cost is nonnegative. Restricting its integral to off-matched pairs at
distance at least $\Delta_{\mathrm{id}}$ gives
\begin{equation*}
\begin{aligned}
    \mathcal W(\P_f,\P_{\mathcal R})
    &=\int\norm{f(A(X_S))-\mathcal R}_2\,d\Q^\star\\
    &\geq\sum_{k=1}^K\sum_{j\neq\tau^\star(k)}
    \int\1\{X_S\in C_S(k),\mathcal C=j\}\\
    &\qquad\times\1\{\norm{f(A(X_S))-\mathcal R}_2\geq\Delta_{\mathrm{id}}\}
    \norm{f(A(X_S))-\mathcal R}_2\,d\Q^\star\\
    &\geq\Delta_{\mathrm{id}}
    \big(M_{\mathrm{off}}-M_{\mathrm{off}}^{<\Delta_{\mathrm{id}}}\big)\\
    &\geq(1-\vartheta_{\mathrm{id}})\Delta_{\mathrm{id}}M_{\mathrm{off}}.
\end{aligned}
\end{equation*}
Moreover, for every $k$,
\begin{equation*}
    p_S(k)-q_k^\star\leq M_{\mathrm{off}},
    \qquad
    \alpha_{\tau^\star(k)}-q_k^\star\leq M_{\mathrm{off}}.
\end{equation*}
Taking the maximum over $k$ in the definition of
$\mathrm{Leak}(\Q^\star)$ gives
\begin{equation*}
    \mathrm{Leak}(\Q^\star)
    \leq2M_{\mathrm{off}}
    \leq\frac{2}{(1-\vartheta_{\mathrm{id}})\Delta_{\mathrm{id}}}
    \mathcal W(\P_f,\P_{\mathcal R}),
\end{equation*}
as claimed.
\end{proof}

\begin{lemma}
\label{lemma: the effect of minimaxing our loss}
Given a finite $(\sigma, \delta)$-augmentation $\mathcal A=\{A_1,\ldots,A_M\}$ equipped with the uniform law $U_{\mathcal A}$, suppose the encoder $f$ satisfying $R_1 \leq \norm{f(\bm{x})}_2 \leq R_2$ for all $\bm{x} \in \mathcal X$ is $L$-Lipschitz continuous. Set $\bar R=\max\{R,R_2\}$. Under Assumption~\ref{assumption: Lip augmentation}, for any $\eps > 0$:
\begin{align}\label{eq: U_T < L}
U_T^2(\eps, f) \lesssim \eps^{-2}\big\{\mathcal{L}_{\mathrm{align}}(f) + \epsilon_1 + \epsilon_2\big\},
\end{align}
If Assumption~\ref{assumption: transport mass} also holds, then
\begin{align}\label{eq: downstream divergence bound}
\max_{i \neq j}\abs{\bm\mu_T(i)^\top\bm\mu_T(j)} \lesssim \mathcal{W}(\P_f, \P_\mathcal{R}) + \epsilon_1.
\end{align}
The hidden constant in \eqref{eq: U_T < L} may depend on $M$.
\end{lemma}

\begin{proof}
We first derive the bound for \eqref{eq: U_T < L}. Since the transformation family $\mathcal{A}$ is finite and carries a uniform distribution, the expected pairwise distance can be written as
\begin{align*}
    \E_{A_1, A_2 \sim U_{\mathcal{A}}}\big\Vert f(A_1(X_T)) - f(A_2(X_T))\big\Vert_2 = \frac{1}{M^2}\sum_{i = 1}^M\sum_{j = 1}^M\big\Vert f\big(A_i(X_T)\big) - f\big(A_j(X_T)\big)\big\Vert_2.
\end{align*}
Consequently, the supremum of variations is bounded by the sum of pairwise distances:
\begin{align*}
    \sup_{A_1,A_2\in\mathcal A}\big\Vert f(A_1(X_T)) - f(A_2(X_T))\big\Vert_2 &\leq M^2\E_{A_1,A_2\sim U_{\mathcal{A}}}\norm{f(A_1(X_T)) - f(A_2(X_T))}_2,
\end{align*}
which implies the following inclusion:
\begin{align*}
    &\Big\{X_T: \sup_{A_1,A_2\in\mathcal A}\norm{f(A_1(X_T)) - f(A_2(X_T))}_2 > \eps\Big\} \\
    &\subseteq \Big\{X_T:\mathop{\E}_{A_1, A_2 \sim U_{\mathcal{A}}}\norm{f(A_1(X_T)) - f(A_2(X_T))}_2 > \frac{\eps}{M^2}\Big\}.
\end{align*}
By invoking Markov's inequality, we obtain
\begin{align}
\label{eq: stability probability bound}
U^2_T(\eps, f) &\leq \P^2_T\Big(\E_{A_1, A_2 \sim U_{\mathcal{A}}}\big\Vert f(A_1(X_T)) - f(A_2(X_T))\big\Vert_2 > \frac{\eps}{M^2}\Big) \nonumber \\
&\lesssim \eps^{-2}\E_{(X_T, Y) \sim \P_T}\E_{A_1, A_2 \sim U_{\mathcal{A}}}\big\Vert f(A_1(X_T)) - f(A_2(X_T))\big\Vert^2_2.
\end{align}
To link the downstream error to the pre-training objective, we expand the expected downstream alignment as
\begin{align}
\label{eq: downstream alignment = upstream alignment + radius}
&\mathop{\E}_{(X_T, Y) \sim \P_T}\mathop{\E}_{A_1,A_2 \sim U_{\mathcal{A}}}\big\Vert f(A_1(X_T)) - f(A_2(X_T))\big\Vert_2^2 \nonumber\\
&= \mathcal{L}_{\mathrm{align}}(f) + \frac{1}{M^2}\sum_{i=1}^M\sum_{j=1}^M\sum_{l=1}^{d^\star}\Big[\E_{(X_T, Y) \sim \P_T}\big\{f_l(A_i(X_T)) - f_l(A_j(X_T))\big\}^2 \nonumber \\
&\qquad - \E_{X_S \sim \P_S}\big\{f_l\big(A_i(X_S)\big) - f_l\big(A_j(X_S)\big)\big\}^2\Big].
\end{align}
Under the distribution shift constraints, for any fixed $i, j$, we have
\begin{align}
\label{eq: alignment shift bound}
&\E_{(X_T, Y) \sim \P_T}\big\{f_l\big(A_i(X_T)\big) - f_l\big(A_j(X_T)\big)\big\}^2 - \E_{X_S \sim \P_S}\big\{f_l\big(A_i(X_S)\big) - f_l\big(A_j(X_S)\big)\big\}^2 \nonumber \\
&= \sum_{k=1}^{K}\Big[p_T(k)\mathop{\E}_{\P_T}\big\{h(X_T) \vert Y = k\big\} - p_S(k)\mathop{\E}_{\P_S}\big\{h(X_S) \vert X_S \in C_S(k)\big\}\Big] \lesssim \epsilon_1 + \epsilon_2,
\end{align}
where $h(X) = \{f_l(A_i(X)) - f_l(A_j(X))\}^2$ is a bounded Lipschitz function; its Lipschitz and range constants depend only on $L,Q,\bar R$. Combining \eqref{eq: stability probability bound}, \eqref{eq: downstream alignment = upstream alignment + radius}, and \eqref{eq: alignment shift bound} completes the proof of \eqref{eq: U_T < L}.

For the Wasserstein bound \eqref{eq: downstream divergence bound}, write
\begin{equation*}
\bm\mu_S(k)=\E\{f(A(X_S))\mid X_S\in C_S(k)\},
\qquad
\bm r(j)=\E\{\mathcal R\mid\mathcal C=j\}.
\end{equation*}
Set $Z=f(A(X_S))$, let $\alpha_j=\P(\mathcal C=j)$ and, for each $k$, define the events
\begin{equation*}
    I_k=\{X_S\in C_S(k)\},\qquad
    J_k=\{\mathcal C=\tau^\star(k)\},\qquad
    E_k=I_k\cap J_k.
\end{equation*}
Define $q_k^\star:=\Q^\star(E_k)$. Then $\Q^\star(I_k)=p_S(k)$ and
$\Q^\star(J_k)=\alpha_{\tau^\star(k)}$. Decomposing the two unnormalised
component means over the matched block and its complements gives
\begin{align*}
&p_S(k)\bm\mu_S(k)
-\alpha_{\tau^\star(k)}\bm r(\tau^\star(k))\\
&\quad=
\int_{E_k}(\bm z-\bm r)\,d\Q^\star
+\int_{I_k\setminus E_k}\bm z\,d\Q^\star
-\int_{J_k\setminus E_k}\bm r\,d\Q^\star.
\end{align*}
Because $\norm{\bm z}_2\leq R_2\leq\bar R$ and
$\norm{\bm r}_2\leq R\leq\bar R$, the triangle inequality and
the optimality of $\Q^\star$ yield
\begin{align*}
&\norm{p_S(k)\bm\mu_S(k)
-\alpha_{\tau^\star(k)}\bm r(\tau^\star(k))}_2\\
&\quad\leq
\int_{E_k}\norm{\bm z-\bm r}_2\,d\Q^\star
+\bar R\{p_S(k)-q_k^\star\}
+\bar R\{\alpha_{\tau^\star(k)}-q_k^\star\}\\
&\quad\leq
\mathcal W(\P_f,\P_{\mathcal R})
+\bar R\big[p_S(k)+\alpha_{\tau^\star(k)}-2q_k^\star\big].
\end{align*}
Moreover,
$|p_S(k)-\alpha_{\tau^\star(k)}|
\leq p_S(k)+\alpha_{\tau^\star(k)}-2q_k^\star$ and
$\norm{\bm r(\tau^\star(k))}_2\leq\bar R$. It follows that
\begin{align*}
p_S(k)\norm{\bm\mu_S(k)-\bm r(\tau^\star(k))}_2
&\leq\mathcal W(\P_f,\P_{\mathcal R})\\
&\quad+2\bar R\big[p_S(k)+\alpha_{\tau^\star(k)}-2q_k^\star\big].
\end{align*}
Since $q_k^\star\leq p_S(k)$ and $q_k^\star>0$ for every $k$,
$p_{S,\min}:=\min_kp_S(k)>0$.
Assumption~\ref{assumption: transport mass} therefore implies
\begin{equation*}
\max_k\norm{\bm\mu_S(k)-\bm r(\tau^\star(k))}_2
\lesssim\mathcal W(\P_f,\P_{\mathcal R}),
\end{equation*}
where the hidden constant may depend on $p_{S,\min}^{-1}$ and
$C_{\mathrm{id}}$.
The rotational symmetry of the $j$th reference component around $\bm c_j$ implies $\bm r(j)=a_\epsilon\bm c_j$ for a scalar $a_\epsilon$. Hence $\bm r(i)^\top\bm r(j)=0$ for $i\neq j$.
Expanding the source cross-products around these orthogonal means yields
\begin{equation*}
\max_{i\neq j}|\bm\mu_S(i)^\top\bm\mu_S(j)|
\lesssim\mathcal W(\P_f,\P_{\mathcal R}).
\end{equation*}
Finally, the structural-shift and Lipschitz assumptions imply
\begin{equation*}
\max_k\norm{\bm\mu_T(k)-\bm\mu_S(k)}_2\lesssim\epsilon_1.
\end{equation*}
Expanding once more proves \eqref{eq: downstream divergence bound}.
\end{proof}

Equipped with the preceding lemmas, we now present the proof of Theorem~\ref{theorem: pop theorem}.

\begin{theorem}[General version of Theorem 1] \label{theorem: general version of the pop theorem}
Given a finite $(\sigma, \delta)$-augmentation equipped with the uniform law, suppose the encoder $f$ with $R_1 \leq \norm{f(\bm{x})}_2 \leq R_2$ for all $\bm x\in\mathcal X$ is $L$-Lipschitz and Assumption~\ref{assumption: Lip augmentation} holds. Set $\bar R=\max\{R,R_2\}$, and suppose Assumption~\ref{assumption: transport mass} holds. Then $\max_{i \neq j}\abs{\bm\mu_T(i)^\top\bm\mu_T(j)} \lesssim \mathcal{L}(f) + \epsilon_1$. Furthermore, for any $\eps>0$, if $\max_{i\neq j}\bm\mu_T(i)^\top\bm\mu_T(j) < R_2^2\psi(\sigma, \delta, \eps, f)$, then the downstream misclassification rate satisfies
\begin{align*}
\mathrm{Err}(G_f) \leq (1 - \sigma) + \mathcal{O}\Big(\eps^{-1}\big\{\mathcal{L}(f) + \epsilon_1 + \epsilon_2\big\}^{\frac{1}{2}}\Big).
\end{align*}
\end{theorem}

\begin{proof}
By \eqref{eq: downstream divergence bound} and the definition
$\mathcal L(f)=\mathcal L_{\mathrm{align}}(f)+\lambda\mathcal W(\P_f,\P_{\mathcal R})$,
\begin{equation*}
    \max_{i\neq j}\abs{\bm\mu_T(i)^\top\bm\mu_T(j)}
    \lesssim \mathcal W(\P_f,\P_{\mathcal R})+\epsilon_1
    \lesssim \mathcal L(f)+\epsilon_1,
\end{equation*}
where the hidden constant may depend on the fixed $\lambda$. Under the stated
separation condition, Lemma~\ref{lemma: sufficient condition of small Err}
and \eqref{eq: U_T < L} give
\begin{align*}
    \mathrm{Err}(G_f)
    &\leq(1-\sigma)+U_T(\eps,f)\\
    &\leq(1-\sigma)+\mathcal O\!\left(
    \eps^{-1}\{\mathcal L_{\mathrm{align}}(f)+\epsilon_1+\epsilon_2\}^{1/2}
    \right)\\
    &\leq(1-\sigma)+\mathcal O\!\left(
    \eps^{-1}\{\mathcal L(f)+\epsilon_1+\epsilon_2\}^{1/2}
    \right).
\end{align*}
This proves the general statement. Theorem~\ref{theorem: pop theorem} follows by setting
$R_1=R_2=R$.
\end{proof}

\subsection{Conditional extension when $K'>K$}\label{section: fine centre aggregation}

We prove the two stages stated in Remark~\ref{remark: overspecified theory}.
First, relabel the $K'$ reference components according to the latent fine index
$J$. Under the fine-level conditions in the remark, Theorem~\ref{theorem: pop theorem} applies directly
with $J$ and $K'$ replacing $Y$ and $K$. At this population stage, take each
fine centre to be its population value $\bm\nu_T(j)$; hence the centre-estimation
error in \eqref{eq: psi population threshold} vanishes. Whenever the corresponding
fine-level separation condition holds, the argument of
Theorem~\ref{theorem: pop theorem}, applied to these fine labels, gives
\begin{equation*}
    \P_T\{J_f^\star(X_T)\neq J\}
    \leq(1-\sigma)+
    \mathcal O\!\left(\eps^{-1}
    \{\mathcal L(f)+\epsilon_1+\epsilon_2\}^{1/2}\right).
\end{equation*}
Since $Y=a^\star(J)$ and $G_f^\star=a^\star\circ J_f^\star$, it follows that
$\mathrm{Err}(G_f^\star)\leq\P_T\{J_f^\star(X_T)\neq J\}$. The second stage
controls the additional error incurred when the fine centres and their
aggregation must be recovered from coarse labels; it does not require observed
fine-component labels.

\begin{proposition}[Fine-centre recovery and aggregation]\label{proposition: fine centre aggregation}
Suppose $\sup_{\bm x\in\mathcal X}\norm{f(\bm x)}_2\leq R$. Let $\{\bm\nu_T(j)\}_{j=1}^{K'}$ and $a^\star$ be the population fine centres and aggregation map defined in Remark~\ref{remark: overspecified theory}, and let $\{\widehat{\bm\nu}_T(j)\}_{j=1}^{K'}$ lie in the radius-$R$ ball. Let $P_{K'}$ denote the set of permutations of $\{1,\ldots,K'\}$. Suppose that, on an event of probability at least $1-\zeta_{n_T}$, there is a permutation $\pi\in P_{K'}$ such that
\begin{equation}\label{eq: supplementary fine recovery}
    \max_{1\leq j\leq K'}
    \norm{\widehat{\bm\nu}_T(\pi(j))-\bm\nu_T(j)}_2
    \leq\eta_{n_T},
    \qquad
    \widehat a(\pi(j))=a^\star(j),\quad j=1,\ldots,K'.
\end{equation}
Then, on this event,
\begin{equation}\label{eq: fine score perturbation}
\begin{aligned}
    \sup_{\bm x\in\mathcal X}\max_{1\leq j\leq K'}
    \big\vert
    \{\widehat{\bm W}f(\bm x)+\hat{\bm b}\}_{\pi(j)}
    -\{\bm W^\star f(\bm x)+\bm b^\star\}_j
    \big\vert
    \leq4R\eta_{n_T}.
\end{aligned}
\end{equation}
Consequently, averaging over the sample used by the recovery algorithm,
\begin{equation}\label{eq: supplementary overspecified risk}
    \E_{\widetilde{\mathcal D}_T}\{\mathrm{Err}(\widehat G)\}
    \leq\mathrm{Err}(G_f^\star)
    +\P_T\!\left\{m_f^\star(X_T)\leq8R\eta_{n_T}\right\}
    +\zeta_{n_T}.
\end{equation}
\end{proposition}

\begin{proof}
Fix $\bm x\in\mathcal X$ and $j\in\{1,\ldots,K'\}$. On the event in \eqref{eq: supplementary fine recovery},
\begin{align*}
\big\vert
\{\widehat{\bm W}f(\bm x)+\hat{\bm b}\}_{\pi(j)}
-\{\bm W^\star f(\bm x)+\bm b^\star\}_j
\big\vert
&\leq2\norm{\widehat{\bm\nu}_T(\pi(j))-\bm\nu_T(j)}_2
\norm{f(\bm x)}_2\\
&\quad+\abs{\norm{\widehat{\bm\nu}_T(\pi(j))}_2^2
-\norm{\bm\nu_T(j)}_2^2}\\
&\leq2R\eta_{n_T}
\quad+\norm{\widehat{\bm\nu}_T(\pi(j))-\bm\nu_T(j)}_2\\
&\qquad\times
\{\norm{\widehat{\bm\nu}_T(\pi(j))}_2+\norm{\bm\nu_T(j)}_2\}\\
&\leq4R\eta_{n_T},
\end{align*}
which proves \eqref{eq: fine score perturbation}. If
$m_f^\star(\bm x)>8R\eta_{n_T}$, the score of $\pi(J_f^\star(\bm x))$ under the estimated centres remains larger than every competing estimated score. Hence
$\widehat J(\bm x)=\pi(J_f^\star(\bm x))$ and, by \eqref{eq: supplementary fine recovery},
\begin{equation*}
    \widehat G(\bm x)
    =\widehat a(\pi(J_f^\star(\bm x)))
    =a^\star(J_f^\star(\bm x))
    =G_f^\star(\bm x).
\end{equation*}
Thus, on the recovery event, $\widehat G$ can make an additional error relative to $G_f^\star$ only when the fine margin is small. Let $\Omega_{\mathrm{rec}}$ denote the recovery event in \eqref{eq: supplementary fine recovery}. On $\Omega_{\mathrm{rec}}$, the score comparison above implies that $\widehat G(X_T)=G_f^\star(X_T)$ whenever $m_f^\star(X_T)>8R\eta_{n_T}$. Consequently, for every $X_T$,
\begin{align*}
\1\left\{\widehat G(X_T)\neq Y\right\}
&\leq \1\left\{G_f^\star(X_T)\neq Y\right\}
  +\1\left\{m_f^\star(X_T)\leq8R\eta_{n_T}\right\},
\qquad\text{on }\Omega_{\mathrm{rec}}.
\end{align*}
Taking expectation with respect to $X_T\sim\P_T$ gives, on $\Omega_{\mathrm{rec}}$,
\begin{align*}
\mathrm{Err}(\widehat G)
&=\E_{(X_T,Y)\sim\P_T}\left[\1\left\{\widehat G(X_T)\neq Y\right\}\right]\\
&\leq \E_{(X_T,Y)\sim\P_T}\left[\1\left\{G_f^\star(X_T)\neq Y\right\}\right]
  +\E_{X_T\sim\P_T}\left[\1\left\{m_f^\star(X_T)\leq8R\eta_{n_T}\right\}\right]\\
&=\mathrm{Err}(G_f^\star)+\P_T\left\{m_f^\star(X_T)\leq8R\eta_{n_T}\right\}.
\end{align*}
On $\Omega_{\mathrm{rec}}^c$, we use $\mathrm{Err}(\widehat G)\leq1$. Therefore,
\begin{align*}
\E_{\widetilde{\mathcal D}_T}\left\{\mathrm{Err}(\widehat G)\right\}
&=\E_{\widetilde{\mathcal D}_T}\left\{\mathrm{Err}(\widehat G)\1_{\Omega_{\mathrm{rec}}}\right\}
  +\E_{\widetilde{\mathcal D}_T}\left\{\mathrm{Err}(\widehat G)\1_{\Omega_{\mathrm{rec}}^c}\right\}\\
&\leq \P(\Omega_{\mathrm{rec}})\left\{\mathrm{Err}(G_f^\star)
  +\P_T(m_f^\star(X_T)\leq8R\eta_{n_T})\right\}
  +\P(\Omega_{\mathrm{rec}}^c)\\
&\leq \mathrm{Err}(G_f^\star)
  +\P_T(m_f^\star(X_T)\leq8R\eta_{n_T})
  +\zeta_{n_T},
\end{align*}
where the last inequality uses $\P(\Omega_{\mathrm{rec}}^c)\leq\zeta_{n_T}$. This proves \eqref{eq: supplementary overspecified risk}.
Combining this inequality with the preceding fine-level population bound gives
the claimed two-stage extension.
\end{proof}

\section{Proof of Theorem 2}\label{section: Appendix B}

This section establishes the non-asymptotic convergence rates for the DM-pre-trained encoder. Throughout, $\mathcal{NN}_{p,q}(W,D,B)$ denotes the layer-wise ReLU network class defined in Section~\ref{subsection: sample complexity}, where $D$ is the number of hidden layers and $W$ is their maximum width. Our derivation of the sample complexity builds upon and extends the analysis framework for self-supervised objectives developed in \citet{duan2025advssl}. To facilitate the analysis, we first formalize the relationship between the population-level risk and its empirical counterpart.

Recall that the population optimal encoder is defined as the minimizer of the total self-supervised risk:
\begin{equation}\label{eq: risk at pop level}
    f^\star \in \argmin_{f \in \mathcal{F}}\mathcal{L}(f) := \mathcal{L}_{\mathrm{align}}(f) + \lambda \cdot \mathcal{W}(\mathbb{P}_f, \mathbb{P}_\mathcal{R}).
\end{equation}
By invoking the Kantorovich--Rubinstein duality \citep{Villani}, the Mallows distance $\mathcal{W}(\mathbb{P}_f, \mathbb{P}_\mathcal{R})$ can be expressed in its dual form. We use the critic class $\mathcal G$ defined in Section~\ref{subsection: estimation and computation}. This allows us to reframe the minimization of \eqref{eq: risk at pop level} as a \textit{saddle-point problem} over the encoder space $\mathcal{F}$ and the critic space $\mathcal{G}$:
\begin{equation}\label{eq: mini-max}
    (f^\star, g^\star) \in \argmin_{f \in \mathcal{F}}\max_{g \in \mathcal{G}}\mathcal{L}(f,g) :=\mathcal{L}_{\mathrm{align}}(f) + \lambda \cdot \mathcal{W}(f, g),
\end{equation}
where the witness term is given by $\mathcal{W}(f, g) = \mathbb{E}_{\mathcal{R} \sim \mathbb{P}_\mathcal{R}}\{g(\mathcal{R})\} - \mathbb{E}_{Z \sim \mathbb{P}_f}\{g(Z)\}$. The supremum of this witness over $\mathcal G$ equals the Mallows distance up to the fixed Lipschitz scale. With that scale absorbed into $\lambda$, we write $\mathcal{L}(f) = \sup_{g \in \mathcal{G}}\mathcal{L}(f, g)$. (The reverse sign gives the same supremum because $\mathcal G$ is symmetric, but this orientation agrees with the empirical criterion.)

In the finite-sample regime, the augmented source dataset is
$\widetilde{\mathcal D}_S=\{(X_{S,1}^{(i)},X_{S,2}^{(i)},\mathcal R^{(i)}):1\leq i\leq n_S\}$,
where each $X_S^{(i)}$ yields two stochastic views $X^{(i)}_{S,1}$ and
$X^{(i)}_{S,2}$, and the reference draws are independent. The empirical risk
at the sample level is constructed as:
\begin{equation}\label{eq: hatW(f,g)}
    \widehat{\mathcal{L}}(f, g) := \widehat{\mathcal{L}}_{\mathrm{align}}(f) + \lambda \cdot \widehat{\mathcal{W}}(f, g),
\end{equation}
where the empirical alignment loss and the empirical measure-matching term are respectively defined as:
\begin{align*}
    \widehat{\mathcal{L}}_{\mathrm{align}}(f)&=\frac{1}{n_S}\sum_{i=1}^{n_S}\norm{f(X^{(i)}_{S,1}) - f(X^{(i)}_{S,2})}_2^2, \\
    \widehat{\mathcal{W}}(f, g) &= \frac{1}{n_S}\sum_{i=1}^{n_S}\Biggl[g(\mathcal{R}^{(i)}) - \frac{1}{2}\Bigl\{g\bigl(f(X^{(i)}_{S,1})\bigr) + g\bigl(f(X^{(i)}_{S,2})\bigr)\Bigr\}\Biggr].
\end{align*}
Correspondingly, the empirical sample-optimum is obtained by solving the minimax optimisation:
\begin{equation}\label{eq: empirical minimax estimator}
    (\hat{f}_{n_S},\hat{g}_{n_S})\in
    \argmin_{f\in\widehat{\mathcal F}}
    \max_{g\in\widehat{\mathcal G}}\widehat{\mathcal L}(f,g).
\end{equation}
We remark that the empirical estimators $\hat{f}_{n_S}$ and $\hat{g}_{n_S}$ defined above correspond precisely to the neural network-parameterized functions $f_{\hat{\bm{\theta}}_{n_S}}$ and $g_{\hat{\bm{\omega}}_{n_S}}$ in \eqref{eq: empirical encoder notation}, where the parameters are obtained through the saddle-point optimisation in \eqref{eq: saddle point}.

\subsection{Error Decomposition}\label{subsection: error decomposition}
To analyze the performance gap, note that $\mathcal{L}(f) = \sup_{g \in \mathcal{G}}\mathcal{L}(f, g)$. We define the stochastic error $\mathcal{E}_{\mathrm{sta}}$, the encoder approximation error $\mathcal{E}_{\mathcal{F}}$, and the critic approximation error $\mathcal{E}_{\mathcal{G}}$ respectively as follows:
\begin{gather*}
    \mathcal{E}_{\mathrm{sta}} := \sup_{f \in \widehat{\mathcal{F}}, g \in \widehat{\mathcal{G}}}\abs{\mathcal{L}(f, g) - \widehat{\mathcal{L}}(f, g)},\\ 
    \mathcal{E}_{\mathcal{F}} := \inf_{f \in \widehat{\mathcal{F}}}\{\mathcal{L}(f) - \mathcal{L}(f^\star)\}, \\
    \mathcal{E}_{\mathcal{G}} := \lambda\sup_{f \in \widehat{\mathcal{F}}}\bigl\vert\sup_{g \in \mathcal{G}}\mathcal{W}(f, g) - \sup_{g \in \widehat{\mathcal{G}}}\mathcal{W}(f, g)\bigr\vert.
\end{gather*}

The following lemma establishes the fundamental relationship between these error components.
\begin{lemma}
$\mathcal{L}(\hat{f}_{n_S}) \leq \mathcal{L}(f^\star) +  2\mathcal{E}_{\mathrm{sta}} + \mathcal{E}_{\mathcal{F}} + 2\mathcal{E}_{\mathcal{G}}$.
\end{lemma}
\begin{proof}
    For any $f \in \widehat{\mathcal{F}}$, consider the following expansion:
    \begin{align*}
        \mathcal{L}(\hat{f}_{n_S}) &= \bigl\{\mathcal{L}(\hat{f}_{n_S}) - \sup_{g \in \widehat{\mathcal{G}}}\mathcal{L}(\hat{f}_{n_S}, g)\bigr\} + \bigl\{\sup_{g \in \widehat{\mathcal{G}}}\mathcal{L}(\hat{f}_{n_S}, g) - \sup_{g \in \widehat{\mathcal{G}}}\widehat{\mathcal{L}}(\hat{f}_{n_S}, g)\bigr\} \\
        &\quad +\bigl\{\sup_{g \in \widehat{\mathcal{G}}}\widehat{\mathcal{L}}(\hat{f}_{n_S}, g) - \sup_{g \in \widehat{\mathcal{G}}}\widehat{\mathcal{L}}(f, g)\bigr\} + \bigl\{\sup_{g \in \widehat{\mathcal{G}}}\widehat{\mathcal{L}}(f, g) - \sup_{g \in \widehat{\mathcal{G}}}\mathcal{L}(f, g)\bigr\} \\
        &\quad +\bigl\{\sup_{g \in \widehat{\mathcal{G}}}\mathcal{L}(f, g) - \mathcal{L}(f)\bigr\} + \bigl\{\mathcal{L}(f) - \mathcal{L}(f^\star)\bigr\} + \mathcal{L}(f^\star).
    \end{align*}
    For the second term and the fourth term, we observe that:
    \begin{equation*}
        \sup_{g \in \widehat{\mathcal{G}}}\mathcal{L}(\hat{f}_{n_S}, g) - \sup_{g \in \widehat{\mathcal{G}}}\widehat{\mathcal{L}}(\hat{f}_{n_S}, g) \leq \sup_{g \in \widehat{\mathcal{G}}}\{\mathcal{L}(\hat{f}_{n_S}, g) - \widehat{\mathcal{L}}(\hat{f}_{n_S}, g)\} \leq  \mathcal{E}_{\mathrm{sta}}
    \end{equation*}
    and
    \begin{equation*}
        \sup_{g \in \widehat{\mathcal{G}}}\widehat{\mathcal{L}}(f, g) - \sup_{g \in \widehat{\mathcal{G}}}\mathcal{L}(f, g) \leq \sup_{g \in \widehat{\mathcal{G}}}\{\widehat{\mathcal{L}}(f, g) - \mathcal{L}(f, g)\}\leq \mathcal{E}_{\mathrm{sta}}.
    \end{equation*}
    The first and the fifth terms are both bounded by $\mathcal{E}_{\mathcal{G}}$. For the first term:
    \begin{align*}
        \mathcal{L}(\hat{f}_{n_S}) - \sup_{g \in \widehat{\mathcal{G}}}\mathcal{L}(\hat{f}_{n_S}, g) &\leq \sup_{f \in \widehat{\mathcal{F}}}\{\mathcal{L}(f) - \sup_{g \in \widehat{\mathcal{G}}}\mathcal{L}(f, g)\} \\
        &= \lambda\sup_{f \in \widehat{\mathcal{F}}}\{\sup_{g \in \mathcal{G}}\mathcal{W}(f, g) - \sup_{g \in \widehat{\mathcal{G}}}\mathcal{W}(f, g)\} \\
        &\leq \lambda\sup_{f \in \widehat{\mathcal{F}}}\bigl\vert\sup_{g \in \mathcal{G}}\mathcal{W}(f, g) - \sup_{g \in \widehat{\mathcal{G}}}\mathcal{W}(f, g)\bigr\vert = \mathcal{E}_{\mathcal{G}}.
    \end{align*}
    Similar logic applies to the fifth term. Finally, taking the infimum over all $f \in \widehat{\mathcal{F}}$ yields the result.
\end{proof}

\subsection{The Stochastic Error}\label{subsection: The Stochastic Error}
To bound the stochastic variation, define the loss kernel $\ell(\bm{v}_1, \bm{v}_2, v_3, v_4, v_5) = \norm{\bm{v}_1 - \bm{v}_2}_2^2 + \lambda\{v_3 - \frac{1}{2} ( v_4 + v_5 )\}$, where $\bm{v}_1,\bm{v}_2 \in \mathbb{R}^{d^\star}$ and $v_3, v_4, v_5 \in \mathbb{R}$. Under this notation, the empirical risk can be expressed as
\begin{align*}
    \widehat{\mathcal{L}}(f,g) = \frac{1}{n_S} \sum_{i=1}^{n_S} \ell \bigl( f(X^{(i)}_{S,1}), f(X^{(i)}_{S,2}), g(\mathcal{R}^{(i)}), g(f(X^{(i)}_{S,1})), g(f(X^{(i)}_{S,2})) \bigr).
\end{align*}
Let $\widetilde{\mathcal{D}}^\prime_S = \bigl\{ (X_{S,1}^{\prime(i)}, X_{S,2}^{\prime(i)}, \mathcal{R}^\prime_i): 1 \leq i \leq n_S \bigr\}$ denote an independent copy of $\widetilde{\mathcal{D}}_S$. It follows that the population-level risk is given by
\begin{align*}
    \mathcal{L}(f,g) = \frac{1}{n_S} \sum_{i=1}^{n_S} \mathbb{E}_{\widetilde{\mathcal{D}}^\prime_S} \bigl\{ \ell \bigl( f(X_{S,1}^{\prime(i)}), f(X_{S,2}^{\prime(i)}), g(\mathcal{R}^\prime_i), g(f(X_{S,1}^{\prime(i)})), g(f(X_{S,2}^{\prime(i)})) \bigr) \bigr\}.
\end{align*}
Substituting these expressions into the definition of $\mathcal{E}_{\mathrm{sta}}$, we have
\begingroup
\small
\begin{align*}
    &\mathbb{E}_{\widetilde{\mathcal{D}}_S} \bigl\{ \mathcal{E}_{\mathrm{sta}} \bigr\} \nonumber\\
    &= \mathbb{E}_{\widetilde{\mathcal{D}}_S} \Bigl\{ \sup_{f \in \widehat{\mathcal{F}}, g \in \widehat{\mathcal{G}}} \lvert \mathcal{L}(f, g) - \widehat{\mathcal{L}}(f, g) \rvert \Bigr\} \\
    &\leq \mathbb{E}_{\widetilde{\mathcal{D}}_S} \Biggl[ \sup_{(f, g) \in \widehat{\mathcal{F}} \times \widehat{\mathcal{G}}} \biggl\lvert \frac{1}{n_S} \sum_{i=1}^{n_S} \mathbb{E}_{\widetilde{\mathcal{D}}^\prime_S} \bigl\{ \ell \bigl( f(X_{S,1}^{\prime(i)}), f(X_{S,2}^{\prime(i)}), g(\mathcal{R}^\prime_i), g(f(X_{S,1}^{\prime(i)})), g(f(X_{S,2}^{\prime(i)})) \bigr) \bigr\} \\
    &\quad - \frac{1}{n_S} \sum_{i=1}^{n_S} \ell \bigl( f(X_{S,1}^{(i)}), f(X_{S,2}^{(i)}), g(\mathcal{R}^{(i)}), g(f(X_{S,1}^{(i)})), g(f(X_{S,2}^{(i)})) \bigr) \biggr\rvert \Biggr] \\
    &\leq \mathbb{E}_{\widetilde{\mathcal{D}}_S, \widetilde{\mathcal{D}}_S^\prime} \Biggl\{ \sup_{(f, g) \in \widehat{\mathcal{F}} \times \widehat{\mathcal{G}}} \biggl\lvert \frac{1}{n_S} \sum_{i=1}^{n_S} \ell \bigl( f(X_{S,1}^{\prime(i)}), f(X_{S,2}^{\prime(i)}), g(\mathcal{R}^{\prime(i)}), g(f(X_{S,1}^{\prime(i)})), g(f(X_{S,2}^{\prime(i)})) \bigr) \\
    &\quad - \ell \bigl( f(X_{S,1}^{(i)}), f(X_{S,2}^{(i)}), g(\mathcal{R}^{(i)}), g(f(X_{S,1}^{(i)})), g(f(X_{S,2}^{(i)})) \bigr) \biggr\rvert \Biggr\} \\
    &\leq \mathbb{E}_{\widetilde{\mathcal{D}}_S, \widetilde{\mathcal{D}}_S^\prime, \bm{\xi}} \Biggl\{ \sup_{(f, g) \in \widehat{\mathcal{F}} \times \widehat{\mathcal{G}}} \biggl\lvert \frac{1}{n_S} \sum_{i=1}^{n_S} \xi_i \Bigl( \ell \bigl( f(X_{S,1}^{\prime(i)}), f(X_{S,2}^{\prime(i)}), g(\mathcal{R}^{\prime(i)}), g(f(X_{S,1}^{\prime(i)})), g(f(X_{S,2}^{\prime(i)})) \bigr) \\
    &\quad - \ell \bigl( f(X_{S,1}^{(i)}), f(X_{S,2}^{(i)}), g(\mathcal{R}^{(i)}), g(f(X_{S,1}^{(i)})), g(f(X_{S,2}^{(i)})) \bigr) \Bigr) \biggr\rvert \Biggr\},
\end{align*}
\endgroup
where the final inequality is a consequence of standard symmetrization techniques in empirical process theory \citep{gine2016mathematical}. Since $\widetilde{\mathcal{D}}^\prime_S$ is a random copy of $\widetilde{\mathcal{D}}_S$, the stochastic error can be further bounded by
\begingroup
\small
\begin{align}\label{eq: Esta < complexity}
    &\mathbb{E}_{\widetilde{\mathcal{D}}_S} \bigl\{ \mathcal{E}_{\mathrm{sta}} \bigr\} \nonumber \\
    &\leq 2 \mathbb{E}_{\widetilde{\mathcal{D}}_S, \bm{\xi}} \Biggl\{ \sup_{(f, g) \in \widehat{\mathcal{F}} \times \widehat{\mathcal{G}}} \biggl\lvert \frac{1}{n_S} \sum_{i=1}^{n_S} \xi_i \ell \bigl( f(X_{S,1}^{(i)}), f(X_{S,2}^{(i)}), g(\mathcal{R}^{(i)}), g(f(X_{S,1}^{(i)})), g(f(X_{S,2}^{(i)})) \bigr) \biggr\rvert \Biggr\} \nonumber \\
    &\lesssim \mathbb{E}_{\widetilde{\mathcal{D}}_S, \bm{\xi}} \Biggl[ \sup_{(f, g) \in \widehat{\mathcal{F}} \times \widehat{\mathcal{G}}} \biggl\lvert \frac{1}{n_S} \sum_{i=1}^{n_S} \sum_{j=1}^{d^\star} \bigl\{ \xi_{i,j,1} f_j(X_{S,1}^{(i)}) + \xi_{i,j,2} f_j(X_{S,2}^{(i)}) \bigr\} + \xi_{i,1} g(\mathcal{R}^{(i)})  \nonumber\\
    &\quad + \xi_{i,2} g \bigl( f(X_{S,1}^{(i)}) \bigr) + \xi_{i,3} g \bigl( f(X_{S,2}^{(i)}) \bigr) \biggr\rvert \Biggr] \nonumber\\
    &\lesssim \mathbb{E}_{\widetilde{\mathcal{D}}_S, \bm{\xi}} \Biggl\{ \sup_{f \in \widehat{\mathcal{F}}} \biggl\lvert \frac{1}{n_S} \sum_{i=1}^{n_S} \sum_{j=1}^{d^\star} \xi_{i,j} f_j(X_{S,1}^{(i)}) \biggr\rvert \Biggr\} + \mathbb{E}_{\widetilde{\mathcal{D}}_S, \bm{\xi}} \Biggl\{ \sup_{g \in \widehat{\mathcal{G}}} \biggl\lvert \frac{1}{n_S} \sum_{i=1}^{n_S} \xi_{i} g(\mathcal{R}^{(i)}) \biggr\rvert \Biggr\} \nonumber\\ 
    &\quad + \mathbb{E}_{\widetilde{\mathcal{D}}_S, \bm{\xi}} \Biggl\{ \sup_{(f,g) \in \widehat{\mathcal{F}} \times \widehat{\mathcal{G}}} \biggl\lvert \frac{1}{n_S} \sum_{i=1}^{n_S} \xi_{i} g \bigl( f(X_{S,1}^{(i)}) \bigr) \biggr\rvert \Biggr\},
\end{align}
\endgroup
where the second inequality results from the vector-contraction principle (Lemma~\ref{lemma: vector-contraction principle}), as established in \citet{maurer2016vectorcontraction} and \citet{gine2016mathematical}.

\begin{lemma}[Vector-contraction principle]
\label{lemma: vector-contraction principle}
Let $\mathcal{X}$ be any set, $(x_1, \ldots, x_n) \in \mathcal{X}^n$, let $\mathcal{H}$ be a class of functions $f:\mathcal{X} \rightarrow \ell_2$, and let $h_i: \ell_2 \rightarrow \mathbb{R}$ be a sequence of functions with Lipschitz constant $L^\prime$ and $h_i(\bm0)=0$. Then
\begin{align*}
    \mathbb{E} \sup_{f \in \mathcal{H}} \Bigl\lvert \sum_i \xi_i h_i \bigl( f(x_i) \bigr) \Bigr\rvert \leq 2\sqrt{2}L^\prime \mathbb{E} \sup_{f \in \mathcal{H}} \Bigl\lvert \sum_{i,k} \xi_{ik} f_k(x_i) \Bigr\rvert,
\end{align*}
where $\xi_{ik}$ is an independent doubly indexed Rademacher sequence and $f_k(x_i)$ denotes the $k$-th component of $f(x_i)$.
\end{lemma}

To bound the three terms in \eqref{eq: Esta < complexity}, we recall the following definitions and auxiliary results regarding covering numbers and sub-Gaussian processes.

\begin{definition}[Covering number]\label{def: covering number}
    Let $n \in \N$, $\mathcal{S} \subseteq \mathbb{R}^n$, and $\varrho > 0$. A set $\mathcal{N} \subseteq \mathcal{S}$ is called a $\varrho$-net of $\mathcal{S}$ with respect to a metric $d$ if for every $\bm{u} \in \mathcal{S}$, there exists $\bm{v} \in \mathcal{N}$ such that $d(\bm{u}, \bm{v}) \leq \varrho$. The covering number of $\mathcal{S}$ is defined as
    \begin{align*}
    \mathcal{N}(\varrho, \mathcal{S}, d) := \min \bigl\{ \lvert \mathcal{Q} \rvert : \mathcal{Q} \text{ is an $\varrho$-cover of }\mathcal{S} \bigr\}.
    \end{align*}
\end{definition}

\begin{definition}[Uniform covering number]\label{def: uniform covering number}
    Let $\mathcal{H}$ be a class of functions from $\mathcal{X}$ to $\mathbb{R}$. Given a sequence $x = (x_1, x_2, \ldots, x_k) \in \mathcal{X}^k$, let $\mathcal{H}_{\vert_x}$ be the subset of $\mathbb{R}^k$ defined by $\mathcal{H}_{\vert_x} = \bigl\{ (f(x_1), f(x_2), \ldots, f(x_k)): f \in \mathcal{H} \bigr\}$. For a positive number $\varrho$, the uniform covering number is given by
    \begin{align*}
        \mathcal{N}_\infty(\varrho, \mathcal{H}, k) = \max \bigl\{ \mathcal{N}(\varrho, \mathcal{H}_{\vert_x}, d_\infty) : x \in \mathcal{X}^k \bigr\}.
    \end{align*}
\end{definition}

\begin{lemma}[\citet{anthony1999neural}]\label{lemma: covering metric hierarchy}
    Let $\mathcal{H}$ be a class of functions from $\mathcal{X}$ to $\mathbb{R}$. For any $\varrho > 0$ and $x \in \mathcal{X}^k$, the following hierarchy for covering numbers holds:
    \begin{align*}
        \mathcal{N}(\varrho, \mathcal{H}_{\vert_x}, d_1) \leq \mathcal{N}(\varrho, \mathcal{H}_{\vert_x}, d_2) \leq \mathcal{N}(\varrho, \mathcal{H}_{\vert_x}, d_\infty),
    \end{align*}
    where $d_1(\bm{x}, \bm{y}) := \frac{1}{k} \sum_{i=1}^k \lvert x_i - y_i \rvert, d_2(\bm{x}, \bm{y}) := \bigl( \frac{1}{k} \sum_{i=1}^k (x_i - y_i)^2 \bigr)^{1/2}$, and $d_\infty(\bm{x}, \bm{y}) := \max_{1\leq i\leq k} \lvert x_i - y_i \rvert$.
\end{lemma}

\begin{definition}[Sub-Gaussian process]
    A centred stochastic process $\{X(t): t \in T\}$ is sub-Gaussian with respect to a distance $d$ if its increments satisfy the sub-Gaussian inequality:
    \begin{align*}
        \mathbb{E} \bigl[ e^{\varsigma \{ X(t) - X(s) \}} \bigr] \leq e^{\varsigma^2 d^2(s,t)/2}, \quad \forall \varsigma \in \mathbb{R}, \quad \forall s,t \in T.
    \end{align*}
\end{definition}

The following result relates the supremum of a sub-Gaussian process to its entropy integral \citep{gine2016mathematical}.
\begin{lemma}[Dudley's entropy integral]\label{lemma: dudley's entropy integral}
    Let $(T,d)$ be a separable pseudo-metric space, and let $\{X(t): t \in T\}$ be a sub-Gaussian process relative to $d$. Then
    \begin{align*}
        \mathbb{E} \sup_{t \in T} \lvert X(t) \rvert \leq \mathbb{E} \lvert X(t_0) \rvert + 4\sqrt{2} \int_0^{D/2} \sqrt{\log 2 \mathcal{N}(\varrho, T, d)} \, d\varrho,
    \end{align*}
    where $t_0$ is an arbitrary point in $T$ and $D$ is the diameter of $(T,d)$.
\end{lemma}
\begin{proof}
    Note that the entropy condition $\int_0^\infty \sqrt{\log\mathcal{N}(\varrho, T, d)} \, d\varrho < \infty$ in the original proof of Theorem 2.3.7 in \citet{gine2016mathematical} essentially ensures the separability of $(T,d)$.
\end{proof}
    By invoking Lemma~\ref{lemma: dudley's entropy integral} and substituting the decomposition from~\eqref{eq: Esta < complexity}, we obtain the following upper bound for the stochastic error:
\begin{align}\label{eq: Esta < entropy}
    &\mathbb{E}_{\widetilde{\mathcal{D}}_S} \bigl\{ \mathcal{E}_{\mathrm{sta}} \bigr\} \nonumber\\
    &\lesssim \frac{1}{\sqrt{n_S}} \mathbb{E}_{\widetilde{\mathcal{D}}_S} \Biggl\{ \int_0^{B_1} \sqrt{\log 2 \mathcal{N} \bigl( \varrho, \mathcal{NN}_{d,1}(W_1, D_1, B_1)_{\vert_{\{X^{(i)}_{S,1}\}_{i=1}^{n_S}}}, d_2 \bigr)} \, d\varrho \nonumber\\
    &\quad+ \int_0^{B_2} \sqrt{\log 2 \mathcal{N} \bigl( \varrho, \mathcal{NN}_{d^\star,1}(W_2, D_2, B_2)_{\vert_{\{\mathcal{R}^{(i)}\}_{i=1}^{n_S}}}, d_2 \bigr)} \, d\varrho \nonumber\\
    &\quad+ \int_0^{B_2} \sqrt{\log 2 \mathcal{N} \bigl( \varrho, \mathcal{NN}_{d,1}(\max\{W_1, W_2\}, D_1 + D_2, B_2)_{\vert_{\{X_{S,1}^{(i)}\}_{i=1}^{n_S}}}, d_2 \bigr)} \, d\varrho \Biggr\}.
\end{align}
We illustrate the bound for the first term in~\eqref{eq: Esta < complexity}. Given that $f \in \widehat{\mathcal{F}} \Rightarrow f_j \in \mathcal{NN}_{d,1}(W_1, D_1, B_1)$ for each $1 \leq j \leq d^\star$, and applying the Fubini theorem, we have: 
\begin{align*}
    &\mathbb{E}_{\widetilde{\mathcal{D}}_S, \bm{\xi}} \biggl\{ \sup_{f \in \widehat{\mathcal{F}}} \Bigl\vert \frac{1}{n_S} \sum_{i=1}^{n_S} \sum_{j=1}^{d^\star} \xi_{i,j} f_j(X_{S,1}^{(i)}) \Bigr\vert \biggr\} \\
    &\leq d^\star \mathbb{E}_{\widetilde{\mathcal{D}}_S} \Biggl[ \mathbb{E}_{\bm{\xi}} \biggl\{ \sup_{f \in \mathcal{NN}_{d,1}(W_1, D_1, B_1)} \Bigl\vert \frac{1}{n_S} \sum_{i=1}^{n_S} \xi_{i} f(X_{S,1}^{(i)}) \Bigr\vert \biggr\} \Biggr] \\
    &= d^\star \mathbb{E}_{\widetilde{\mathcal{D}}_S} \Biggl[ \mathbb{E}_{\bm{\xi}} \biggl\{ \sup_{f \in \mathcal{NN}_{d,1}(W_1, D_1, B_1)} \Bigl\vert \frac{1}{n_S} \sum_{i=1}^{n_S} \xi_{i} f(X_{S,1}^{(i)}) \Bigr\vert \Bigm\vert X_{S,1}^{(i)}, 1 \leq i \leq n_S \biggr\} \Biggr].    
\end{align*}
Therefore, it suffices to establish that
\begin{align*}
    &\mathbb{E}_{\bm{\xi}} \biggl\{ \sup_{f \in \mathcal{NN}_{d,1}(W_1, D_1, B_1)} \Bigl\vert \frac{1}{\sqrt{n_S}} \sum_{i=1}^{n_S} \xi_{i} f(X_{S,1}^{(i)}) \Bigr\vert \Bigm\vert \widetilde{\mathcal{D}}_S \biggr\} \\
    &\lesssim \int_0^{B_1} \sqrt{\log 2 \mathcal{N} \bigl( \varrho, \mathcal{NN}_{d,1}(W_1, D_1, B_1)_{\vert_{\{X^{(i)}_{S,1}\}_{i=1}^{n_S}}}, d_2 \bigr)} \, d\varrho.
\end{align*}
Specifically, conditioned on $\widetilde{\mathcal{D}}_S$, the samples $\{X_{S,1}^{(i)}\}_{i=1}^{n_S}$ are fixed. Consequently, the stochastic process $\bigl\{ \frac{1}{\sqrt{n_S}} \sum_{i=1}^{n_S} \xi_{i} f(X_{S,1}^{(i)}) : f \in \mathcal{NN}_{d,1}(W_1, D_1, B_1) \bigr\}$ is sub-Gaussian, as $\xi_i$ are independent Rademacher variables. Let $f_{\vert_{\{X_{S,1}^{(i)}\}}} = \bigl( f(X_{S,1}^{(1)}), \dots, f(X_{S,1}^{(n_S)}) \bigr) \in \mathbb{R}^{n_S}$ for any $f \in \mathcal{NN}_{d,1}$. Defining the distance on the index set $\mathcal{NN}_{d,1}$ as
\begin{align*}
    d_{\mathcal{NN}}(f_1, f_2) &:= \sqrt{\mathbb{E} \Bigl[ \Bigl\{ \frac{1}{\sqrt{n_S}} \sum_{i=1}^{n_S} \xi_i \bigl(f_1(X_{S,1}^{(i)})-f_2(X_{S,1}^{(i)})\bigr) \Bigr\}^2 \Bigr]} \\
    &= \sqrt{\frac{1}{n_S} \sum_{i=1}^{n_S} \bigl( f_1(X_{S,1}^{(i)}) - f_2(X_{S,1}^{(i)}) \bigr)^2} = d_2 \bigl( {f_1}_{\vert_{\{X_{S,1}^{(i)}\}_{i=1}^{n_S}}}, {f_2}_{\vert_{\{X_{S,1}^{(i)}\}_{i=1}^{n_S}}} \bigr),
\end{align*}
we note that $\bigl( \mathcal{NN}_{d,1}(W_1, D_1, B_1)_{\vert_{\{X^{(i)}_{S,1}\}_{i=1}^{n_S}}}, d_2 \bigr)$ is a separable subset of $\mathbb{R}^{n_S}$, thereby satisfying the requirements of Lemma~\ref{lemma: dudley's entropy integral}. Setting $t_0$ as the zero-parameter network $f_{\bm{0}}$ yields $\mathbb{E} \lvert X(t_0) \rvert = 0$. Furthermore, for any $f \in \mathcal{NN}_{d,1}$, we have:
\begin{equation*}
    d_{\mathcal{NN}}(f, f_{\bm{0}}) = d_2 \bigl( f_{\vert_{\{X_{S,1}^{(i)}\}}}, {f_{\bm{0}}}_{\vert_{\{X_{S,1}^{(i)}\}}} \bigr) = \sqrt{\frac{1}{n_S} \sum_{i=1}^{n_S} f^2(X_{S,1}^{(i)})} \leq B_1,
\end{equation*}
which implies that the diameter $D$ satisfies $D/2 \leq B_1$. The bounds for the remaining terms in~\eqref{eq: Esta < entropy} follow from an identical argument.

We now introduce the following definitions and auxiliary results regarding the capacity of neural network function classes to conclude the bound for \eqref{eq: Esta < entropy}.

\begin{definition}[VC-dimension] 
    Let $\mathcal{H}$ denote a class of functions from $\mathcal{X}$ to $\{0,1\}$. For any non-negative integer $m$, the growth function of $\mathcal{H}$ is defined as
    \begin{equation*}
        \Pi_\mathcal{H}(m) := \max_{x_1, \ldots, x_m \in \mathcal{X}} \bigl\vert \bigl\{ (h(x_1), \ldots, h(x_m)) : h \in \mathcal{H} \bigr\} \bigr\vert.
    \end{equation*}
    If $\bigl\lvert \bigl\{ (h(x_1), \ldots, h(x_m)) : h \in \mathcal{H} \bigr\} \bigr\rvert = 2^m$, we say that $\mathcal{H}$ shatters the set $\{x_1, \ldots, x_m\}$. The Vapnik--Chervonenkis dimension $\mathrm{VCdim}(\mathcal{H})$ is the size of the largest shattered set. For a real-valued function class $\mathcal{H}$, we define $\mathrm{VCdim}(\mathcal{H}) := \mathrm{VCdim}(\mathrm{sgn}(\mathcal{H}))$, where $\mathrm{sgn}(x) = \1\{x > 0\}$.
\end{definition}

\begin{definition}[Pseudodimension]
    Let $\mathcal{H}$ be a class of functions from $\mathcal{X}$ to $\mathbb{R}$. The pseudodimension $\mathrm{Pdim}(\mathcal{H})$ is the largest integer $m$ for which there exists a witness $(x_1, \dots, x_m, y_1, \dots, y_m) \in \mathcal{X}^m \times \mathbb{R}^m$ such that for any $(b_1, \dots, b_m) \in \{0,1\}^m$, there exists $f \in \mathcal{H}$ satisfying $f(x_i) > y_i \Leftrightarrow b_i = 1$ for all $i$.
\end{definition}

\begin{lemma}[\citet{anthony1999neural}]\label{lemma: covering number < Pdim}
    Let $\mathcal{H}$ be a class of real functions into the bounded interval $[0, B]$, with $v=\mathrm{Pdim}(\mathcal{H})\geq1$. For $0<\varrho\leq B$ and $m\geq v$, the uniform covering number satisfies
    \begin{equation*}
        \mathcal{N}_\infty(\varrho, \mathcal{H}, m)
        \leq\sum_{i=0}^{v}\binom{m\lceil B/\varrho\rceil}{i}
        \leq\left(\frac{2emB}{\varrho v}\right)^v.
    \end{equation*}
    This follows by discretising the range and applying the subgraph growth bound. For functions with range $[-B,B]$, shifting by $B$ gives the same bound with $B$ replaced by $2B$.
\end{lemma}

\begin{lemma}[\citet{bartlett2019nearly}]\label{lemma: Pdim < DSlogS}
    Let $S$ denote the total number of parameters in $\mathcal{NN}_{d,1}(W, D)$. Then:
    \begin{equation*}
        \mathrm{Pdim} \bigl( \mathcal{NN}_{d,1}(W, D) \bigr) \lesssim DS \log S.
    \end{equation*}
\end{lemma}

\begin{lemma}\label{lemma: Pdim(NN(W,D,B)) < Pdim(NN(W,D))}
    For any $d, W, D \in \N$, the range constraint does not increase the pseudodimension:
    \begin{equation*}
        \mathrm{Pdim} \bigl( \mathcal{NN}_{d,1}(W, D, B) \bigr) \leq \mathrm{Pdim} \bigl( \mathcal{NN}_{d,1}(W, D) \bigr).
    \end{equation*}
\end{lemma}
    Following the preliminaries established above, we now proceed with the formal derivation of the stochastic error bound in \eqref{eq: Esta < entropy}:

\begin{align*}
   &\mathbb{E}_{\widetilde{\mathcal{D}}_S} \bigl\{ \mathcal{E}_{\mathrm{sta}} \bigr\} \nonumber \\
   &\lesssim \frac{1}{\sqrt{n_S}} \mathbb{E}_{\widetilde{\mathcal{D}}_S} \Biggl[ \int_0^{B_1} \sqrt{\log 2 \mathcal{N} \bigl( \varrho, \mathcal{NN}_{d,1}(W_1, D_1, B_1)_{\vert_{\{X^{(i)}_{S,1}\}_{i=1}^{n_S}}}, d_2 \bigr)} \, d\varrho \nonumber\\
    &\quad+ \int_0^{B_2} \sqrt{\log 2 \mathcal{N} \bigl( \varrho, \mathcal{NN}_{d^\star,1}(W_2, D_2, B_2)_{\vert_{\{\mathcal{R}^{(i)}\}_{i=1}^{n_S}}}, d_2 \bigr)} \, d\varrho \nonumber\\
    &\quad+ \int_0^{B_2} \sqrt{\log 2 \mathcal{N} \bigl( \varrho, \mathcal{NN}_{d,1}(\max\{W_1, W_2\}, D_1 + D_2, B_2)_{\vert_{\{X_{S,1}^{(i)}\}_{i=1}^{n_S}}}, d_2 \bigr)} \, d\varrho \Biggr] \\
    &\leq \frac{1}{\sqrt{n_S}} \mathbb{E}_{\widetilde{\mathcal{D}}_S} \Biggl[ \int_0^{B_1} \sqrt{\log 2 \mathcal{N} \bigl( \varrho, \mathcal{NN}_{d,1}(W_1, D_1, B_1)_{\vert_{\{X^{(i)}_{S,1}\}_{i=1}^{n_S}}}, d_\infty \bigr)} \, d\varrho \nonumber\\
    &\quad+ \int_0^{B_2} \sqrt{\log 2 \mathcal{N} \bigl( \varrho, \mathcal{NN}_{d^\star,1}(W_2, D_2, B_2)_{\vert_{\{\mathcal{R}^{(i)}\}_{i=1}^{n_S}}}, d_\infty \bigr)} \, d\varrho \nonumber\\
    &\quad+ \int_0^{B_2} \sqrt{\log 2 \mathcal{N} \bigl( \varrho, \mathcal{NN}_{d,1}(\max\{W_1, W_2\}, D_1 + D_2, B_2)_{\vert_{\{X_{S,1}^{(i)}\}_{i=1}^{n_S}}}, d_\infty \bigr)} \, d\varrho \Biggr] \tag{Lemma~\ref{lemma: covering metric hierarchy}}\\
    &\leq \frac{1}{\sqrt{n_S}} \Biggl\{ \int_0^{B_1} \sqrt{\log 2 \mathcal{N}_\infty \bigl( \varrho, \mathcal{NN}_{d,1}(W_1, D_1, B_1), n_S \bigr)} \, d\varrho \nonumber\\
    &\quad+ \int_0^{B_2} \sqrt{\log 2 \mathcal{N}_\infty \bigl( \varrho, \mathcal{NN}_{d^\star,1}(W_2, D_2, B_2), n_S \bigr)} \, d\varrho \nonumber\\
    &\quad+ \int_0^{B_2} \sqrt{\log 2 \mathcal{N}_\infty \bigl( \varrho, \mathcal{NN}_{d,1}(\max\{W_1, W_2\}, D_1 + D_2, B_2), n_S \bigr)} \, d\varrho \Biggr\} \tag{Definition~\ref{def: uniform covering number}}\\
    &\lesssim \biggl( \frac{\mathrm{Pdim}(\mathcal{NN}_{d,1}(W_1, D_1)) \log n_S}{n_S} \biggr)^{1/2} + \biggl( \frac{\mathrm{Pdim}(\mathcal{NN}_{d^\star,1}(W_2, D_2)) \log n_S}{n_S} \biggr)^{1/2} \nonumber\\
    &\quad+ \biggl( \frac{\mathrm{Pdim}(\mathcal{NN}_{d,1}(\max\{W_1,W_2\}, D_1+D_2)) \log n_S}{n_S} \biggr)^{1/2} \tag{Lemma~\ref{lemma: covering number < Pdim} and \ref{lemma: Pdim(NN(W,D,B)) < Pdim(NN(W,D))}}\\
    &\leq \widetilde{\mathcal O} \Biggl( \sqrt{\frac{(D_1 + D_2)^2 \max\{W_1, W_2\}^2}{n_S}} \Biggr) \tag{Lemma~\ref{lemma: Pdim < DSlogS} and $S \lesssim W^2D$}\\
    &\lesssim \frac{D_1 W_1}{\sqrt{n_S}} \tag{$W_2 \lesssim W_1$ and $D_2 \lesssim D_1$}.
\end{align*}

Logarithmic factors are suppressed when recording the polynomial rate.

\subsection{The Approximation Error}\label{section: The Approximation Error}

In this section, we determine the upper bounds for the approximation errors $\mathcal{E}_{\mathcal{F}}$ and $\mathcal{E}_{\mathcal{G}}$, following the analytical framework established in \citet{yang2023nearly} and \citet{gao2024convergencecontinuousnormalizingflows}. To facilitate this analysis, we first introduce several foundational definitions and lemmas regarding Sobolev spaces and neural network approximation capacity. Let $d \in \N$ and $U$ be an open subset of $\mathbb{R}^d$. We denote $L^\infty(U)$ as the standard Lebesgue space on $U$ equipped with the $L^\infty$ norm.

\begin{definition}[Sobolev space]
    For $n \in \{0\} \cup \N$, the Sobolev space $W^{n, \infty}(U)$ is defined as
    \begin{align*}
        W^{n, \infty}(U) := \bigl\{ f \in L^\infty(U) : D^{\bm{\alpha}} f \in L^{\infty}(U) \text{ for all } \bm{\alpha} \in \N_0^{d} \text{ with } \norm{\bm{\alpha}}_1 \leq n \bigr\}.
    \end{align*}
    Here $\N_0=\{0\}\cup\N$. Moreover, for any $f \in W^{n, \infty}(U)$, the Sobolev norm $\norm{\cdot}_{W^{n, \infty}(U)}$ is defined by
    \begin{align*}
        \norm{f}_{W^{n, \infty}(U)} := \max_{0 \leq \norm{\bm{\alpha}}_1 \leq n} \norm{D^{\bm{\alpha}} f}_{L^\infty(U)}.
    \end{align*}
\end{definition}

The following lemma provides a characterization of Lipschitz continuity within the context of Sobolev spaces.
\begin{lemma}[\citet{evans2010partial}]
    \label{lemma: lipschitz v.s sobolev}
    Let $U$ be an open and bounded set with a $C^1$ boundary $\partial U$. Then $f : U \to \mathbb{R}$ belongs to $W^{1, \infty}(U)$ if and only if it admits a Lipschitz representative.
\end{lemma}

To bound the approximation error, we utilize the following result on the approximation power of ReLU networks for functions with Lipschitz regularity.
\begin{lemma}[Corollary B.2 in \citet{gao2024convergencecontinuousnormalizingflows}]
    \label{lemma: approximation with lipschitz regularity}
    For any $f \in W^{1, \infty}((0, 1)^d)$ such that $\norm{f}_{W^{1, \infty}((0,1)^d)} < \infty$, and for any $N, D \in \N$, there exists a function $f_{\bm{\theta}}$ implemented by a deep ReLU network with width $\widetilde{\mathcal{O}}(N)$, depth $\widetilde{\mathcal{O}}(D)$, and range constraint $B \geq \norm{f}_\infty$ such that $\norm{f_{\bm{\theta}}}_{W^{1, \infty}((0, 1)^d)} \lesssim \norm{f}_{W^{1, \infty}((0,1)^d)}$ and
    \begin{align*}
        \norm{f_{\bm{\theta}} - f}_{L^\infty([0,1]^d)} \lesssim \norm{f}_{W^{1, \infty}((0,1)^d)} (ND)^{-2/d}.
    \end{align*}
    Here we write $a_n = \widetilde{\mathcal{O}}(b_n)$ for two sequences $\{a_n\}$ and $\{b_n\}$ if there exists a constant $k \geq 0$ such that $a_n = \mathcal{O} \bigl( b_n \log^k n \bigr)$.
\end{lemma}

\paragraph{The Encoder Approximation Error \texorpdfstring{$\mathcal{E}_{\mathcal{F}}$}{EF}}\label{subsection: encoder approximation error}

Apply Lemma~\ref{lemma: approximation with lipschitz regularity} coordinatewise
to $f^\star$. For sufficiently large fixed $B_1$ and $\bar L$, the resulting
coordinate networks can be concatenated into an
$f\in\mathcal{NN}_{d,d^\star}(W_1,D_1,B_1)\cap\mathrm{Lip}(\bar L)$ satisfying
\begin{equation}\label{eq: encoder uniform approximation}
    \sup_{\bm x\in\mathcal X}
    \norm{f(\bm x)-f^\star(\bm x)}_2
    \leq\widetilde{\mathcal O}\{(D_1W_1)^{-2/d}\}.
\end{equation}
Choose the fixed constant in $\Delta_{n_S}$ large enough to dominate the
right-hand side. Since $\norm{f^\star(\bm x)}_2=R$, the reverse triangle
inequality then gives
\[
R-\Delta_{n_S}\leq\norm{f(\bm x)}_2
\leq R+\Delta_{n_S},\qquad \bm x\in\mathcal X.
\]
Thus this approximant belongs to the annular class $\widehat{\mathcal F}$.

For this $f$, the difference of the alignment terms is bounded using
$|\norm{\bm u}_2^2-\norm{\bm v}_2^2|
\leq(\norm{\bm u}_2+\norm{\bm v}_2)\norm{\bm u-\bm v}_2$.
Moreover, the triangle inequality for $\mathcal W$ and the coupling obtained
from the same $(X_S,A)$ give
\[
\left|\mathcal W(\P_f,\P_{\mathcal R})
-\mathcal W(\P_{f^\star},\P_{\mathcal R})\right|
\leq\mathcal W(\P_f,\P_{f^\star})
\leq\E\norm{f(A(X_S))-f^\star(A(X_S))}_2.
\]
Because $\norm{f(\bm x)}_2\leq R_{2,n_S}$ and
$\norm{f^\star(\bm x)}_2=R$, these two bounds and
\eqref{eq: encoder uniform approximation} yield
\begin{equation}\label{eq: encoder approximation rate}
    \mathcal E_{\mathcal F}
    \leq\widetilde{\mathcal O}\{(D_1W_1)^{-2/d}\}.
\end{equation}

\paragraph{The Critic Approximation Error \texorpdfstring{$\mathcal{E}_{\mathcal{G}}$}{EG}}\label{subsection: critic approximation error}

We bound the critic approximation error $\mathcal{E}_{\mathcal{G}}$ using integral probability metrics (IPMs), following \citet{liu2021non}.

\begin{definition}[Integral Probability Metric, \citet{Muller_1997}]\label{def: IPM}
    For any two probability distributions $\mu$ and $\nu$ defined on a common domain, and a symmetric function class $\mathcal{H}$, the Integral Probability Metric (IPM) is defined as
    \begin{align*}
        d_{\mathcal{H}}(\mu, \nu) = \sup_{h \in \mathcal{H}} \Bigl[ \mathbb{E}_{X_1 \sim \mu} \bigl\{ h(X_1) \bigr\} - \mathbb{E}_{X_2 \sim \nu} \bigl\{ h(X_2) \bigr\} \Bigr].
    \end{align*}
    \begin{remark}
         Notably, when the function class $\mathcal{H}$ is chosen as the set of $1$-Lipschitz functions (i.e., $\mathcal{H} = \mathrm{Lip}(1)$), the IPM $d_{\mathcal{H}}(\mu, \nu)$ coincides exactly with the Mallows distance $\mathcal{W}(\mu, \nu)$.
    \end{remark}
\end{definition}

To quantify the discrepancy introduced by using a restricted hypothesis class, we define the approximation error between two function classes as follows.

\begin{definition}[Approximation error between function classes]\label{def: function class approx error}
    The approximation error of a function class $\mathcal{H}_1$ with respect to a target class $\mathcal{H}_2$ is given by
    \begin{align*}
        \mathcal{E}(\mathcal{H}_2, \mathcal{H}_1) = \sup_{h_2 \in \mathcal{H}_2} \inf_{h_1 \in \mathcal{H}_1} \norm{h_2 - h_1}_\infty.
    \end{align*}
\end{definition}

The following lemma bounds the IPM difference by this approximation error.

\begin{lemma}\label{lemma: difference between IPMs < 2 approx error}
For any probability distributions $\mu$ and $\nu$, and symmetric function classes $\mathcal{H}_1$ and $\mathcal{H}_2$, the difference between their respective IPMs satisfies the inequality:
\begin{equation*}
    d_{\mathcal{H}_2}(\mu, \nu) - d_{\mathcal{H}_1}(\mu, \nu) \leq 2 \mathcal{E}(\mathcal{H}_2, \mathcal{H}_1).
\end{equation*}
\end{lemma}
\begin{proof}
     By decomposing the supremum over $\mathcal{H}_2$ and $\mathcal{H}_1$, we obtain:
     \begin{align*}
        &d_{\mathcal{H}_2}(\mu, \nu) - d_{\mathcal{H}_1}(\mu, \nu) \\
        &= \sup_{h_2 \in \mathcal{H}_2} \Bigl[ \mathbb{E}_{X_1 \sim \mu} \bigl\{ h_2(X_1) \bigr\} - \mathbb{E}_{X_2 \sim \nu} \bigl\{ h_2(X_2) \bigr\} \Bigr] \\
        &\quad - \sup_{h_1 \in \mathcal{H}_1} \Bigl[ \mathbb{E}_{X_1 \sim \mu} \bigl\{ h_1(X_1) \bigr\} - \mathbb{E}_{X_2 \sim \nu} \bigl\{ h_1(X_2) \bigr\} \Bigr] \\
        &= \sup_{h_2 \in \mathcal{H}_2} \inf_{h_1 \in \mathcal{H}_1} \Bigl[ \mathbb{E}_{X_1 \sim \mu} \bigl\{ h_2(X_1) - h_1(X_1) \bigr\} + \mathbb{E}_{X_2 \sim \nu} \bigl\{ h_1(X_2) - h_2(X_2) \bigr\} \Bigr] \\
        &\leq 2 \sup_{h_2 \in \mathcal{H}_2} \inf_{h_1 \in \mathcal{H}_1} \norm{h_2 - h_1}_\infty = 2 \mathcal{E}(\mathcal{H}_2, \mathcal{H}_1).
    \end{align*}
\end{proof}

Because $\widehat{\mathcal G}\subseteq\mathcal G$, its IPM cannot exceed the
IPM over $\mathcal G$. By invoking
Lemma~\ref{lemma: difference between IPMs < 2 approx error}, the task of
bounding $\mathcal{E}_{\mathcal{G}}$ is reduced to the following corollary.

\begin{corollary}\label{corollary: critic approx bound}
    The critic approximation error satisfies the bound $\mathcal{E}_{\mathcal{G}} \leq 2\lambda \mathcal{E}(\mathcal{G}, \widehat{\mathcal{G}})$, where $\widehat{\mathcal{G}}$ is the sieve hypothesis class for the critic.
\end{corollary}

Concretely, we use the sieve approximation condition stated in
Section~\ref{subsection: sample complexity}, with
$\widehat{\mathcal G}\subseteq\mathcal G$. Lemma~\ref{lemma: approximation with lipschitz regularity}
provides an approximation rate with Lipschitz regularity controlled up to a
fixed multiplicative constant. Membership in the prescribed Lipschitz ball
is imposed by the sieve condition. Under this condition, because $\lambda$
is fixed, $d^\star\leq d$ and
$D_2W_2\asymp D_1W_1$,
\begin{align*}
    \mathcal{E}_{\mathcal{G}} \lesssim (D_2 W_2)^{-2/d^\star}
    \lesssim (D_2 W_2)^{-2/d}
    \lesssim (D_1 W_1)^{-2/d}.
\end{align*}

\subsection{Trade-off Analysis}\label{subsection: Trade off between statistic error and approximation error}

By balancing the stochastic error $\mathcal{E}_{\mathrm{sta}}$ and the approximation components $\mathcal{E}_{\mathcal{F}}, \mathcal{E}_{\mathcal{G}}$ with $D_1 W_1 \asymp n_S^{\frac{d}{2d+4}}$ and $D_2 W_2 \asymp D_1 W_1$, we obtain the following convergence rate for the total population risk, up to logarithmic factors:
\begin{align*}
    \mathbb{E}_{\widetilde{\mathcal{D}}_S} \bigl\{ \mathcal{L}(\hat{f}_{n_S}) \bigr\} \lesssim \mathcal{L}(f^\star) + \frac{D_1 W_1}{\sqrt{n_S}} + (D_1 W_1)^{-2/d} \lesssim \mathcal{L}(f^\star) + n_S^{-\frac{1}{d+2}}.
\end{align*}

\subsection{Existence of a Vanishing Population Risk}\label{subsection: vanish L(f*)}

This subsection demonstrates the existence of an ideal encoder $\tilde{f} \in \mathcal{F}$ that satisfies $\mathcal{L}(\tilde{f}) = 0$. Consequently, the population optimum satisfies $\mathcal{L}(f^\star) = 0$ by the definition of $f^\star$, providing the necessary foundation for the end-to-end theoretical guarantee. To this end, we first recall the Kirszbraun theorem \citep{schwartz1969nonlinear}.

\begin{lemma}[Kirszbraun theorem]\label{lemma: kirszbraun theorem}
    If $U$ is a subset of a Hilbert space $\mathcal{H}_1$, and $\mathcal{H}_2$ is another Hilbert space, then any Lipschitz-continuous map $f: U \to \mathcal{H}_2$ can be extended to a Lipschitz-continuous map $F: \mathcal{H}_1 \to \mathcal{H}_2$ that preserves the same Lipschitz constant as $f$.
\end{lemma}

\begin{lemma}[Existence of a vanishing population risk]\label{lem:vanishing}
Under Assumption~\ref{ass: realizable}, and provided that the Lipschitz constant of the composite map does not exceed the constant defining $\mathcal F$, there exists $\tilde f\in\mathcal F$
such that $\mathcal L(\tilde f)=0$. Consequently $\mathcal L(f^\star)=0$.
\end{lemma}

\begin{proof}
For the given $\mathcal A=\mathcal A_{n_S}$, the construction follows a two-stage scheme $\tilde f=T\circ\phi$: invariance of $\phi$ makes the alignment loss zero, while the componentwise Lipschitz maps $T_k$ match the source components to their reference components. The maps need not be optimal.

\medskip
\noindent\emph{Step 1 (Invariant map zeroes the alignment loss).}
By the realizable transport geometry assumption, $\phi$ is Lipschitz and augmentation-invariant. Hence, for any $\bm{x}\in\mathcal X$ and any
$A_1,A_2\in\mathcal A$,
\[
\norm{\phi(A_1(\bm{x}))-\phi(A_2(\bm{x}))}_2=\norm{\phi(\bm{x})-\phi(\bm{x})}_2=0 .
\]

\medskip
\noindent\emph{Step 2 (Gluing the componentwise transports).}
Use the reference-component labelling specified after Assumption~\ref{ass: realizable},
and write $S_k:=\phi(C_S(k))\subset\mathbb R^{d^\star}$. By that assumption,
the sets $\{S_k\}_{k=1}^K$ are pairwise
$\rho$-separated. Define
\[
T:=\bigsqcup_{k=1}^{K}T_k,\qquad T(z)=T_k(\bm{z})\ \text{ for } \bm{z}\in S_k,
\]
which is well defined since the $S_k$ are disjoint. We verify that $T$ is
Lipschitz on $\bigsqcup_k S_k$:
\begin{itemize}
  \item \emph{(Within a component)} for $\bm{z},\bm{z}'\in S_k$,
  $\norm{T(\bm{z})-T(\bm{z}')}_2\le \mathrm{Lip}(T_k)\norm{\bm{z}-\bm{z}'}_2$;
  \item \emph{(Across components)} for $\bm{z}\in S_i$, $\bm{z}'\in S_j$ with $i\neq j$,
  using $\norm{T(\bm{z})-T(\bm{z}')}_2\le\mathrm{diam}\bigl(\mathrm{supp}(\P_\mathcal{R})\bigr)\le 2R$
  and $\|\bm{z}-\bm{z}'\|_2\ge\rho$,
  \[
  \frac{\|T(\bm{z})-T(\bm{z}')\|_2}{\|\bm{z}-\bm{z}'\|_2}\;\le\;\frac{2R}{\rho}.
  \]
\end{itemize}
Hence $T$ is Lipschitz on $\bigsqcup_k S_k$: the cross-component bound is
supplied by the separation $\rho$, independently of whether each $T_k$ is
optimal. Since $\{C_S(k)\}_{k=1}^K$ is a partition of $\mathcal X$,
$\bigsqcup_{k=1}^K S_k=\phi(\mathcal X)$; moreover,
$T\{\phi(\mathcal X)\}\subseteq\partial B(\bm0,R)$ because every $T_k$ maps
into the support of a reference component.

\medskip
\noindent\emph{Step 3 (Kirszbraun extension).}
By Lemma~\ref{lemma: kirszbraun theorem}, the map constructed in Step~2 extends to
$T:\mathbb R^{d^\star}\to\mathbb R^{d^\star}$ with the same Lipschitz
constant $L_T$. Its values off $\phi(\mathcal X)$ are immaterial.

\medskip
\noindent\emph{Step 4 (Composition and verification).}
Set $\tilde f:=T\circ\phi$.
\begin{itemize}
  \item \emph{(Membership in $\mathcal F$.)} As a composition of Lipschitz
  maps, $\tilde f$ is naturally Lipschitz continuous as well. Since
  $T\{\phi(\mathcal X)\}\subseteq\partial B(\bm0,R)$,
  $\norm{\tilde f(\bm x)}_2=R$ on $\mathcal X$, so
  $\tilde f\in\mathcal F$.
  \item \emph{(Alignment loss.)} By Step~1,
  $\tilde f(A_1(X))=T(\phi(A_1(X)))=T(\phi(X))=\tilde f(A_2(X))$, implying $\mathcal{L}_{\mathrm{align}}(\tilde f)=0$.
  \item \emph{(Measure matching.)} Augmentation invariance gives
  $\phi_\sharp \P_{\mathcal A}=\phi_\sharp \P_S$, so
  \[
  \tilde f_\sharp \P_{\mathcal A}
  =T_\sharp\,\phi_\sharp \P_S
  =\sum_{k=1}^{K}p_S(k)\,(T_k)_\sharp\nu_k
  =\sum_{k=1}^{K}p_S(k)\P_\mathcal{R}(k).
  \]
  Choosing the design weights $\alpha_k=p_S(k)$, the right-hand side equals
  $\P_\mathcal{R}$, and therefore $\mathcal{W}(\P_{\tilde f},\P_\mathcal{R})=0$.
\end{itemize}

Combining the last two items,
\[
\mathcal{L}(\tilde f)=\mathcal{L}_{\mathrm{align}}(\tilde f)+\lambda\,\mathcal{W}(\P_{\tilde f},\P_{\mathcal R})=0 .
\]
Since $f^\star\in\argmin_{f\in\mathcal F}\mathcal L(f)$ and
$\mathcal L\ge 0$, we conclude $\mathcal L(f^\star)=0$.
\end{proof}

\begin{mythm}{2}
Under the assumptions of Theorem~\ref{theorem: sample theorem}, the displayed end-to-end risk bound holds.
\end{mythm}

\begin{proof}
    We prove the bound at the given sufficiently large $n_S$. Probabilities and expectations involving the target training sample are conditional on the observed class counts $\{n_T(k)\}_{k=1}^K$. We suppress this conditioning in the intermediate steps and display it explicitly in the final bound.
    \phantomsection\label{proof: sample theorem}
    The local identifiability condition in Theorem~\ref{theorem: sample theorem} means that there is a fixed threshold $c_0>0$, independent of $n_S$, such that Assumption~\ref{assumption: transport mass} holds with a common $C_{\mathrm{id}}$ whenever $f\in\widehat{\mathcal F}$ and $\mathcal L(f)\leq c_0$. Since the population DM risk is nonnegative, Markov's inequality gives
    \[
        \P_{\widetilde{\mathcal D}_S}\{\mathcal L(\hat f_{n_S})>c_0\}
        \leq c_0^{-1}\E_{\widetilde{\mathcal D}_S}\{\mathcal L(\hat f_{n_S})\}.
    \]
    By the annular encoder restriction,
    $R_{1,n_S}\leq\norm{\hat{f}_{n_S}(\bm{x})}_2\leq R_{2,n_S}$ on
    $\mathcal X$. On $\{\mathcal L(\hat f_{n_S})\leq c_0\}$, the local identifiability condition allows us to apply Theorem~\ref{theorem: general version of the pop theorem} to the fitted encoder, with a common constant. On the complementary event, Cauchy--Schwarz gives $\max_{i\neq j}|\bm\mu_T(i)^\top\bm\mu_T(j)|\leq R_{2,n_S}^2$. Splitting the expectation over these two events and using the preceding probability bound, we obtain
    \begin{align}\label{eq: E[divergence] < E[L]}
        \mathbb{E}_{\widetilde{\mathcal{D}}_S} \Bigl\{ \max_{i \neq j} \lvert \bm\mu_T(i)^{\top} \bm\mu_T(j) \rvert \Bigr\} \lesssim \mathbb{E}_{\widetilde{\mathcal{D}}_S} \bigl\{ \mathcal{L}(\hat{f}_{n_S}) \bigr\} + \epsilon_1.
    \end{align}
    Here $c_0$ is fixed and $R_{2,n_S}$ is uniformly bounded for sufficiently large $n_S$, so the complementary event changes only the implicit constant. The augmentation-stability bound \eqref{eq: U_T < L} does not require Assumption~\ref{assumption: transport mass} and remains valid for every encoder in the sieve.
    Let $\mathcal{E}$ denote the event $\bigl\{ \max_{i \neq j} \lvert \bm\mu_T(i)^\top \bm\mu_T(j) \rvert < R_{2,n_S}^2 \psi(\sigma_{n_S}, \delta_{n_S}, \eps, \hat{f}_{n_S}) \bigr\}$. This event depends on the source sample through the fitted encoder and on the target sample through the empirical centres entering $\psi$. Here $\mathbb{P}$ denotes the joint probability law of these samples. We decompose the classification risk as:
    \begin{align}\label{eq: E[Err(G)] < (1 - sigma) + E[L] + P}
        \mathbb{E}_{\widetilde{\mathcal{D}}_S, \widetilde{\mathcal{D}}_T} \bigl\{ \mathrm{Err}(G_{\hat{f}_{n_S}}) \bigr\} &= \mathbb{E}_{\widetilde{\mathcal{D}}_S, \widetilde{\mathcal{D}}_T} \bigl\{ \mathrm{Err}(G_{\hat{f}_{n_S}}) \1_{\mathcal{E}} \bigr\} + \mathbb{E}_{\widetilde{\mathcal{D}}_S, \widetilde{\mathcal{D}}_T} \bigl\{ \mathrm{Err}(G_{\hat{f}_{n_S}}) \1_{\mathcal{E}^c} \bigr\} \nonumber \\
        &\leq \mathbb{E}_{\widetilde{\mathcal{D}}_S, \widetilde{\mathcal{D}}_T} \Bigl[ \bigl\{ (1 - \sigma_{n_S}) + U_T(\eps, \hat{f}_{n_S}) \bigr\} \1_{\mathcal{E}} \Bigr] + \mathbb{P} \bigl( \mathcal{E}^c \bigr) \nonumber \\
        &\leq (1 - \sigma_{n_S}) + \mathbb{E}_{\widetilde{\mathcal{D}}_S} \bigl\{ U_T(\eps, \hat{f}_{n_S}) \bigr\} + \mathbb{P} \bigl( \mathcal{E}^c \bigr) \nonumber \\
        &\leq (1 - \sigma_{n_S}) + \mathcal{O} \biggl( \eps^{-1} \mathbb{E}_{\widetilde{\mathcal{D}}_S} \Bigl[ \bigl\{ \mathcal{L}(\hat{f}_{n_S}) + \epsilon_1 + \epsilon_2 \bigr\}^{\frac{1}{2}} \Bigr] \biggr) + \mathbb{P} \bigl( \mathcal{E}^c \bigr) \nonumber \\
        &\leq (1 - \sigma_{n_S}) + \mathcal{O} \biggl( \eps^{-1} \Bigl[ \mathbb{E}_{\widetilde{\mathcal{D}}_S} \bigl\{ \mathcal{L}(\hat{f}_{n_S}) \bigr\} + \epsilon_1 + \epsilon_2 \Bigr]^{\frac{1}{2}} \biggr) + \mathbb{P} \bigl( \mathcal{E}^c \bigr),
    \end{align}
    where we have utilized Lemma~\ref{lemma: sufficient condition of small Err}, \eqref{eq: U_T < L}, and Jensen's inequality. 

    Substituting the convergence rate $\mathbb{E}_{\widetilde{\mathcal{D}}_S} \{ \mathcal{L}(\hat{f}_{n_S}) \} \lesssim n_S^{-\frac{1}{d+2}}$ into \eqref{eq: E[divergence] < E[L]} and \eqref{eq: E[Err(G)] < (1 - sigma) + E[L] + P}, we obtain:
    \begin{align}\label{eq: derivation_final_err}
        \mathbb{E}_{\widetilde{\mathcal{D}}_S, \widetilde{\mathcal{D}}_T} \bigl\{ \mathrm{Err}(G_{\hat{f}_{n_S}}) \bigr\} \leq (1 - \sigma_{n_S}) + \mathcal{O} \Bigl( \eps^{-1} \bigl( n_S^{-\frac{1}{d+2}} + \epsilon_1 + \epsilon_2 \bigr)^{\frac{1}{2}} \Bigr) + \mathbb{P} \bigl( \mathcal{E}^c \bigr).
    \end{align}
    To bound $\mathbb{P}(\mathcal{E}^c)$, keep the sufficiently small
    $\eps>0$ in the theorem fixed. Recall that
    $R_{1,n_S}=R-\Delta_{n_S}$ and $R_{2,n_S}=R+\Delta_{n_S}$, where
    $\Delta_{n_S}\to0$. For any sequence along which
    $U_T(\eps,\hat f_{n_S})\to0$, the annular branch of
    $\Gamma_{\min}$ in \eqref{eq: psi population threshold} satisfies
    \begin{equation}\label{eq: gamma min limit}
        \frac{R_{1,n_S}}{R_{2,n_S}}\longrightarrow1,
        \qquad
        \frac{\bar d_{\eps}}{R_{2,n_S}}\longrightarrow\frac{2\eps}{R},
        \qquad
        \Gamma_{\min}(\sigma_{n_S},\delta_{n_S},\eps,\hat f_{n_S})
        \longrightarrow1-\frac{2\eps}{R}.
    \end{equation}
    Here the second limit uses $\delta_{n_S}\to0$ and the fact that $\eps$
    is sufficiently small. Moreover, whenever
    $\sigma_{n_S}p_T(k)-U_T(\eps,\hat f_{n_S})>0$, the stable part of class
    $k$ is nonempty. For any point in that set, the population-centre version
    of \eqref{eq: correct score lower bound} and Cauchy--Schwarz give
    \begin{equation}\label{eq: population centre norm lower bound}
        R_{2,n_S}\norm{\bm\mu_T(k)}_2
        \geq \hat f_{n_S}(X_T)^\top\bm\mu_T(k)
        \geq R_{2,n_S}^2
        \Gamma_{\min}(\sigma_{n_S},\delta_{n_S},\eps,\hat f_{n_S}).
    \end{equation}
    Consequently,
    \begin{equation}\label{eq: empirical centre norm lower bound}
        \frac{\min_k\norm{\widehat{\bm\mu}_T(k)}_2}{R_{2,n_S}}
        \geq
        \Gamma_{\min}(\sigma_{n_S},\delta_{n_S},\eps,\hat f_{n_S})
        -\frac{\max_k\norm{\widehat{\bm\mu}_T(k)-\bm\mu_T(k)}_2}
        {R_{2,n_S}}.
    \end{equation}
    The function
    $\gamma-\sqrt{2-2\gamma}-\{1-\gamma^2\}/2$ is continuous and equals
    one at $\gamma=1$. We may therefore first choose $\eps$ sufficiently
    small and then choose fixed constants $u_0,v_0,c_\psi>0$ so that, for this
    sufficiently large $n_S$, the conditions
    \begin{equation*}
        U_T(\eps,\hat f_{n_S})\leq u_0,
        \qquad
        \max_k\norm{\widehat{\bm\mu}_T(k)-\bm\mu_T(k)}_2
        \leq v_0R_{2,n_S}
    \end{equation*}
    imply
    $\psi(\sigma_{n_S},\delta_{n_S},\eps,\hat f_{n_S})\geq c_\psi$.
    This follows directly by substituting \eqref{eq: empirical centre norm lower bound}
    into \eqref{eq: psi population threshold} and using \eqref{eq: gamma min limit}.
    Markov's inequality gives
    \[
        \P\{U_T(\eps,\hat f_{n_S})>u_0\}
        \leq u_0^{-1}\E U_T(\eps,\hat f_{n_S})
        \lesssim \E U_T(\eps,\hat f_{n_S}).
    \]
    It remains to control the target-centre error. Conditional on the observed class counts, the $n_T(k)$ averages of two augmented representations in $\widehat{\bm\mu}_T(k)$ are independent across target observations and have Euclidean norm at most $R_{2,n_S}$. Hence Jensen's inequality and the variance identity give
    \begin{align}
        \E_{\widetilde{\mathcal D}_T}
        \norm{\widehat{\bm\mu}_T(k)-\bm\mu_T(k)}_2
        \leq \frac{C R_{2,n_S}}{\sqrt{n_T(k)}}.
        \label{eq: target centre concentration}
    \end{align}
    Markov's inequality at $v_0R_{2,n_S}$ and a union bound over the fixed
    number $K$ of classes therefore give
    \begin{equation*}
        \P\left\{
        \max_k\norm{\widehat{\bm\mu}_T(k)-\bm\mu_T(k)}_2
        >v_0R_{2,n_S}\right\}
        =\mathcal O\!\left(\frac1{\min_k\sqrt{n_T(k)}}\right).
    \end{equation*}

    When both the augmentation-stability and target-centre error bounds above hold, failure of $\mathcal E$ requires the maximum absolute cross-class inner product to exceed a fixed positive multiple of $R^2$. Markov's inequality, \eqref{eq: U_T < L}, Jensen's inequality, and \eqref{eq: E[divergence] < E[L]} give the first bound below; the second follows from Assumption~\ref{assumption: rates} at the given $n_S$:
    \begin{align*}
        \mathbb{P}(\mathcal{E}^c)
        &\lesssim \bigl\{\E\mathcal L(\hat f_{n_S})+\epsilon_1+\epsilon_2\bigr\}^{1/2}
        +\E\mathcal L(\hat f_{n_S})+\epsilon_1
        +\mathcal O\biggl(\frac1{\min_k\sqrt{n_T(k)}}\biggr)\\
        &\leq \widetilde{\mathcal O}\!\left(n_S^{-\min\{1/(2d+4),\alpha/2,\beta/2\}}\right)
        +\mathcal O\biggl(\frac1{\min_k\sqrt{n_T(k)}}\biggr).
    \end{align*}
    Combining these results gives the end-to-end risk bound at the specified sample size:
    \begin{align*}
        \begin{aligned}
        \mathbb{E}_{\widetilde{\mathcal{D}}_S, \widetilde{\mathcal{D}}_T} \Bigl[ \mathrm{Err}(G_{\hat{f}_{n_S}}) \Bigm| \{n_T(k)\}_{k=1}^K \Bigr]
        &\leq \bigl( 1 - \sigma_{n_S} \bigr) + \widetilde{\mathcal O} \Bigl( n_S^{-\min \{ \frac{1}{2d+4}, \frac{\alpha}{2}, \frac{\beta}{2} \}} \Bigr) \\
        &\quad + \mathcal{O} \biggl( \frac{1}{\min_{k} \sqrt{n_T(k)}} \biggr).
        \end{aligned}
    \end{align*}
\end{proof}

\section{EXPERIMENTAL DETAILS}\label{appendix: experimental details}

\subsection{Within-Domain Transfer on Real-World Image Datasets}\label{subsection: Benchmark Classification Performance on Real-World Image Datasets}
To assess the practical utility of the DM framework on high-dimensional complex signals, we evaluate transfer from self-supervised pretraining to downstream classification within each of three standard benchmarks: CIFAR-10, CIFAR-100 \citep{Krizhevsky09cifar}, and STL-10 \citep{coates2011stl10}. The experiments compare the downstream classification accuracy of representations learned by DM and representative SSL methods. The statistical procedure and computational configurations are detailed below.

\paragraph{Dataset Specifications and Sampling} Following standard self-supervised evaluation protocols \citep{chen2020simclr, zbontar2021barlow, duan2025advssl}, we use the official training and test splits. For CIFAR-10 and CIFAR-100, $\mathcal{D}_S$ is the official training set with labels hidden during pretraining, and $\mathcal{D}_T$ is the same training set with labels used for probe fitting. For STL-10, pretraining uses the official training and unlabelled sets together, while probe fitting uses the labelled training set. In all cases, the official test set is held out from both pretraining and probe fitting.

\paragraph{Estimation Procedure} The training process follows a two-stage estimation paradigm. In the pretraining phase, images are randomly cropped and resized to $32 \times 32$ pixels for CIFAR-10/100 and $64 \times 64$ pixels for STL-10. We employ the Adam optimiser to update the encoder $f_{\bm{\theta}}$ using the unlabelled dataset $\mathcal{D}_S$. In the transfer stage, the backbone is frozen and both probes use its $512$-dimensional features before the projection head. We fit an affine linear probe to these backbone features, using a linear transformation from $\R^{512}$ to $\R^K$ followed by a softmax layer and a separate Adam optimiser to minimise the cross-entropy risk. We also report the performance of a $k$-nearest neighbours ($k$-NN) classifier with $k=5$ as a non-parametric probe.

\paragraph{Network Architectures} The encoder $f_{\bm{\theta}}$ uses a ResNet-18 backbone \citep{he2016resent} (\texttt{weights="DEFAULT"}) followed by a two-layer head with hidden width 1000 \citep{chen2020simclr}; throughout the manuscript, $f_{\bm{\theta}}$ denotes the resulting representation map used by the DM objective. The critic $g_{\bm{\omega}}$ is implemented as a three-layer MLP with a dimensionality path of $d^\star \to 128 \to d^\star \to 1$. We apply layer normalisation \citep{ba2016layernormalization} and LeakyReLU activations with slope 0.2 to the hidden layers of $g_{\bm{\omega}}$ for numerical stability.

\paragraph{Dual Distance Estimation via Gradient Penalty} To encourage the $1$-Lipschitz constraint required by Kantorovich--Rubinstein duality, we regularise the empirical critic objective. Following \citet{gulrajani2017wgangp}, the quantity maximised with respect to the critic is
\begin{equation}\label{eq: gp_refined}
    \widehat{\mathcal{W}}_{\mathrm{gp}}(f, g) = \E_{\mathcal{R} \sim \P_\mathcal{R}} \left\{ g(\mathcal{R}) \right\} - \E_{Z \sim \P_f} \left\{ g(Z) \right\} - \eta \cdot \E_{\hat{Z} \sim \P_{\hat{Z}}}  \left\{ \left( \norm{\nabla_{\hat{Z}} g(\hat{Z})}_2 - 1 \right)^2 \right\},
\end{equation}
where $\eta > 0$ is the penalty weight, set to $1$ in the image experiments. The random variable $\hat{Z} = u Z + (1 - u) \mathcal{R}$ with $u \sim \mathrm{Uniform}(0, 1)$ lies on the line segment joining a sampled representation $Z$ and a sampled reference point $\mathcal{R}$. The critic is updated five times for every encoder update. The gradient penalty regularises the critic during optimisation, while the theoretical analysis uses the Lipschitz-constrained critic class.

\paragraph{Optional batch-wise OT acceleration}
We also provide a critic-free computational variant, denoted DM-Batch, in the
code repository.  Given the normalised representations $\bm z_{i,1}$ and
$\bm z_{i,2}$ of two augmented views in a minibatch of size $B$, the method
first forms $B$ reference slots by repeating the $K'$ centres as evenly as
possible.  Writing these slots as $\{\bm s_j\}_{j=1}^B$, it computes the
balanced assignment
\begin{equation}\label{eq: supplementary batch OT}
    \hat\tau_B\in\argmax_{\tau\in P_B}
    \sum_{i=1}^B
    \left\langle\frac{\bm z_{i,1}+\bm z_{i,2}}{2},
    \bm s_{\tau(i)}\right\rangle,
\end{equation}
where $P_B$ is the set of permutations of $\{1,\ldots,B\}$.  This assignment
is obtained by the Hungarian algorithm.  Each assigned slot is independently
perturbed as in the reference construction and renormalised, giving a target
$\widetilde{\bm s}_i$.  The encoder is then updated directly by
\begin{equation}\label{eq: supplementary batch regression}
    \frac{1}{2Bd^\star}\sum_{i=1}^B\sum_{v=1}^2
    \norm{\bm z_{i,v}-\widetilde{\bm s}_i}_2^2
    +\frac{\lambda_{\mathrm B}}{Bd^\star}\sum_{i=1}^B
    \norm{\bm z_{i,1}-\bm z_{i,2}}_2^2.
\end{equation}
Thus the global distribution-matching problem is replaced computationally by
a balanced assignment within each minibatch followed by regression to the
assigned reference targets.  The variant removes the critic and its repeated
updates.  Table~\ref{tab: batchwise results} compares its representation quality
and training speed with the optimised critic-based DM implementation on
CIFAR-10. The reported epoch time decreases from $72.28$ to approximately
$31$ seconds, with linear-probe and $5$-NN accuracies of $90.22\%$ and
$87.16\%$, respectively.

\begin{table}[htbp]
    \centering
    \caption{CIFAR-10 representation quality and training speed of optimised DM and DM-Batch. Linear and $5$-NN denote test accuracy (\%); epoch time is measured on a single Tesla V100.}
    \label{tab: batchwise results}
    \begin{tabular}{lccc}
        \toprule
        Method & Linear & $5$-NN & Seconds per epoch \\
        \midrule
        Optimised DM & $92.11$ & $89.17$ & $72.28$ \\
        DM-Batch & $90.22$ & $87.16$ & $\approx 31$ \\
        \bottomrule
    \end{tabular}
\end{table}

DM-Batch is provided as an optional acceleration; the principal results in the
main text use the stochastic dual estimator in \eqref{eq: nn objective}.

\paragraph{Hyperparameters and Optimisation} Across all datasets, we set the number of reference components to $K' = 384$, the radius to $R = 1$, and the concentration parameter to $\epsilon = 10^{-3}$. The projection-head output dimension $d^\star$ is fixed at 384. Pretraining is conducted for 1000 epochs with a mini-batch size of 512. The learning rates for the encoder and critic are fixed at $3 \times 10^{-5}$ and $10^{-3}$, respectively, with weight decays of $10^{-4}$ for both. A learning rate warm-up is applied for the initial 500 iterations. In the transfer stage, the classification head is estimated over 500 epochs with a batch size of 512, using an exponential learning rate decay from $10^{-2}$ to $10^{-6}$ and a weight decay of $5 \times 10^{-6}$.

\paragraph{Data Augmentation Protocol} The augmentation family $\mathcal{A}$ follows the composition of stochastic transformations proposed in \citet{chen2020simclr}. For any instance $X$, the augmented views are generated by sequentially applying: (\romannumeral 1) random resized cropping with scale $\in [0.2, 1.0]$ and aspect ratio $\in [3/4, 4/3]$; (\romannumeral 2) horizontal flipping with probability $0.5$; (\romannumeral 3) color jittering with intensity $(0.4, 0.4, 0.4, 0.1)$ and probability $0.8$; and (\romannumeral 4) grayscaling with probability $0.2$.

\paragraph{Computational Environment} All experiments were executed on a single NVIDIA Tesla V100 GPU utilizing the PyTorch framework (version 2.2.1+cu118) and CUDA version 11.8.

\subsection{Qualitative inspection of spherical representations}\label{subsection: qualitative spherical groups}

Figure~\ref{fig: spherical groups} uses a saved 1000-epoch CIFAR-10 DM checkpoint as a qualitative illustration separate from the benchmark comparisons. Its architecture is a CIFAR-stem ResNet-18 followed by a $512\to1000\to384$ projection head with batch normalisation and ReLU. We extract the projected outputs for all $10,000$ official test images and normalise each output to unit Euclidean norm. The deterministic input transformation converts each image to a tensor and normalises its channels using means $(0.4914,0.4822,0.4465)$ and standard deviations $(0.2023,0.1994,0.2010)$; no model parameters are updated.

The checkpoint does not retain the training reference centres. We therefore inspect its signed-coordinate geometry: each representation is assigned the index and sign of its largest-magnitude coordinate, without using labels or fitting a clustering model. Equivalently, each unit representation is assigned to the signed coordinate direction with the largest cosine similarity. These groups describe the learned spherical geometry rather than verified original reference-component identities. After assignment, we compute group sizes and coarse-class proportions. For each coarse category, the candidate pool consists of the three largest groups containing at least ten images and having coarse-class purity at least $0.8$. The figure displays two contrasting candidate groups for each of horse, truck and dog. For every group, the left panel contains four members with the largest assignment margins, where the margin is the difference between the largest and second-largest absolute coordinates. The right panel shows the first four of eight members sampled without replacement from the group after excluding its eight highest-margin members, using a fixed random seed of $0$ and without excluding off-class images. The observations concern recurring appearance and viewpoint patterns, not named fine-label categories.

\subsection{Experimental Details for Quantitative Simulation Studies}\label{subsection: experimental details simulation}

This section details the configuration for the quantitative simulations (Studies I and II), including the results in Table~\ref{tab:sim_results}. Both studies are implemented in PyTorch and controlled via a fixed random seed (42) to ensure reproducibility of the data generating process and model initialization. The separate configuration for Figure~\ref{fig:sim_tsne} is given in Section~\ref{subsection: experimental details tsne}.

\paragraph{Data Generating Process (DGP)}
The synthetic environment consists of a latent space $\mathbb{R}^{32}$ with $K=10$ semantic categories. The class centres are fixed as orthogonal standard basis vectors. Latent variables are sampled as $\bm{z} = \bm{c}_y + \bm{\epsilon}$, where $\bm{\epsilon} \sim \mathcal{N}(\bm{0}, 0.4^2 \mathbf{I})$. To simulate high-dimensional observations, we apply a fixed warping map $G: \mathbb{R}^{32} \to \mathbb{R}^{512}$, implemented as a three-layer MLP with $\tanh$ activations. The resulting ambient features are standardised by their empirical mean and standard deviation. For self-supervised signals, positive pairs are generated by adding isotropic noise $\bm{\delta} \sim \mathcal{N}(\bm{0}, 0.05^2 \mathbf{I})$ to the latent state $\bm{z}$ before passing through the warping map.

\paragraph{Model Architectures and Optimisation}
The encoder $f_{\bm{\theta}}$ is a two-layer MLP ($512 \to 256 \to 32$) with ReLU activations. The critic $g_{\bm{\omega}}$ consists of three layers ($32 \to 128 \to 128 \to 1$) with LeakyReLU (0.2). During pre-training, the encoder output is $L_2$-normalised when computing the alignment and distribution matching losses. We use the Adam optimiser with encoder and critic learning rates of $3 \times 10^{-5}$ and $1 \times 10^{-4}$, respectively. The critic is updated 5 times for every encoder update, and a gradient penalty weight of 10.0 is used to encourage the $1$-Lipschitz constraint. The weight for the Mallows distance ($\lambda$) is fixed at 1.5.

\paragraph{Study-Specific Settings}
In both studies, downstream classification uses multiclass logistic regression. For each comparison, the classifiers use the same labelled target training set ($n_T=200$) and test set ($2,000$ observations). The latent-oracle baseline fits this type of classifier directly to the true latent representations $Z$ in the labelled target training set by minimising cross-entropy and evaluates it on the corresponding latent test representations.
\begin{itemize}
    \item \textbf{Simulation I (Rectification Capacity):} The unlabelled source size is fixed at $n_S = 15,000$, and the model is trained for $19,000$ iterations. The warping map has layer widths $32 \to 64 \to 128 \to 512$, and we use $K'=32$ reference components with $R=1$. Two augmented views are generated once for each source observation and kept fixed during training; minibatches of size 256 are sampled with replacement. Downstream evaluation fits a multiclass logistic-regression model (\texttt{max\_iter=5000}) to the unnormalised outputs $f(\bm{x})$ of the frozen encoder.
    \item \textbf{Simulation II (Dependence on Source Sample Size):} We vary the source sample size $n_S \in \{$1k, 5k, 10k, 20k$\}$. The number of training iterations is set to $T = 4000 + n_S$, so the optimisation budget also increases with the source sample size. The warping map, reference distribution, augmentation pairs and minibatch settings are as in Simulation I. Downstream evaluation fits a multiclass logistic-regression model (\texttt{max\_iter=3000}) to the unnormalised outputs of the frozen encoder. The classification errors are reported on a log--log scale to illustrate performance as the unlabelled sample and optimisation budget increase together.
\end{itemize}

\subsection{Experimental Details for t-SNE Visualisation}\label{subsection: experimental details tsne}

To illustrate the geometric rectification associated with Distribution Matching (DM), we conduct a visualisation study using t-SNE \citep{hinton2008tsne}. Figure~\ref{fig:sim_tsne} uses the visualisation-specific configuration below, distinct from the quantitative simulations in Section~\ref{subsection: experimental details simulation}. The classification errors in Table~\ref{tab:sim_results} belong to Simulation I, not to this visualisation study.

\paragraph{Visualisation-Specific DGP} 
For visual clarity in a two-dimensional projection, we reduce the number of latent categories to $K = 5$ and the latent dimension to $10$. The within-class noise level is set to $\sigma_c = 0.35$, while the augmentation noise remains at $\sigma_a = 0.05$. The observations are generated via a fixed non-linear warping map $G: \mathbb{R}^{10} \to \mathbb{R}^{512}$, followed by the same standardization procedure described in Appendix~\ref{subsection: experimental details simulation}.

\paragraph{Pre-training and Optimisation}
The encoder and critic architectures follow the same MLP structures as in the quantitative simulations. We train the model for $10,000$ iterations with an unlabelled source pool of $n_S = 15,000$. The learning rates are fixed at $3 \times 10^{-5}$ for the encoder and $1 \times 10^{-4}$ for the critic. The distribution matching weight $\lambda$ is set to $1.5$, and the $1$-Lipschitz constraint is encouraged using a gradient penalty weight of $10.0$.

\paragraph{t-SNE Configuration}
Upon completion of pre-training, we extract representations for $2,000$ independent test samples. The visualisation contrasts two feature spaces:
\begin{itemize}
    \item \textbf{Ambient Space ($X$):} t-SNE is applied directly to the $512$-dimensional standardised observations.
    \item \textbf{Representation Space ($f(X)$):} t-SNE is applied to the $L_2$-normalised outputs of the frozen encoder. Normalisation is used here to emphasise the geometric structure induced by the spherical reference prior.
\end{itemize}
For both spaces, we employ the TSNE implementation from \texttt{scikit-learn} with
\begin{equation*}
\begin{gathered}
\texttt{n\_components=2},\qquad \texttt{perplexity=40},\\
\texttt{init='pca'},\qquad \texttt{random\_state=42},
\end{gathered}
\end{equation*}
where the fixed random state makes the projection reproducible.

\end{appendices}

\end{document}